\documentclass{article}
\usepackage{microtype}
\usepackage{graphicx}
\usepackage{subcaption}
\usepackage{booktabs} 
\usepackage{comment}
\usepackage{wrapfig}
\usepackage{etoolbox}

\usepackage{amsmath,amsfonts,bm}

\def\eqref#1{equation~\ref{#1}}

\def\1{\bm{1}}

\def\ra{{\textnormal{a}}}
\def\rb{{\textnormal{b}}}
\def\rc{{\textnormal{c}}}

\def\rx{{\textnormal{x}}}
\def\ry{{\textnormal{y}}}

\def\rva{{\mathbf{a}}}

\def\rvh{{\mathbf{h}}}

\def\rvo{{\mathbf{o}}}

\def\rvq{{\mathbf{q}}}

\def\rvv{{\mathbf{v}}}

\def\rvx{{\mathbf{x}}}
\def\rvy{{\mathbf{y}}}
\def\rvz{{\mathbf{z}}}

\def\rmG{{\mathbf{G}}}
\def\rmH{{\mathbf{H}}}

\def\rmO{{\mathbf{O}}}

\def\rmV{{\mathbf{V}}}
\def\rmW{{\mathbf{W}}}
\def\rmX{{\mathbf{X}}}

\def\rmZ{{\mathbf{Z}}}

\DeclareMathAlphabet{\mathsfit}{\encodingdefault}{\sfdefault}{m}{sl}
\SetMathAlphabet{\mathsfit}{bold}{\encodingdefault}{\sfdefault}{bx}{n}

\newcommand{\E}{\mathbb{E}}

\DeclareMathOperator*{\argmin}{arg\,min}

\usepackage[dvipsnames]{xcolor}
\usepackage{array}
\usepackage{colortbl}
\usepackage[hyperfootnotes=false]{hyperref}
\usepackage{xurl}
\usepackage{etoc}

\usepackage[main,final]{neurips_2026}
\setcitestyle{authoryear,round}

\usepackage{amsmath}
\usepackage{amssymb}
\usepackage{mathtools}
\usepackage{amsthm}

\usepackage{adjustbox}
\usepackage[capitalize,noabbrev]{cleveref}
\theoremstyle{plain}

\theoremstyle{definition}

\theoremstyle{remark}

\definecolor{refcolor}{RGB}{70, 60, 140}
\definecolor{cardline}{HTML}{D7DEE9}

\definecolor{cliqueindigo}{HTML}{2C2F6E}
\definecolor{cliquecopper}{HTML}{B76B43}
\definecolor{cliqueteal}{HTML}{146B6B}

\definecolor{resourcetext}{HTML}{075F73}
\definecolor{resourceaccent}{HTML}{168DA6}
\definecolor{resourcefill}{HTML}{EAF8FB}
\definecolor{CFheading}{HTML}{075F73}
\definecolor{CFheaderfill}{HTML}{DFF2F6}
\definecolor{CFaccent}{HTML}{168DA6}
\definecolor{CFmetricfill}{HTML}{EFF9FB}
\definecolor{CFbestink}{HTML}{087568}
\definecolor{CFsecondink}{HTML}{245DA8}
\definecolor{CFbodyink}{HTML}{222932}
\newcommand{\CFbest}[1]{\textcolor{CFbestink}{\textbf{#1}}}
\newcommand{\CFsecond}[1]{\textcolor{CFsecondink}{#1}}
\newcommand{\CFmetric}[1]{\textcolor{CFheading}{\textbf{\boldmath #1}}}
\newcommand{\CFheading}[1]{\textcolor{CFheading}{#1}}
\newcommand{\CFnumber}[1]{\makebox[27pt][c]{\strut#1}}
\newsavebox{\CFtablemeasure}
\newcommand{\CFwidetable}[1]{%
  \begingroup
  \setlength{\tabcolsep}{4pt}%
  \renewcommand{\arraystretch}{1.16}%
  \setlength{\extrarowheight}{0.25pt}%
  \setlength{\aboverulesep}{0pt}%
  \setlength{\belowrulesep}{0pt}%
  \arrayrulecolor{CFaccent}%
  \sbox{\CFtablemeasure}{\begin{tabular}{lcccccc}#1\end{tabular}}%
  \setlength{\tabcolsep}{\dimexpr\tabcolsep+(\linewidth-\wd\CFtablemeasure)/14\relax}%
  \begin{tabular}{lcccccc}#1\end{tabular}%
  \arrayrulecolor{black}%
  \endgroup
}
\newcommand{\resourcelink}[2]{%
  \begingroup
  \setlength{\fboxsep}{0.8pt}%
  \setlength{\fboxrule}{0.35pt}%
  \href{#1}{\fcolorbox{resourceaccent}{resourcefill}{\textcolor{resourcetext}{\textbf{#2}}}}%
  \endgroup
}

\hypersetup{
    colorlinks=true,
    linkcolor=cliqueindigo,
    citecolor=cliqueindigo,
    urlcolor=cliqueindigo,
    pdftitle={Offline Materials Optimization with CliqueFlowmer},
    pdfauthor={Jakub Grudzien Kuba, Benjamin Kurt Miller, Sergey Levine, Pieter Abbeel}
}
\usepackage[disable,textsize=tiny]{todonotes}

\title{Offline Materials Optimization with CliqueFlowmer}

\author{
Jakub Grudzien Kuba\thanks{BAIR, UC Berkeley, Berkeley, California, USA. Corresponding author: \texttt{kuba@berkeley.edu}}
\And
Benjamin Kurt Miller\thanks{FAIR, Meta, San Francisco, California, USA}
\And
Sergey Levine\footnotemark[1]
\And
Pieter Abbeel\footnotemark[1]
}

\begin{document}

\etocdepthtag.toc{main}

\maketitle

\begin{abstract}
Recent advances in deep learning have inspired neural network-based approaches to \emph{computational materials discovery} (CMD).
A plethora of problems in this field involve finding materials that optimize a target property.
Nevertheless, the increasingly popular generative modeling methods are ineffective at boldly exploring attractive regions of the materials space due to their maximum likelihood training.
In this work, we offer an alternative CMD technique based on offline \emph{model-based optimization} (MBO) that fuses direct optimization of a target material property into generation.
To that end, we introduce a domain-specific model, \emph{CliqueFlowmer}, that incorporates recent advances in clique-based MBO into transformer and flow generation.
We validate this model's optimization abilities and show that materials it produces strongly outperform those from generative baselines. 
To support specialized materials discovery applications and broader interdisciplinary research, we release our
\resourcelink{https://github.com/znowu/CliqueFlowmer}{code},
\resourcelink{https://huggingface.co/iamkuba/CliqueFlowmer}{model weights}, and an
\resourcelink{https://cliqueflowmer.znowu.workers.dev}{interactive demo}.
\end{abstract}
\section{Introduction}
Large neural network models have recently enabled solving challenging artificial intelligence (AI) tasks such as language modeling, automated coding, and image generation \citep{achiam2023gpt, team2024gemini, ho2020denoising, esser2024scaling}.
Meanwhile, scientific problems that deal with the world of atoms, rather than the world of bits, are yet to benefit from this revolution \citep{Thiel_2025_EconTimes}.
Indeed, exploration of new physical breakthroughs continues to take place in physical wet labs and is driven by costly physical experimentation \citep{xiang1995combinatorial, freese2024relevance, shahzad2024accelerating}.
One such problem is \emph{computational materials discovery} (CMD), wherein scientists strive to invent new chemical structures that display properties absent in known materials \citep{danielson1997combinatorial, jain2013commentary, xue2016accelerated, lin2025high}.
A systematic, sample-efficient method for CMD has the potential to deliver structures that serve as catalysts for key energy-conversion reactions and as functional components of advanced biomaterials.
Such advances would greatly accelerate the development of clean-energy technologies and medical therapies, effectively extending the recent AI-driven progress from the world of bits to the world of atoms \citep{hutmacher2000scaffolds, nocera2009chemistry, gokcekuyu2024artificial, han2025ai}.

Having realized the importance of this problem, AI researchers have been increasingly turning their attention to AI-driven CMD. 
Most commonly, the introduced methods have been utilizing the celebrated \emph{diffusion} and \emph{flow} models that had become very successful in the domain of image generation \citep{ho2020denoising, lipman2022flow, miller2024flowmm, inizangenerative}.
While effective at harnessing the distribution of viable materials presented to them in the dataset, likelihood-based generative models do not actively explore the relation between the materials and their properties.
This is a major impediment in settings where CMD is expected to optimize materials with respect to a specified metric \citep{yang2024generative, havens2025adjoint}.
In the meantime, a new paradigm, known as \emph{offline model-based optimization} (MBO) has made steps towards techniques that optimize \emph{scientific designs} by bootstrapping models trained entirely on offline data \citep{kumar2021data, trabucco2022design, kuba2024cliqueformer}. 
Nevertheless, MBO methods have been mainly deployed in standard benchmark tasks, free, for example, of the challenging intricacies of CMD in which data is inherently of the hybrid discrete-continuous nature, irregular shape, and subject to physical constraints.

To address the need for AI-driven CMD and the limitations of generative approaches, we introduce \emph{CliqueFlowmer}---a model that renders materials data tractable by MBO and enables direct optimization of material structures. 
At its core is an auto-encoder that transforms multi-modal, irregular materials data into finite-dimensional vectors that can be optimized with \emph{clique-based} MBO \citep{kuba2024cliqueformer} and decoded back into material form. 
Because CliqueFlowmer is built from general-purpose deep learning primitives, it makes CMD more accessible to the broader AI community while reducing reliance on specialized domain-specific software.
Empirically, using data from Materials Project \citep{jain2013commentary} together with M3GNet \citep{chen2022universal} and MEGNet \citep{chen2019graph} as property oracles, we show that materials optimized by CliqueFlowmer vastly outperform those sampled by generative baselines. While our experiments focus on offline materials optimization, the pretrained auto-encoder also provides a practical continuous representation space for materials. 
\begin{wrapfigure}{r}{0.45\textwidth}
    \centering
    \vspace{-3pt}
    \includegraphics[width=0.45\textwidth]{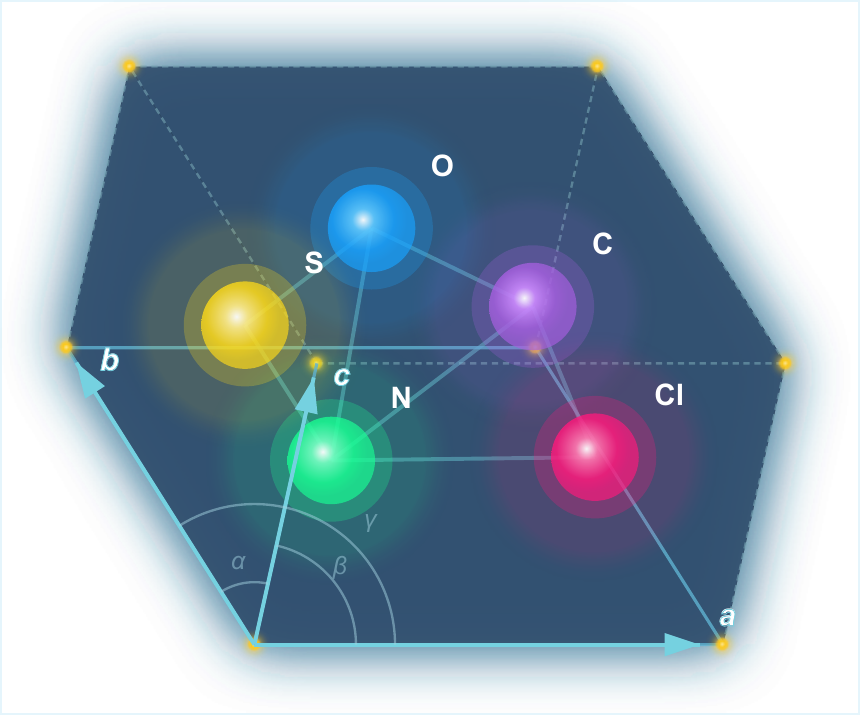}
    \caption{The unit cell of a hypothetical material. The cell has a shape of a parallelepiped determined by three axes, $\vec{\ra}$, $\vec{\rb}$, and $\vec{\rc}$. The angles between the axes are $\operatorname{ang}(\vec{\rb}, \vec{\rc})=\alpha$, $\operatorname{ang}(\vec{\rc}, \vec{\ra})=\beta$, $\operatorname{ang}(\vec{\ra}, \vec{\rb})=\gamma$. In this cell, there are five atoms, whose type sequence is $\rva=[\text{N, Cl, C, O, S}]$.}
    \label{fig:unit-cell}
    \vspace{-47pt}
\end{wrapfigure}
\vspace{-5pt}
\section{Background}
\vspace{-5pt}
This section provides the necessary background on the key notions discussed by our work.
First, we lay down foundations of material modeling that enables us to use machine learning methods in CMD.
Then, we introduce the basic ideas of continuous normalizing flows (\emph{i.e.,} flow matching).
Lastly, we state the problem that we aim to solve (CMD) and formulate it through the lens of offline MBO.

\subsection{Modeling Materials}
We characterize a material  $\mathsf{M}$ by its \emph{unit cell}---the material's unit amount. 
The cell is a parallelepiped with axis lengths $(a, b, c)\in\mathbb{R}_{+}^{3}$ in \AA\ and interaxial angles $(\alpha, \beta, \gamma)\in(0, \pi)^3$ in radians.
The content of the cell is a set of $N_{\text{atom}}$ (which varies between materials) atoms that is represented by the sequence $\rva\in\mathcal{A}^{N_{\text{atom}}}$ of their types, as well as their positions $\rmX\in [0, 1)^{N_{\text{atom}}\times 3}$ expressed in the basis induced by $(a, b, c)$ and $(\alpha, \beta, \gamma)$.
Thus, the space of materials $\mathsf{M}$ can be embedded in the product $\mathcal{M}=\mathbb{R}_{+}^{3}\times(0, \pi)^{3}\times(\mathcal{A}\times [0,1)^{3})^{*}$, and a single material can be characterized as a tuple $[\rva, (a,b,c), (\alpha, \beta, \gamma), \rmX]$.
\footnote{$\mathcal{S}^{*}$ denotes the countable union of the products of the set $\mathcal{S}$ with itself. That is, $\mathcal{S}^{*}=\cup_{n=1}^{\infty}\mathcal{S}^{n}$.}
We denote the geometrical part of the material as $\rmG=[(a,b,c), (\alpha, \beta, \gamma), \rmX]$, and thus can write $[\rva, \rmG]$ to represent material $\mathsf{M}$.

\subsection{Flow Matching}
\label{subsec:flow}
The goal of a continuous normalizing flow \citep{lipman2022flow} is to learn sampling from a data distribution $p_{\text{data}}(\rvx)$. 
To that end, one sets the \emph{target} distribution $p_{1}(\rvx)=p_{\text{data}}(\rvx)$, as well as \emph{source} distribution $p_{0}(\rvx)=p_{\text{source}}(\rvx)$, such as the standard-normal. 
To learn to turn a sample from $p_{\text{source}}(\rvx)$ into one from $p_{\text{data}}(\rvx)$, one interpolates noise and data,
\begin{align}
    \rvx_t = (1-t) \cdot \rvx_0 + t \cdot \rvx_1, \quad \text{where} \quad \rvx_{0} \sim &p_{0}(\rvx), \ \rvx_1 \sim p_{1}(\rvx) \ \ \ \ \& \ \ \ \  t\sim p_{\text{time}}(t), \nonumber
\end{align}
and passes the mixture, together with timestep information, to a \emph{velocity} neural network that minimizes
\begin{align}
    \E_{\rvx_{0}\sim p_{0}, \rvx_{1}\sim p_1, t\sim p_{\text{time}}}\big[\big(v_{\theta}(\rvx_t, t) - (\rvx_1-\rvx_0)\big)^2\big]. \nonumber
\end{align}
Once learned, the velocity network can be used to sample from the target distribution by solving the following ordinary differential equation (ODE) with initial conditions,
\begin{align}
    \text{d}\rvx_t = v_{\theta}(\rvx_t, t) \text{d}t, \ \quad \rvx_{0}\sim p_{0}(\rvx), \nonumber
\end{align}
from $t=0$ to $t=1$, which results in a sample $\rvx_1 \sim p_{1}(\rvx)$.
Typically, the ODE is solved numerically, by discretizing the time interval $[0, 1]$ into $N_{\text{step}}$ steps and using, \emph{e.g.,} the Euler method.
In this paper, we will use continuous normalizing flows to model the geometry of materials.

\subsection{Model-Based Optimization in Computational Materials Discovery}
We consider a function $f(\mathsf{M})\in\mathbb{R}$, often referred to as \emph{target property}, and assume that we are given a finite dataset $\mathcal{D}=\{\mathsf{M}^{i}, \ry^i=f(\mathsf{M}^i)\}_{i=1}^{N_{\text{data}}}$ of $N_{\text{data}}$ examples of materials and their values of the property.
Given this dataset, our goal is to discover materials that optimize the property,
\begin{align}
    \min_{\mathsf{M} \in\mathcal{M}} f(\mathsf{M})
    \label{eq:mbo}
\end{align}
without the necessity to evaluate consecutive candidates in a wet lab\footnote{Maximization problems can be represented analogously and solved using the same techniques.} \citep{danielson1997combinatorial}.
That is, we want to do this \emph{fully offline}.
The standard approach to this problem proceeds in two stages: \emph{1)} train a generative model $p_{\theta}(\mathsf{M})$ to sample candidate materials, and \emph{2)} select the most promising candidate among $N_{\text{prop}}$ proposals \citep{gomez2018automatic, noh2019inverse, park2024has, zeni2025generative}.
However, this approach is quite inefficient for optimization, since sampling from $p_{\theta}(\mathsf{M})$ only explores the regions of the materials space supported by the training data; methods that more effectively reach promising regions are therefore desirable (more background in Appendix \ref{sec:appendix-background}).

In this work, we propose to solve the optimization problem in Equation~(\ref{eq:mbo}) directly. 
Specifically, we use tools from offline \emph{model-based optimization} (MBO) to model the target property $f(\mathsf{M})$ and search for candidate minimizers. 
Since the main obstacle is the irregular structure of the materials space, we adopt the \emph{clique}-based MBO paradigm, which enables optimization over irregular designs by encoding them into clique-structured latent representations and optimizing those latents \citep{kuba2024cliqueformer}. 
To make this approach applicable to CMD, for the first time, we introduce \emph{CliqueFlowmer}, a neural-network model that maps materials into representations tractable by MBO.

\vspace{-5pt}
\section{CliqueFlowmer}
\vspace{-5pt}
This section introduces the architecture and the training algorithm of our MBO model for CMD. 
The neural network consists of several carefully designed components, built from standard deep learning primitives, that enable it to handle materials data.
In Section \ref{sub:enc} we introduce an encoder that maps a material $\mathsf{M}$ to a continuous latent representation $\rvz$.
Then, Section \ref{sub:pred} describes a decomposable model that predicts the property of interest from $\rvz$.  
Section \ref{sub:dec} delineates a decoder that reconstructs the material---both the atom types and the geometry, from the representation.
As we demonstrate later, such a setup allows to parameterize materials as continuous vectors which can be optimized for properties with gradient-based techniques (see Figure (\ref{fig:mbo-clique})).

\vspace{-5pt}
\subsection{Encoder}
\vspace{-5pt}
\label{sub:enc}
Our encoder has to combine four distinct pieces of information: lattice lengths $(a,b,c)$, lattice angles $(\alpha,\beta,\gamma)$, atom positions $\rmX$, and atom types $\rva$ (the latter two being irregular), and produce a fixed-dimensional continuous vector $\rvz\in\mathbb{R}^{d_{\rvz}}$.
To do this, first, we map the continuous inputs to vectors of size $d_{\text{model}}$ with their corresponding MLPs, 
\begin{align}
    \vspace{-10pt}
    \rvh^{\text{len}} = \operatorname{MLP}^{\text{len}}_{\theta}(a,b,c), \ 
    \rvh^{\text{ang}} =\operatorname{MLP}^{\text{ang}}_{\theta}(\alpha, \beta, \gamma) \in \mathbb{R}^{d_{\text{model}}}, \ \& \ \ 
    \rmH^{\text{pos}} = \operatorname{MLP}^{\text{pos}}_{\theta}(\rmX) \in \mathbb{R}^{N_{\text{atom}} \times d_{\text{model}}}\nonumber
    \vspace{-10pt}
\end{align}
and we map the atom types to continuous $d_{\text{model}}$-dimensional embeddings $\rmH^{\text{atom}} \in \mathbb{R}^{N_{\text{atom}} \times d_{\text{model}}}$.
The MLP-produced hidden states are then concatenated, $\rmH^{\text{in}} = [\rvh^{\text{len}}, \rvh^{\text{ang}}, \rmH^{\text{pos}}]$, and passed into a transformer, where they are conditioned by $\rmH^{\text{atom}}$ via adaptive layer-norm \citep[AdaLN]{peebles2023scalable} 
in place of the standard layer-norm \citep{lei2016layer, vaswani2017attention},
\begin{align}
    \rmH^{\text{out}} = T^{\text{enc}}_{\theta}(\rmH^{\text{in}}, \rmH^{\text{atom}}).\nonumber
    \vspace{-15pt}
\end{align}
\begin{figure*}[t]
    \centering
    \includegraphics[width=\textwidth]{figures/figma/CliqueFlowmer-figma-glow-bigger.pdf}
    \caption{\small Computational materials discovery through MBO with CliqueFlowmer.
    Known materials are encoded by a transformer, and an attention-based pooling layer, into a fixed-dimensional latent space. 
    The latent variable admits a clique decomposition, with respect to the target property,
    where each clique contributes additively to the target property.
    The representations are optimized with evolution strategies.
    Atom types are then decoded with an autoregressive transformer and beam search. 
    The flow model reconstructs the material's geometry from the atom sequence and the latent representation by which it is conditioned via cross-attention.}
    \label{fig:mbo-clique}
    \vspace{-10pt}
\end{figure*}

For a single instance of $N_{\text{atom}}$ atoms in the unit cell, the output of the transformer is a tensor of shape $(2+N_{\text{atom}})\times d_{\text{model}}$, thus depending on the cell size. 
To produce a fixed-dimensional representation $\rvz$, we pool the tensor along the first dimension, using attention with a learnable query vector $\rvq\in\mathbb{R}^{d_{\text{model}}}$, yielding a hidden state,
\begin{align}   
    \label{eq:pool}
    \rvh^{\text{pool}} = \operatorname{Att}(\rvq, \rmH^{\text{out}}\rmW^{K}, \rmH^{\text{out}}\rmW^{V}) \in\mathbb{R}^{d_{\text{model}}},
\end{align}
which is further post-processed by GELU \citep{hendrycks2016gaussian} and layer-norm, yielding $\rvh^{\text{post}}$.
It is then fed to  a linear layer that produces the mean $\mu_{\rvz}$ and log-standard deviation $\log \sigma_{\rvz}$ parameters of the normal distribution of the latent representation.
We highlight that the pooling layer from Equation (\ref{eq:pool}) is what enables us to navigate the transdimensional materials space $\mathcal{M}$ with a fixed-dimensional vector space, paving the road for MBO.

\subsection{Predictor}
\label{sub:pred}
Once sampled, in addition to the standard flat form, the latent vector attains a form of a chain of cliques $\texttt{chain}(\rvz, d_{\text{clique}}, d_{\text{knot}})$ of clique size $d_{\text{clique}}$ and knot size $d_{\text{knot}}$. 
That is, it is a $N_{\text{clique}}\times d_{\text{clique}}$ matrix $\rmZ$ with entry values from $\rvz$ such that $\rmZ_{c,i}=\rvz_{k}$, where $k=(c-1)\cdot (d_{\text{clique}}-d_{\text{knot}}) + i$, \emph{i.e.}, the last $d_{\text{knot}}$ entries of row $c$ and the first $d_{\text{knot}}$ entries of row $c+1$, for $c=1, \dots, N_{\text{clique}}-1$, are equal.
We write $\rmZ_c$ to denote its $c$th row.
The matrix is then fed to an MLP predictor of property $\rvy=f(\mathsf{M})$ that decomposes its prediction over the cliques of $\rmZ$,
\begin{align}
    f(\mathsf{M}) \approx f_{\theta}(\rvz) = \sum_{c=1}^{N_{\text{cliques}}} f_{\theta}(\rmZ_{c}, c). \nonumber
    \vspace{-10pt}
\end{align}

Imposing such a decomposable structure on the latent space is known to improve the effectiveness of MBO \citep[\& Appendix \ref{sec:appendix-fgm}]{kuba2024cliqueformer}. 
Intuitively (see Figure \ref{fig:mbo-clique}), it enables composing optimal in-distribution examples of each clique to form a competitive in-distribution solution---a property also known as \emph{stitching} \citep{fu2020d4rl, shin2025treatment}.
See Appendix \ref{appendix:clique} for ablation studies.

\subsection{Decoder}
\label{sub:dec}
The decoder consists of two modules: the atom type decoder and the geometry decoder. The former infers the atom types given the latent representation, $p_{\theta}(\rva|\rvz)$, and the latter infers the unit cell geometry given the latent representation and the atom types, $p_{\theta}(\rmG|\rvz, \rva)$.

\vspace{-5pt}
\paragraph{Atom types.} The atom type decoder is quite straightforward---we utilize the standard causal transformer with next-token prediction which we condition on the latent representation via AdaLN. 
To that end, we map the $d_{\rvz}$-dimensional $\rvz$ to the $d_{\text{model}}$-dimensional space and then use it in combination with adaptive layer-norm inside of the causal transformer to produce a hidden state 
\begin{align}
    \rvz^{\text{mod}} = \operatorname{LayerNorm}(\operatorname{GELU}(\operatorname{Lin}_{\theta}(\rvz))),
    \label{eq:atom-type-mod}
    \quad \& \quad 
    \rvh_{k+1} = T^{\text{dec}}_{\theta}(\rva_{0:k}, \rvz^{\text{mod}})_{k+1},
\end{align}
for $k=0, \dots, N_{\text{atom}}$, where $\ra_{0}$ and $\ra_{N_{\text{atom}}+1}$ are $\langle\operatorname{Start}\rangle$ and $\langle\operatorname{Stop}\rangle$ tokens.
We found the layer-norm very helpful at ensuring equitable training progress of all modalities.
That hidden state is then refined into log-likelihoods with an MLP and a log-softmax layer that predicts the next (atom type) token.

\paragraph{Geometry.} We decode the shape of the unit cell and the atom positions with a continuous normalizing flow, conditioned on $\rvz$, for which we build another specialized transformer.
We use the flow-matching framework, wherein we decode the material geometry $\rmG = [(a,b,c), \ (\alpha, \beta, \gamma), \ \rmX]$ by, first, initializing at a sample from a prior distribution and, then, solving an ODE,
\begin{align}
    \text{Sample} \quad \rmG_{0} = [(a,b,c), \ (\alpha, \beta, \gamma), \ \rmX]_{0} \sim p_{0}(\rmG | \rva),\quad 
    \text{Solve} \quad 
    \text{d}\rmG_{t} = V_{\theta}(\rmG_{t}, t | \rva, \rvz)\text{d}t,  \nonumber
\end{align}
from $t=0$ to $t=1$,
where $V_{\theta}(\rmG_t, t | \rva, \rvz)$ is a transformer denoiser.
The prior distribution $p_{0}(\rmG|\rva)=p^{\text{len}}_{0}(a,b,c|\rva)\cdot p^{\text{ang}}_{0}(\alpha, \beta, \gamma|\rva)\cdot p^{\text{pos}}_{0}(\rmX|\rva)$ is modeled meticulously---first, the angles are modeled as random uniform in interval $(\pi/3, 2\pi/3)$, and positions are random uniform on $(0, 1)$.
The prior over the lengths is chosen so that the density of the unit cell $\operatorname{den}(\mathsf{M})=N_{\text{atom}}/\operatorname{Vol}(\mathsf{M})$ is invariant to the number of atoms (more details in Appendix \ref{app:prior}).
Similarly to the encoder, given a noisy unit cell shape and atom positions $\rmG_{t} = [(a,b,c), (\alpha, \beta, \gamma), \rmX]_{t}$ at time $t$, we embed them in the $d_{\text{model}}$-dimensional space with MLPs. 
This is then fed to a (non-causal) transformer that is conditioned on the embeddings of the decoded atom types $\rmH^{\text{atom}}$ via AdaLN. 
Crucially, each block of the transformer contains a cross-attention layer \citep{vaswani2017attention} between the hidden states and the latent representations in the form $\rmZ=\texttt{chain}(\rvz, d_{\text{clique}}, d_{\text{knot}})$. 
More specifically, cliques of $\rmZ$ get first mapped to the $d_{\text{model}}$-dimensional space
and a cross-attention layer is added between the self-attention and feed-forward layers of the transformer block
\begin{align}
    \rmH^{\rvz} = \operatorname{LayerNorm}(\operatorname{GELU}(\operatorname{MLP}^{\text{lat}}_{\theta}(\rmZ)))
    \quad \& \quad 
    \rmH^{\text{cross}}_{t} = \rmH^{\text{self}}_{t} + \operatorname{CrossAtt}(\rmH^{\text{self}}_{t}, \rmH^{\rvz})\nonumber
\end{align}
(more information in Appendix \ref{appendix:cross-attention}).
The output hidden state $\rmO_{t}$ of the transformer is then decomposed into the lengths, angles, and position parts, and each of them is used to predict the corresponding modality with an MLP,
\begin{align}
    \hat{\rvv}^{\text{len}}_{t} = \operatorname{MLP}^{\text{len}}_{\theta}(\rvo^{\text{len}}_{t}), \quad 
    \hat{\rvv}^{\text{ang}}_{t} = \operatorname{MLP}^{\text{ang}}_{\theta}(\rvo^{\text{ang}}_{t}), \quad 
    \hat{\rmV}^{\text{pos}}_{t} = \operatorname{MLP}^{\text{pos}}_{\theta}(\rmO^{\text{pos}}_{t})\nonumber
\end{align}
which together form the prediction from our denoiser, $V_{\theta}(\rmG_{t}, t | \rva, \rvz) = [\hat{\rvv}^{\text{len}}_{t}, \hat{\rvv}^{\text{ang}}_{t}, \hat{\rmV}^{\text{pos}}_{t}]$. 
Lastly, to sample the denoising timestep for training, we draw a sample from, our novel, \textit{lifted logit-normal} distribution, which is a Bernoulli mixture of the logit-normal distribution and the uniform distribution (consult Appendix \ref{app:lifted} for more information).

\vspace{-5pt}
\subsection{Training}
\vspace{-5pt}
The goal of training CliqueFlowmer is to learn a map $p_{\theta}(\rvz|\rva, \rmG)$ of a material $\mathsf{M}=[\rva, \rmG]$ to a fixed-dimensional continuous vector $\rvz$, a decomposable approximator $f_{\theta}(\rvz)$ of the target property, and the inverse $p_{\theta}(\rva, \rmG|\rvz)=p_{\theta}(\rva|\rvz)p_{\theta}(\rmG|\rva, \rvz)$ of the encoding map.
Thus, to train CliqueFlowmer, we must combine training of its four integrated submodules.
At the nucleus of this fusion there is sampling the latent representation $\rvz\sim p_{\theta}(\rvz|\rva, \rmG)$---given this vector, the atom type decoder learns to maximize the conditional next-token log-likelihood 
\begin{align}
    -\mathcal{L}_{\text{atom}}(\rva, \rvz) = \sum_{i=0}^{N_{\text{atom}}}\log p_{\theta}(\ra_{i+1}|\rva_{0:i}, \rvz), \nonumber
    \vspace{-30pt}
\end{align}
where $\ra_{0}=\langle\operatorname{Start}\rangle$ and $\ra_{N_{\text{atom}}+1}=\langle\operatorname{Stop}\rangle$.
Meanwhile, the flow-matching model that uses cross-attention to condition on the clique form $\rmZ$ of the latent, samples a noise geometry $\rmG_{0}$ for a cell with $N_{\text{atom}}$ atoms, interpolates it with the ground-truth geometry at a random timestep, $\rmG_t=(1-t)\cdot\rmG_0 + t\cdot \rmG$, and minimizes the difference between the velocity network's prediction and the path between geometries,
\begin{align}
    \mathcal{L}_{\text{len}}(\rva, \rvz&, \rmG, \rmG_{0}) = \big( (a-a_0, b-b_0, c-c_0) - \hat{\rvv}^{\text{len}}_t\big)^2 \nonumber\\
    \mathcal{L}_{\text{ang}}(\rva, \rvz,& \rmG, \rmG_{0})= \big((\alpha-\alpha_0, \beta-\beta_0, \gamma-\gamma_0) - \hat{\rvv}^{\text{ang}}_t\big)^2 \nonumber \\
    \mathcal{L}_{\text{pos}}(\rva&, \rvz, \rmG, \rmG_{0}) = \sum_{i=1}^{N_{\text{atom}}}\big((\rvx^i - \rvx^i_{0}) - \hat{\rvv}^{\text{pos}, i}_t\big)^2. \nonumber
\end{align}
While training the flow denoiser, we mask the latent $\rvz$ and replace it with a random normal noise, $\epsilon_{\rvz}\sim \mathcal{N}(\mathbf{0}_{d_{\rvz}}, \mathbf{I}_{d_{\rvz}})$ with probability $p_{\text{lat}}=0.1$.
We decompose the flow loss into distinct components to ensure that the unit cell (lengths and angles) receives equal weight for every material in the batch. 
For clarity, we define the sum of flow losses as
\begin{align}
    \mathcal{L}_{\text{flow}}&(\rva, \rvz, \rmG, \rmG_0)  = \mathcal{L}_{\text{len}}(\rva, \rvz, \rmG, \rmG_{0})
    + \mathcal{L}_{\text{ang}}(\rva, \rvz, \rmG, \rmG_{0})
    + \tau_{\text{pos}}\cdot\mathcal{L}_{\text{pos}}(\rva, \rvz, \rmG, \rmG_{0}),
    \nonumber
\end{align}
where $\tau_{\text{pos}}$ is a positive scalar.
Further, the latent representation, in its clique form, is also tasked with minimizing the prediction error of the target property and the clique-based KL-divergence is computed for the parameters of the normal distribution of $\rvz$,
\begin{align}
    \label{eq:kl}
    \mathcal{L}_{\text{pred}}(\mathsf{M}, \rvz) = \big(f_{\theta}(\rvz) - f(\mathsf{M})\big)^2 \quad \& \quad  
    \mathcal{L}_{\text{lat}}(\mu_{\rvz}, \sigma_{\rvz}) = \text{KL}\big(\mathcal{N}(\mu_{\rvz, c}, \sigma^{2}_{\rvz, c}),\ \mathcal{N}(\mathbf{0}_{d_{\text{clique}}}, \mathbf{I}_{d_{\text{clique}}})\big),
\end{align}
where the clique $c$ is drawn from the uniform distribution $\mathcal{U}[N_{\text{clique}}]$ \citep{kuba2024cliqueformer}. To summarize, the total expected loss takes form
\begin{align}
    \mathcal{L}(\theta) = \E_{\mathsf{M}\sim\mathcal{D}, \rvz\sim p_{\theta}, \rmG_{0}\sim p_{0}}\big[\mathcal{L}_{\text{atom}}(\rva, \rvz) 
    + \mathcal{L}_{\text{flow}}(\rva, \rvz, \rmG, \rmG_0) 
    + \tau_{\text{pred}} \mathcal{L}_{\text{pred}}(\mathsf{M}, \rvz) 
    + \beta  \mathcal{L}_{\text{lat}}(\mu_{\rvz}, \sigma_{\rvz})\big],
    \nonumber
\end{align}
where $\beta$ and $\tau_{\text{pred}}$ are positive scalars which we warm up linearly one after another.
While this loss consists of multiple terms, the architectural building blocks introduced in the previous subsections allow for their minimization and, as a result, learning representations that enable MBO for CMD.

\subsection{Computational Materials Discovery with CliqueFlowmer}
\label{sec:CMD-clique}
To optimize new materials with our model, we conduct a form of gradient-based search in the representation space.
We initialize this process by encoding a sample of existing materials with the encoder, $\rvz=\text{Enc}_{\theta}(\rva, \rmG)$.
Then, we optimize the prediction of the target property,
\begin{align}
    \label{eq:clique-mbo}
    \vspace{-10pt}
    \rvz_{\star} = \argmin_{\rvz^{\text{new}}}f_{\theta}(\rvz^{\text{new}}).
\end{align}
The choice of the minimization algorithm plays a significant role. 
While we have found that exact back-propagation of Equation (\ref{eq:clique-mbo}) is prone to adversarial exploitation of the model,
back-propagation-free evolution strategies \citep[ES]{salimans2017evolution} were very effective (see Section \ref{subsec:opt}).
That is, at every iteration, we draw $N_{\text{pert}}$ noise vectors $\epsilon$ from the standard normal distribution, $\epsilon \sim \mathcal{N}(\mathbf{0}_{d_{\rvz}}, \mathbf{I}_{d_{\rvz}})$, and compute the predicted values $f_{\theta}(\rvz + \sigma \epsilon)$ of the perturbations of $\rvz$ at scale $\sigma$. 
Then, we rank these perturbations, $\epsilon^1, \dots, \epsilon^{N_{\text{pert}}}$, based on the predicted values of $f$ (so that the smallest value gets the lowest rank). 
We standardize the ranks to have zero mean and unit variance and compute the ES gradient as a linear combination of the perturbations weighted by the ranks.
Additionally, we use antithetic sampling (see Appendix \ref{app:es} for more information about ES).
Then, we take a gradient step with AdamW optimizer \citep{loshchilov2017fixing}. 
The weight decay step of the optimizer, $\rvz \mapsto (1-\eta\lambda)\rvz$, where $\eta$ is the learning rate, is crucial since it brings the optimized latent variable closer to the origin.
In effect, this step increases the likelihood of the latent under the Gaussian prior that CliqueFlowmer was trained with (see Equation (\ref{eq:kl})), which mitigates the problem of going out of distribution with $\rvz$.

Given an optimized latent representation $\rvz_{\star}$, we modulate it like in Equation (\ref{eq:atom-type-mod}), and pass to the atom-type decoder, from which we sample the atom-type sequence using beam search (see Appendix \ref{app:beam}) with beam width $N_{\text{beam}}=10$,
\begin{align}
    \rva_{\star} \sim \operatorname{BeamSearch}(T^{\text{dec}}_{\theta}, \rvz_{\star}^{\text{mod}}).\nonumber
\end{align}
Next, we consider the clique form of the latent representation, $\rmZ_{\star}=\texttt{chain}(\rvz_{\star}, d_{\text{clique}}, d_{\text{knot}})$, and pass it to our continuous normalizing flow model, from which we decode the material geometry.
For this end, we use the \emph{Euler method} with \emph{classifier-free guidance} \citep[CFG]{ho2022classifier}, in which the effective velocity used in the ODE is
\begin{align}
    V^{\omega}_{\theta}(\rmG_t, t |\rva_{\star}, \rvz_{\star}) = (1 + \omega) \cdot V_{\theta}(\rmG_t, t|\rva_{\star}, \rvz_{\star}) - \omega \cdot V_{\theta}(\rmG_t, t | \rva_{\star}, \epsilon_{\rvz}),\nonumber
\end{align}
where $\epsilon_{\rvz}$ is standard normal noise. In our experiments, we used $\omega=2$ (see Appendix \ref{sec:appendix-cfg} for ablations).
We provide a discussion on the time complexity of this procedure in Appendix \ref{app:inference}.

\begin{table*}[t]
    \centering
    {\small
    \CFwidetable{
        \toprule[0.8pt]
        \rowcolor{CFheaderfill}
        \CFheading{Metric} & \CFheading{CrystalFormer} & \CFheading{DiffCSP} & \CFheading{DiffCSP++} & \CFheading{MatterGen} & \CFheading{\textbf{CliqueFlowmer}} & \CFheading{\textbf{-Top}} \\
        \midrule[0.4pt]
        \rowcolor{CFmetricfill}
        \CFmetric{E$_{\text{form}}$ ($\downarrow$)} & \CFnumber{\textcolor{CFbodyink}{0.71}} & \CFnumber{\textcolor{CFbodyink}{0.59}} & \CFnumber{\textcolor{CFbodyink}{0.65}} & \CFnumber{\textcolor{CFbodyink}{0.60}} & \CFnumber{\CFsecond{\ensuremath{-}0.81}} & \CFnumber{\CFbest{\ensuremath{-}0.99}} \\
        Stable ($\uparrow$) & \CFnumber{\textcolor{CFbodyink}{14.1}} & \CFnumber{\CFsecond{19.2}} & \CFnumber{\CFbest{19.8}} & \CFnumber{\textcolor{CFbodyink}{18.5}} & \CFnumber{\textcolor{CFbodyink}{14.5}} & \CFnumber{\textcolor{CFbodyink}{13.6}} \\
        S.U.N. ($\uparrow$) & \CFnumber{\textcolor{CFbodyink}{12.8}} & \CFnumber{\CFbest{18.6}} & \CFnumber{\CFsecond{18.5}} & \CFnumber{\textcolor{CFbodyink}{17.6}} & \CFnumber{\textcolor{CFbodyink}{14.4}} & \CFnumber{\textcolor{CFbodyink}{13.4}} \\
        E$_{\text{form}}$ / S.U.N. ($\downarrow$) & \CFnumber{\textcolor{CFbodyink}{1.06}} & \CFnumber{\textcolor{CFbodyink}{1.02}} & \CFnumber{\textcolor{CFbodyink}{1.10}} & \CFnumber{\textcolor{CFbodyink}{0.98}} & \CFnumber{\CFsecond{\ensuremath{-}0.65}} & \CFnumber{\CFbest{\ensuremath{-}1.06}} \\
        \midrule[0.35pt]
        \rowcolor{CFmetricfill}
        \CFmetric{Band Gap ($\downarrow$)} & \CFnumber{\textcolor{CFbodyink}{0.52}} & \CFnumber{\textcolor{CFbodyink}{0.63}} & \CFnumber{\textcolor{CFbodyink}{0.48}} & \CFnumber{\textcolor{CFbodyink}{0.57}} & \CFnumber{\CFbest{0.03}} & \CFnumber{\CFsecond{0.07}} \\
        Stable ($\uparrow$) & \CFnumber{\textcolor{CFbodyink}{14.1}} & \CFnumber{\textcolor{CFbodyink}{19.2}} & \CFnumber{\textcolor{CFbodyink}{19.8}} & \CFnumber{\textcolor{CFbodyink}{18.5}} & \CFnumber{\CFsecond{61.3}} & \CFnumber{\CFbest{69.4}} \\
        S.U.N. ($\uparrow$) & \CFnumber{\textcolor{CFbodyink}{12.8}} & \CFnumber{\textcolor{CFbodyink}{18.6}} & \CFnumber{\textcolor{CFbodyink}{18.5}} & \CFnumber{\textcolor{CFbodyink}{17.6}} & \CFnumber{\CFsecond{61.3}} & \CFnumber{\CFbest{69.4}} \\
        Band Gap / S.U.N. ($\downarrow$) & \CFnumber{\textcolor{CFbodyink}{0.37}} & \CFnumber{\textcolor{CFbodyink}{0.55}} & \CFnumber{\textcolor{CFbodyink}{0.23}} & \CFnumber{\textcolor{CFbodyink}{0.40}} & \CFnumber{\CFbest{0.02}} & \CFnumber{\CFsecond{0.05}} \\
        \bottomrule[0.65pt]
    }
    \caption{\small
    Target properties and stability metrics across generative baselines and CliqueFlowmer. Formation energies are in eV/atom, band gaps in eV, and Stable and S.U.N. rates in \%. ``Property / S.U.N.'' denotes the mean property among S.U.N. structures, retaining the property's units.
    Row-wise best results are \CFbest{bold teal}; second-best are \CFsecond{underlined blue}.
    Among other results, \textbf{CliqueFlowmer} and \textbf{CliqueFlowmer-Top} reduce average formation energy from 0.46 in the initial samples to -0.81 and -0.99, respectively, and reduce band gap from 0.57 to 0.03 and 0.07. 
    More metrics in Appendix \ref{app:additional}.}
        \label{tab:eform_sun_comparison}
    }
    \vspace{-5pt}
\end{table*}

\vspace{-5pt}
\section{Experiments}
\label{sec:exp}
\vspace{-5pt}
We tested CliqueFlowmer in two tasks for which we trained two separate models.
Both of them were trained on the MP-20 dataset, with the standard 60-20-20\% train-val-test split, that contains 45K example materials altogether \citep{xie2021crystal} for a total of $7\times 10^{5}$ gradient steps with batch size of 1024. 
In the first task, for the role of the target property, we chose the predicted formation energy per atom $E_{\text{form}}(\mathsf{M})$ (eV/atom) from M3GNet \citep{chen2022universal}, and for the second one, the band gap $\Delta_{\text{band}}(\mathsf{M})$ (eV) predicted by MEGNet \citep{chen2019graph}.
Formation energy provides an accessible benchmark for property optimization, although minimizing it alone does not ensure thermodynamic stability. The band-gap optimization objective includes a formation-energy penalty (Appendix~\ref{sec:appendix-hparams}).

\begin{figure}
    \centering
    \hspace{-10pt}
    \includegraphics[width=\linewidth]{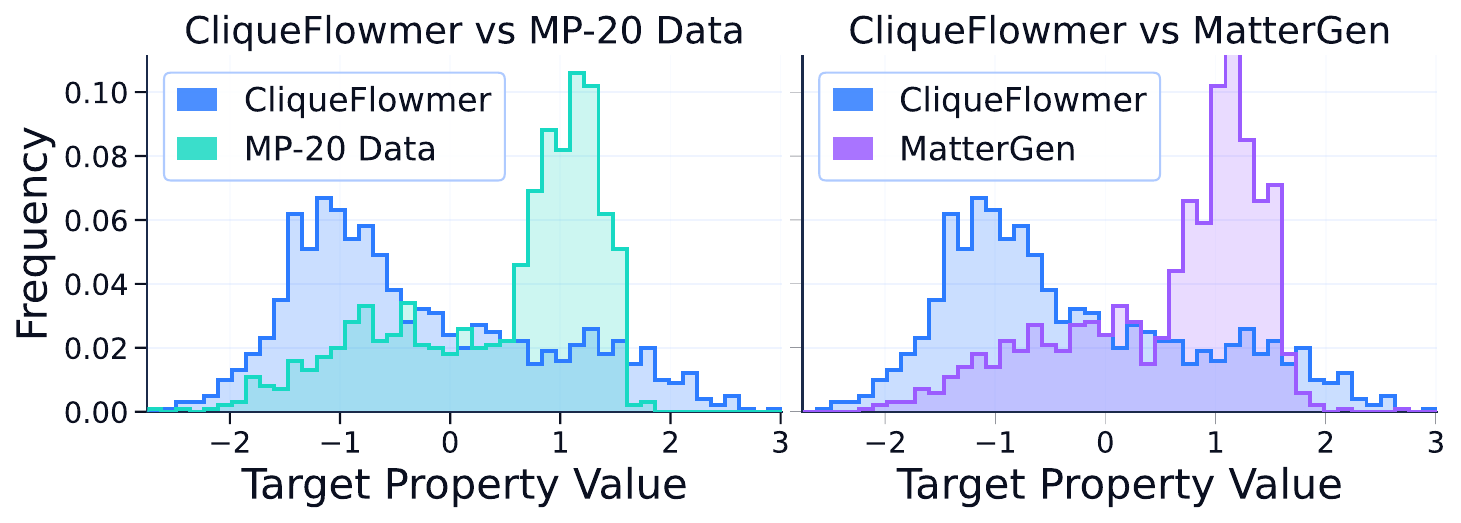}
    \caption{\small The distribution of formation energy per atom (horizontal axis, eV/atom) among discovered materials. We compare CliqueFlowmer (blue) to materials from MP-20 dataset (green, \emph{left}) and to those generated by MatterGen (purple, \emph{right}). 
    CliqueFlowmer significantly shifts the distribution of the target property to its lower values.
    For more histograms, see Appendix \ref{appendix:add-fig}.}
    \label{fig:histograms}
    \vspace{-10pt}
\end{figure}

\vspace{-5pt}
\begin{figure*}[t]
    \centering
    \captionsetup[subfigure]{justification=centering}
    \begin{subfigure}[b]{0.19\textwidth}
        \centering
        \includegraphics[width=\linewidth]{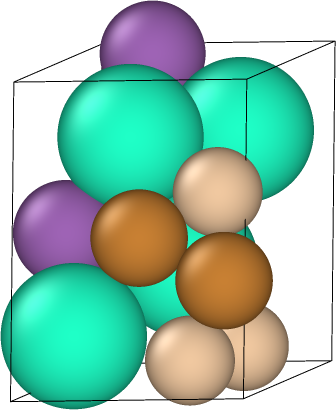}
        \caption{Dy$_2$CuSi$_2$Sb\\0.06\,eV}
    \end{subfigure}\hfill
    \begin{subfigure}[b]{0.19\textwidth}
        \centering
        \includegraphics[width=\linewidth]{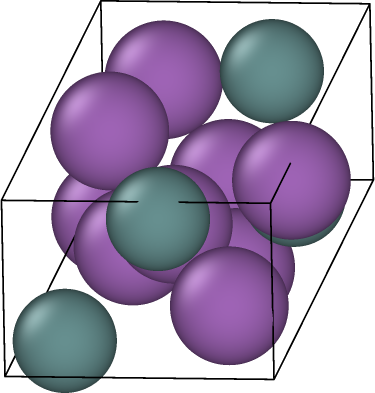}
        \caption{Ge$_4$Sb$_9$\\0.01\,eV}
    \end{subfigure}\hfill   
    \begin{subfigure}[b]{0.19\textwidth}
        \centering
        \includegraphics[width=\linewidth]{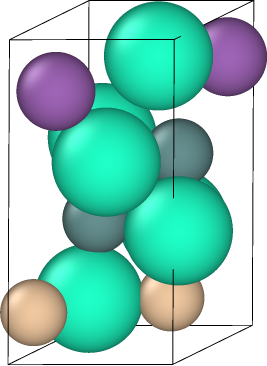}
        \caption{Dy$_3$SiGeSb\\0.14\,eV}
    \end{subfigure}\hfill
    \begin{subfigure}[b]{0.19\textwidth}
        \centering
        \includegraphics[width=\linewidth]{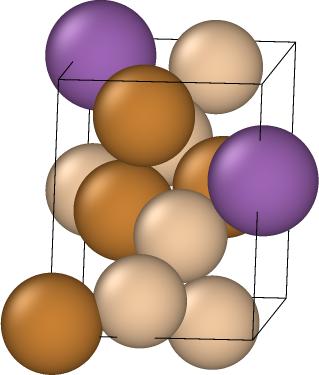}
        \caption{Cu$_2$Si$_3$Sb\\0.02\,eV}
    \end{subfigure}\hfill
    \begin{subfigure}[b]{0.19\textwidth}
        \centering
        \includegraphics[width=\linewidth]{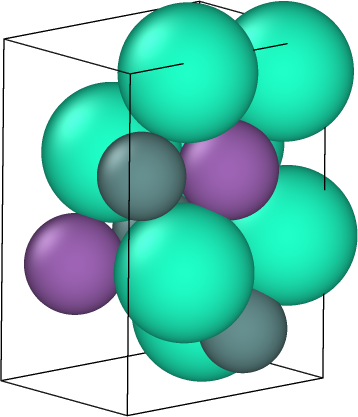}
        \caption{Dy$_7$(Ge$_2$Sb)$_2$\\0.03\,eV}
    \end{subfigure}\hfill
    \caption{Examples of crystals optimized by CliqueFlowmer for band-gap minimization. Each panel shows a unit cell, its composition, and its reported band gap in eV.}
    \label{fig:vesta-materials}
\end{figure*}

\subsection{MBO With CliqueFlowmer}
\vspace{-5pt}
To test CliqueFlowmer's ability to conduct CMD through MBO, we use the following protocol. 
We sample $N$ existing materials and encode them with the CliqueFlowmer encoder. Then, we optimize the materials' representations, using $T=2000$ ES-gradient steps. 
The converged representations are then decoded with the CliqueFlowmer decoder, thus rendering $N$ new materials (see Figure~\ref{fig:vesta-materials} for examples).
Additionally, one can filter out the most promising candidates prior to their structure-level oracle evaluation.
Namely, once optimization converges, one can predict the target property with the prediction head, $f_{\theta}(\rvz)$, and select top-$k\%$ solutions with the best values. 
The attractiveness of this approach stems from the computational savings it offers---having run the cheap optimization stage in the latent space, one can reject the majority of suboptimal latents before passing the most promising ones to the expensive denoising and relaxation stages (Appendix \ref{app:inference} discusses the wall-clock time complexity of our method).
We refer to this approach \emph{CliqueFlowmer-Top}, set $k=10$ in our experiments, and consider it the final method of our paper.

We measure the target property value of the $N=10^4$ original materials and the value of the newly-proposed ones. 
Additionally, to examine if our method produces both interesting and stable materials, we calculate the S.U.N. rate (stable, unique, novel) \citep{zeni2023mattergen} among the proposed solutions. 
Then, to study the optimization performance among the interesting materials, the target property is re-evaluated for examples that are classified as S.U.N. 
We compare our results to CrystalFormer \citep{cao2025space}, DiffCSP \citep{jiao2023crystal}, DiffCSP++ \citep{jiao2024space}, and MatterGen \citep{zeni2023mattergen}, 
whose generated structures we obtained from \citet{kazeev2025generatedcrystals}.
Due to computational limitations, we relax all the materials (including those from the baselines) approximately with M3GNet \citep{chen2022universal} and determine stability with energy above local hull\footnote{As a result, our Stable and S.U.N. rates overestimate the true numbers but they do so for all methods.}.
To compensate for the bias induced by these approximations, we use the \emph{strict} criterion for stability (see Appendix \ref{sec:appendix-CMD}).
We also evaluate selected candidates with VASP using Materials Project relaxation inputs and energy corrections; Appendix~\ref{appendix:dft} details this DFT protocol and its implications for stability estimates.

\begin{wrapfigure}{r}{0.45\textwidth} 
\centering 
\vspace{-10pt}
\includegraphics[width=0.95\linewidth]{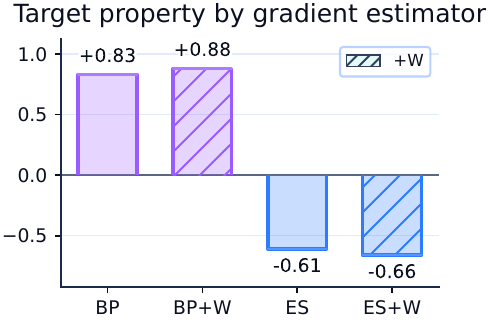} 
\caption{\small Comparison of gradient estimators---back-propagation (BP) vs. evolution strategies (ES), and weight decay. We plot the average change in formation energy per atom (eV/atom) due to optimization, over 100 materials.} 
\label{fig:optimization} 
\end{wrapfigure}

Results in Table \ref{tab:eform_sun_comparison} and Figure (\ref{fig:histograms}) show that CliqueFlowmer drastically reduces the value of the target property (M3GNet formation energy \& MEGNet band gap), significantly outperforming materials generated by the baselines. 
In the formation energy task, this quality is further amplified by 
CliqueFlowmer-Top, which reduces the objective value even further. 
We suspect that, in the band gap task, we do not observe this effect because the band gap is lower-bounded by zero, and thus the ordering of the top materials in the latent space may be driven by approximation errors.
Furthermore, while the purpose of CliqueFlowmer is property optimization, the structures proposed by the model outperform the baselines in terms of the S.U.N. rate in the band gap optimization task.
This may be surprising given that it is formation energy that is more tightly related to stability.
However, it is known in materials science that simply minimizing the formation energy may lead to low-energy regions of the materials space with many competing phases, resulting in fewer stable materials \citep{riebesell2025framework}.
Furthermore, in both tasks, the S.U.N. materials from our method outperformed those from the baselines, in terms of the target property, regardless of the S.U.N. rate.
Ultimately, the strong optimization performance (overall and /S.U.N) and high S.U.N. rates exceeded our expectations.
See Appendix \ref{app:opt} \& \ref{sec:appendix-hparams} for more information.

\begin{wrapfigure}{L}{0.45\textwidth} 
\centering 
\includegraphics[width=0.99\linewidth]{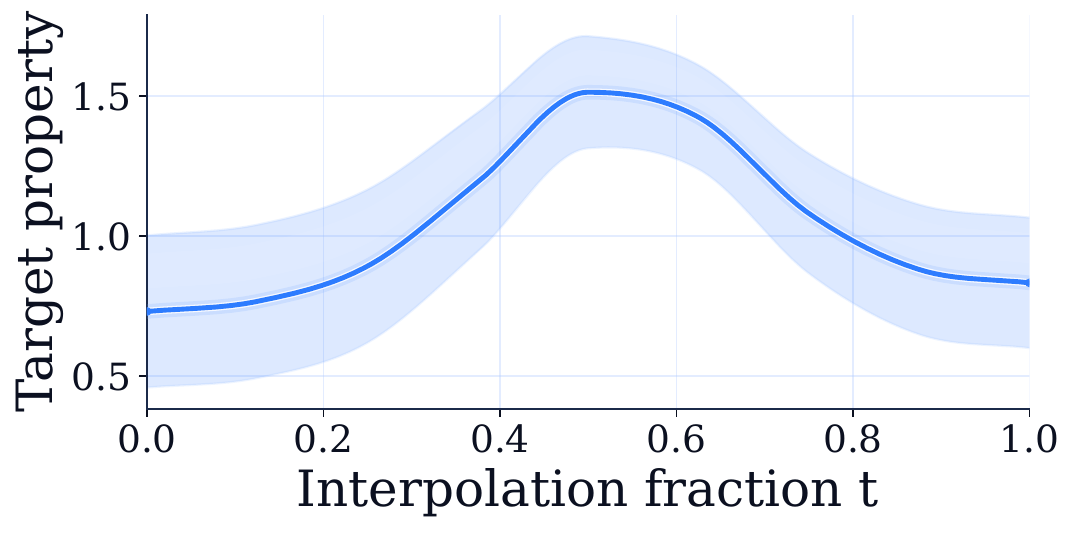} 
\caption{\small Average $f(\mathsf{M})$ (formation energy per atom, eV/atom) with standard error of the mean (SEM) over the course of linear interpolation. The curve was smoothed out with the Gaussian kernel. The value tends to be higher along the interpolation trajectory.} 
\label{fig:interpolation-f} 
\vspace{-5pt} 
\end{wrapfigure}
\subsection{Optimization Algorithm}
\label{subsec:opt}
\vspace{-5pt}
As described in Section \ref{sec:CMD-clique}, having trained CliqueFlowmer, we discover new materials by solving the optimization problem from Equation (\ref{eq:clique-mbo}).
In this paper, we utilized the back-propagation-free ES algorithm \citep{salimans2017evolution}. 
This section provides an empirical justification of this counterintuitive choice.
We compare the performance of the gradient derived by back-propagation (BP) to ES by optimizing $N=100$ structures, in CliqueFlowmer's latent space, for $T=1000$ steps.
In our experiments, both gradients were applied by Adam optimization algorithm \citep{kingma2014adam}, as well as by its decoupled weight decay variant \citep[BP+W, ES+W]{loshchilov2017fixing}.
Figure (\ref{fig:optimization}) shows that, instead of decreasing the minimized property, BP and BP+W have drastically increased it. 
Meanwhile, ES has successfully progressed towards the property minimization, and ES+W has done that most effectively.
Thus, our main experiments use ES+W.
More information in Appendix \ref{app:opt}.

\begin{figure*}[t]
\centering
\setlength{\tabcolsep}{3pt}
\newcommand{\cellcardheight}{0.68\linewidth}
\newcommand{\cellxshift}{-0.6mm}

\newcommand{\platerule}{0.55pt}
\newcommand{\platepad}{3pt}
\newcommand{\headgap}{1.5pt}   
\newcommand{\capgap}{-4pt}    
\newcommand{\rowgap}{6pt}      

\definecolor{plate}{RGB}{80,80,80}
\newcommand{\cell}[3]{
  \begin{minipage}[t]{0.18\textwidth}\centering
    \if\relax\detokenize{#1}\relax\else
      {\footnotesize\textbf{#1}}\par\vspace{0.4pt}%
      \vspace{\headgap}%
    \fi
    #2\par\vspace{\capgap}%
    {\footnotesize #3}%
  \end{minipage}%
}

\vspace{-5pt}
{\color{plate}\rule{\textwidth}{\platerule}}\par
\vspace{\platepad}
\noindent\begin{tabular}{@{}ccccc@{}}
\cell{$t=0.00$}{\includegraphics[width=\linewidth,height=\cellcardheight,keepaspectratio,trim=10 20 18 20,clip]{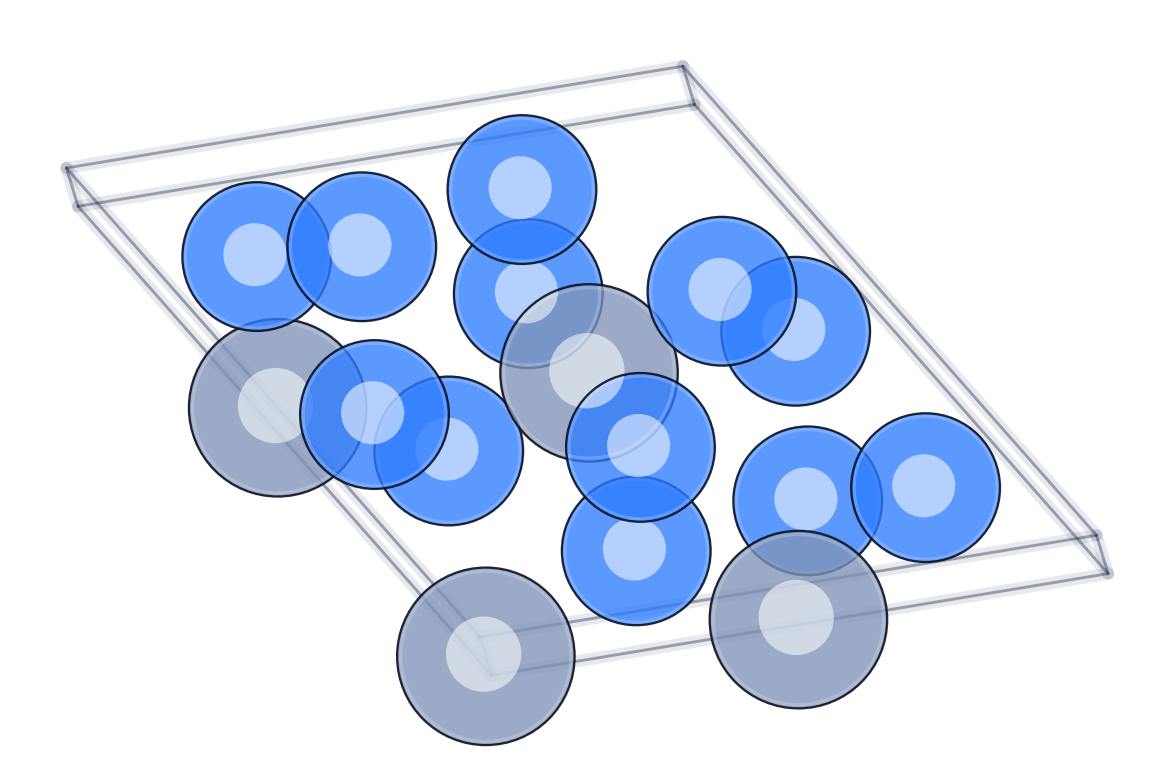}}{As$_3$Rh} &
\cell{$t=0.25$}{\includegraphics[width=\linewidth,height=\cellcardheight,keepaspectratio,trim=10 20 18 20,clip]{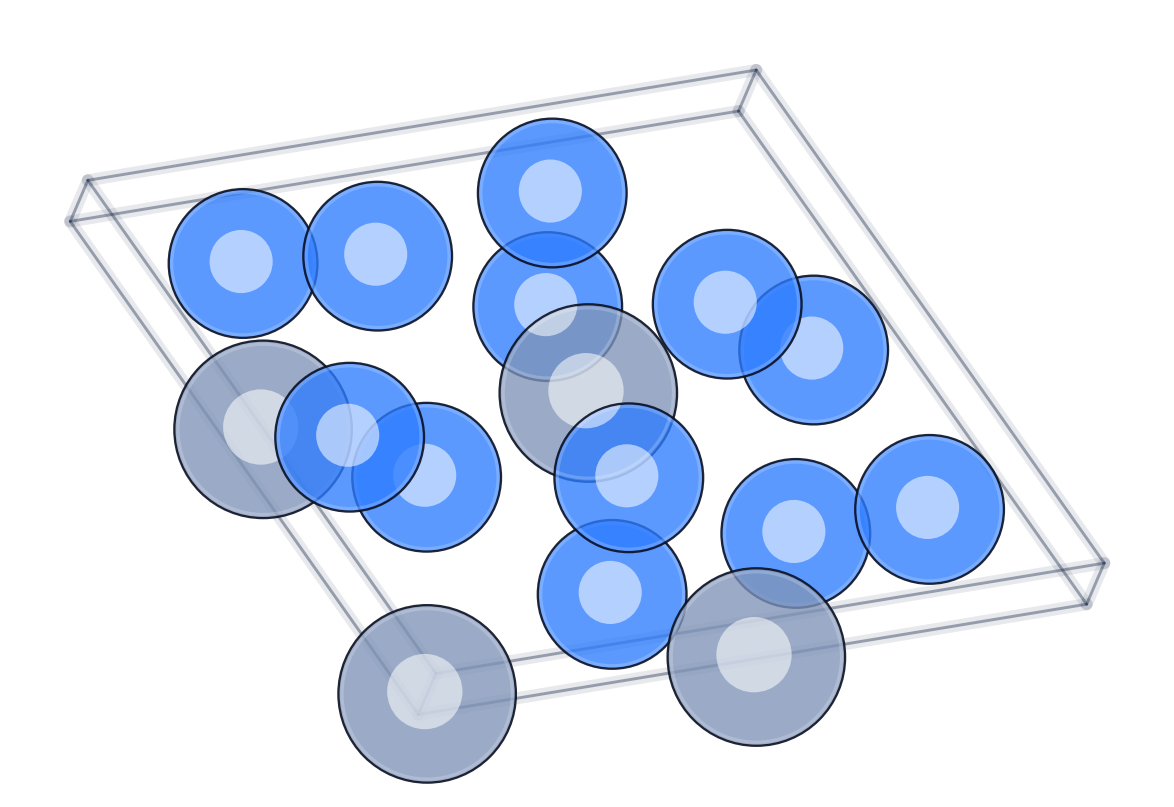}}{As$_3$Rh} &
\cell{$t=0.50$}{\includegraphics[width=\linewidth,height=\cellcardheight,keepaspectratio,trim=10 20 18 20,clip]{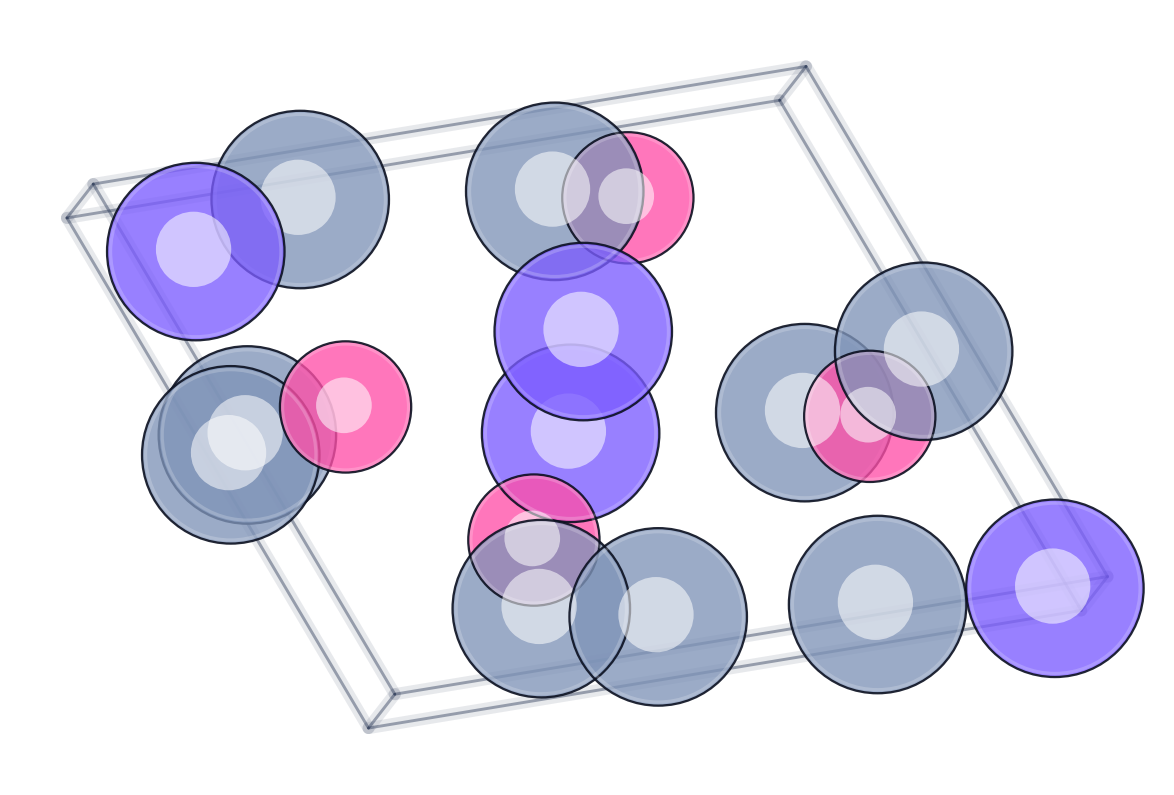}}{In$_4$Rh$_9$S$_4$} &
\cell{$t=0.75$}{\includegraphics[width=\linewidth,height=\cellcardheight,keepaspectratio,trim=10 20 18 20,clip]{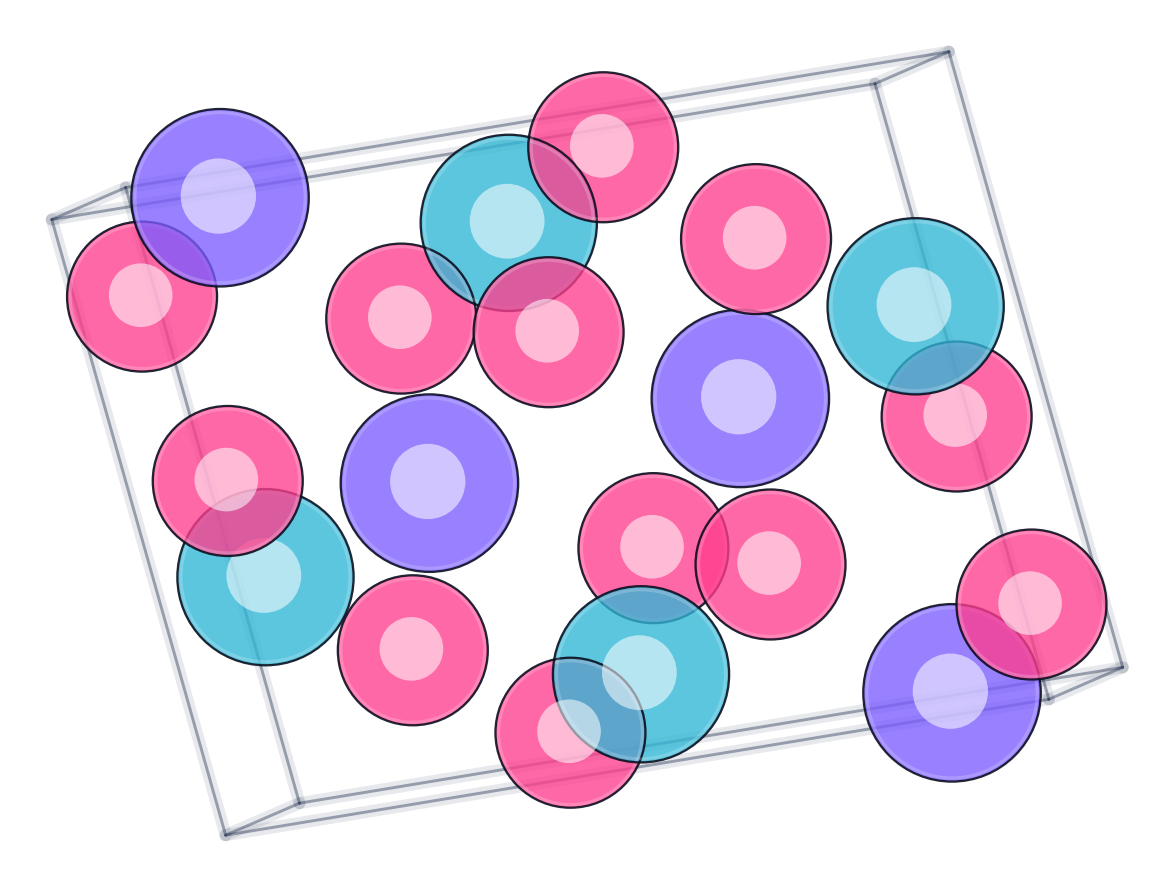}}{MgInBr$_3$} &
\cell{$t=1.00$}{\includegraphics[width=\linewidth,height=\cellcardheight,keepaspectratio,trim=10 20 18 20,clip]{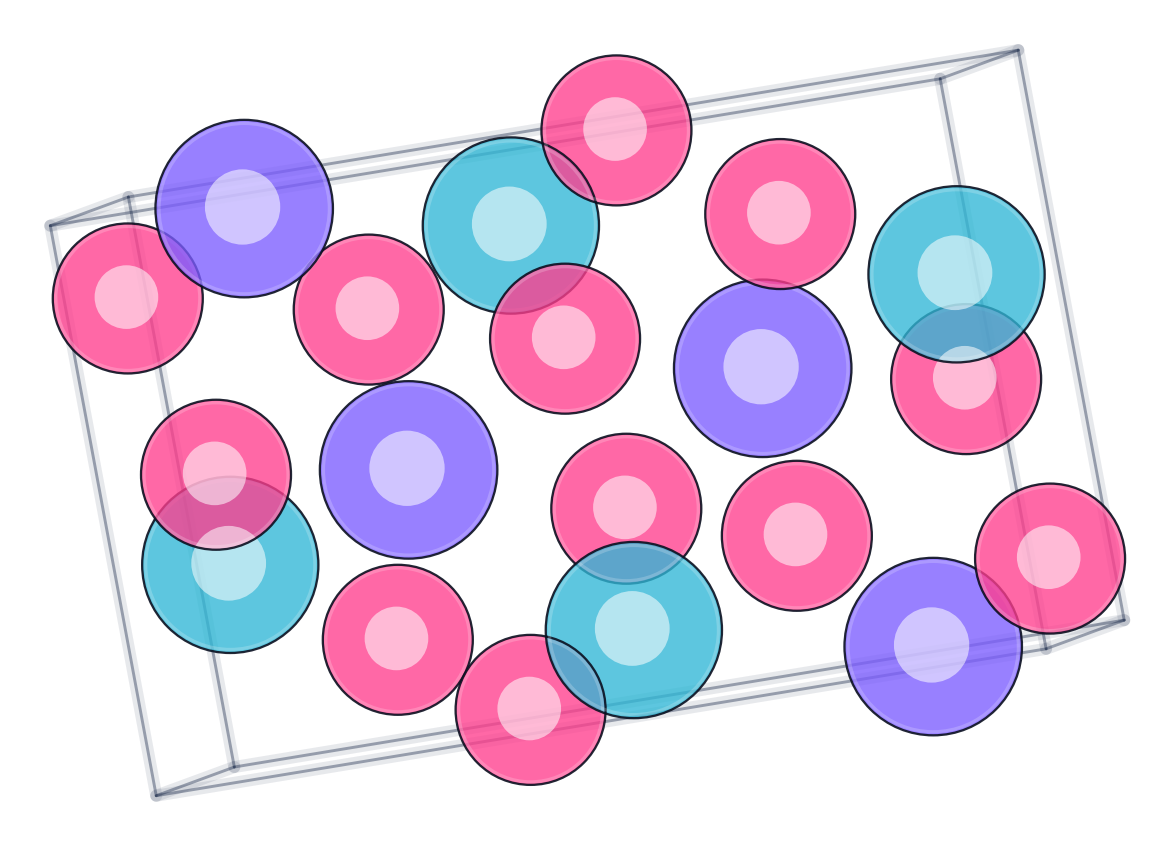}}{MgInBr$_3$} 
\\[\rowgap]
\cell{}{\includegraphics[width=\linewidth,height=\cellcardheight,keepaspectratio,trim=10 20 18 20,clip]{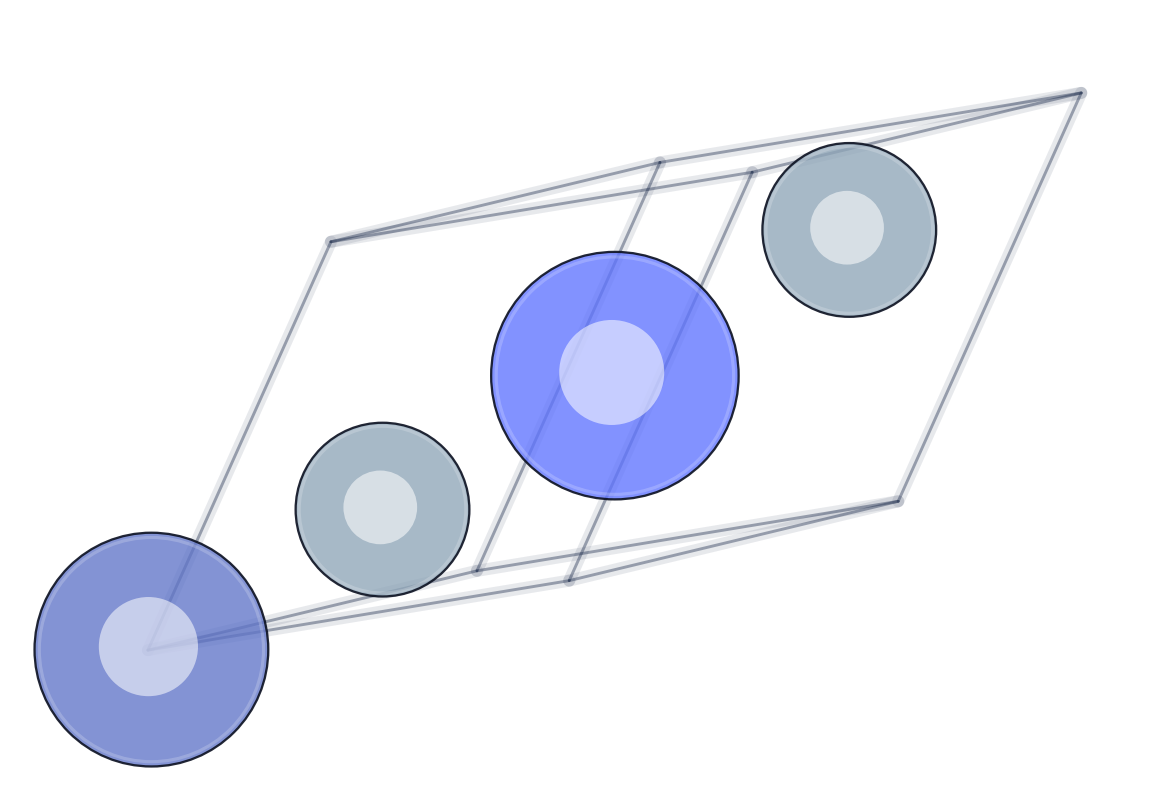}}{YbEuPd$_2$} &
\cell{}{\includegraphics[width=\linewidth,height=\cellcardheight,keepaspectratio,trim=10 20 18 20,clip]{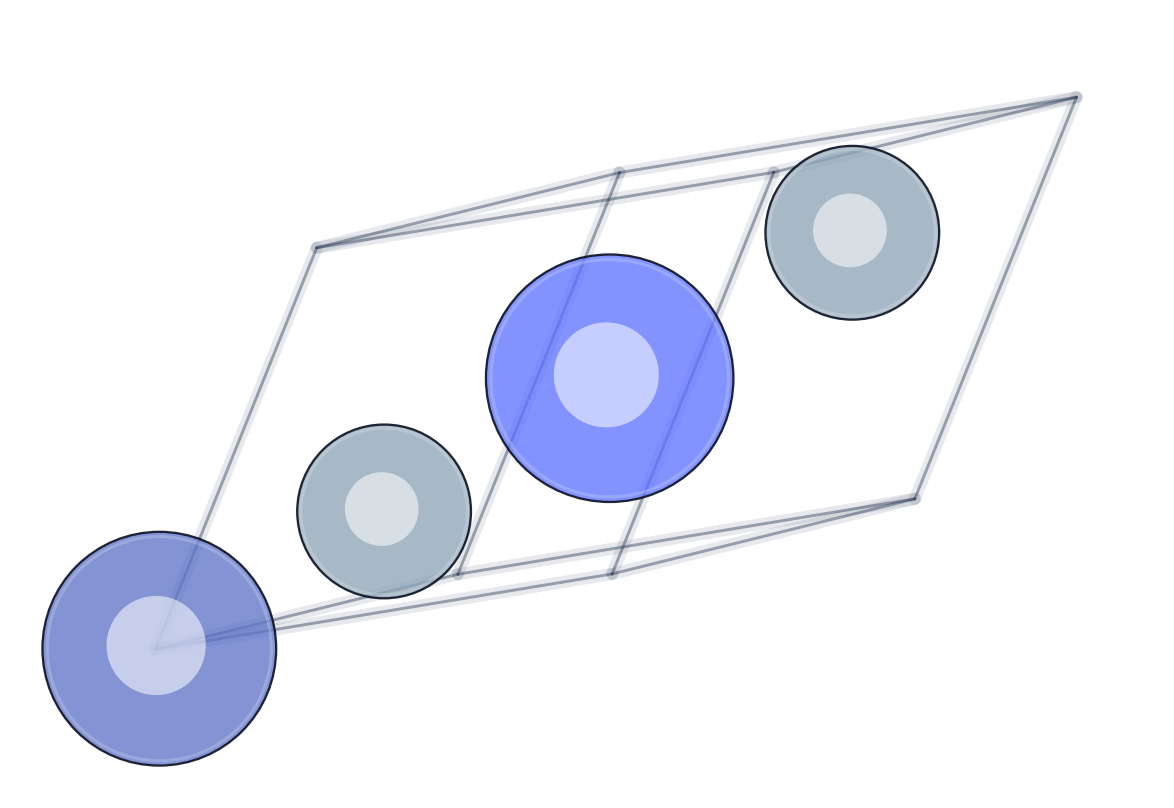}}{YbEuPd$_2$} &
\cell{}{\includegraphics[width=\linewidth,height=\cellcardheight,keepaspectratio,trim=10 20 18 20,clip]{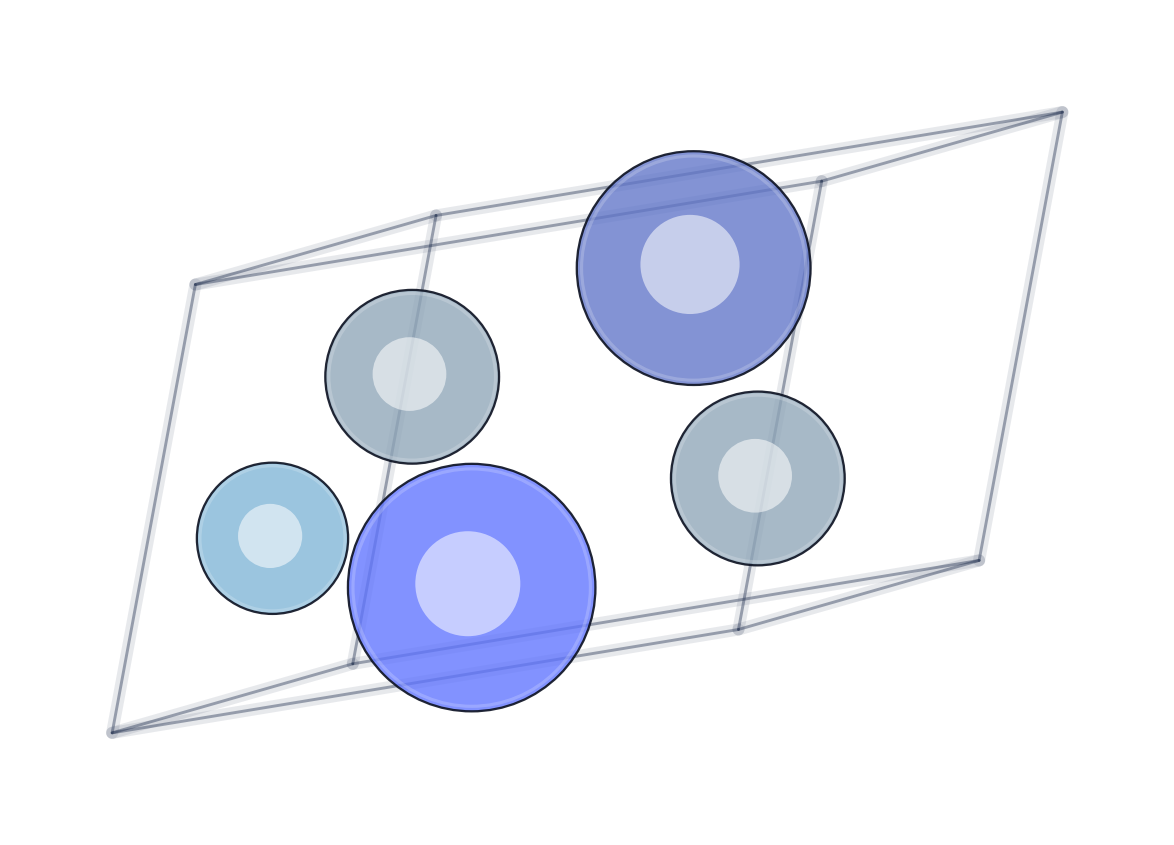}}{YbEuAlPd$_2$} &
\cell{}{\includegraphics[width=\linewidth,height=\cellcardheight,keepaspectratio,trim=10 20 18 20,clip]{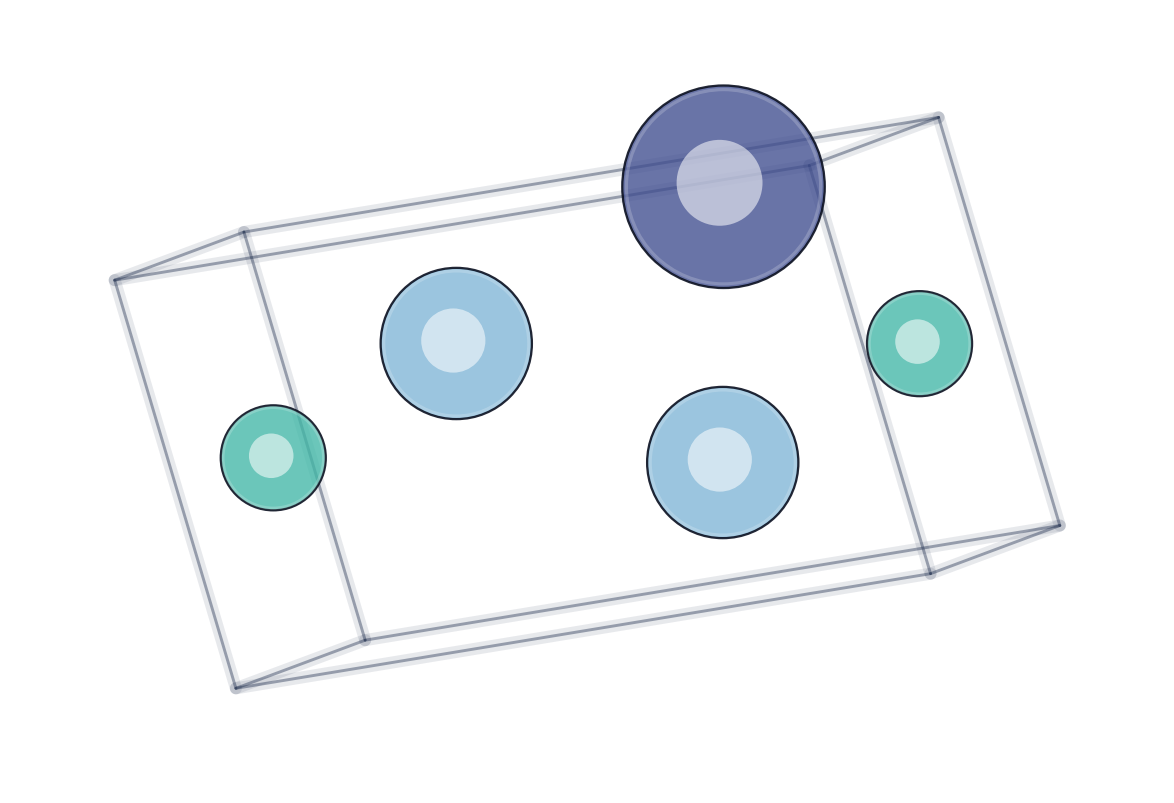}}{Al$_2$B$_2$W} &
\cell{}{\includegraphics[width=\linewidth,height=\cellcardheight,keepaspectratio,trim=10 20 18 20,clip]{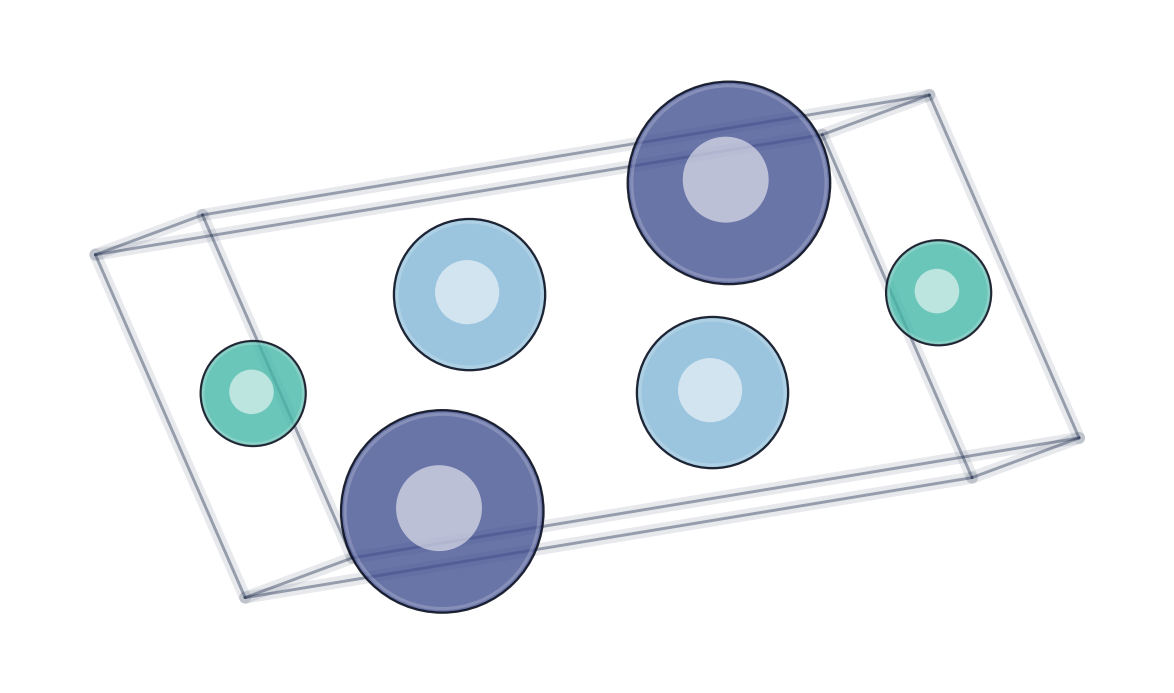}}{AlBW} 
\\[\rowgap]
\cell{}{\includegraphics[width=\linewidth,height=\cellcardheight,keepaspectratio,trim=10 20 18 20,clip]{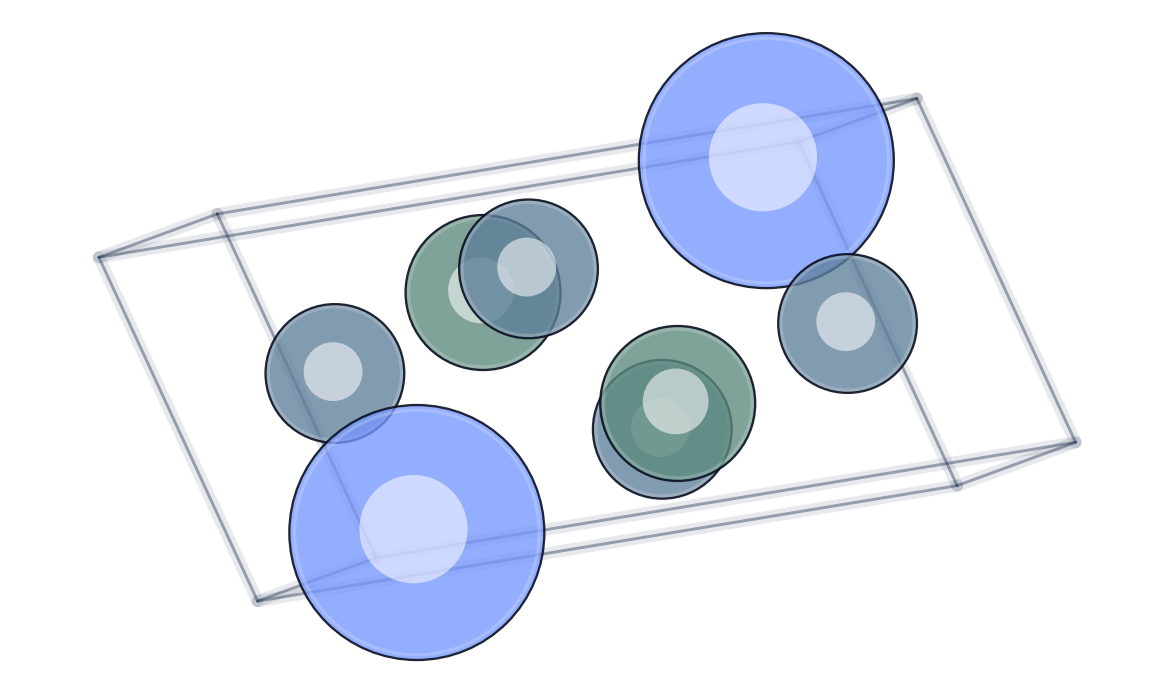}}{CeSi$_2$Ni} &
\cell{}{\includegraphics[width=\linewidth,height=\cellcardheight,keepaspectratio,trim=10 20 18 20,clip]{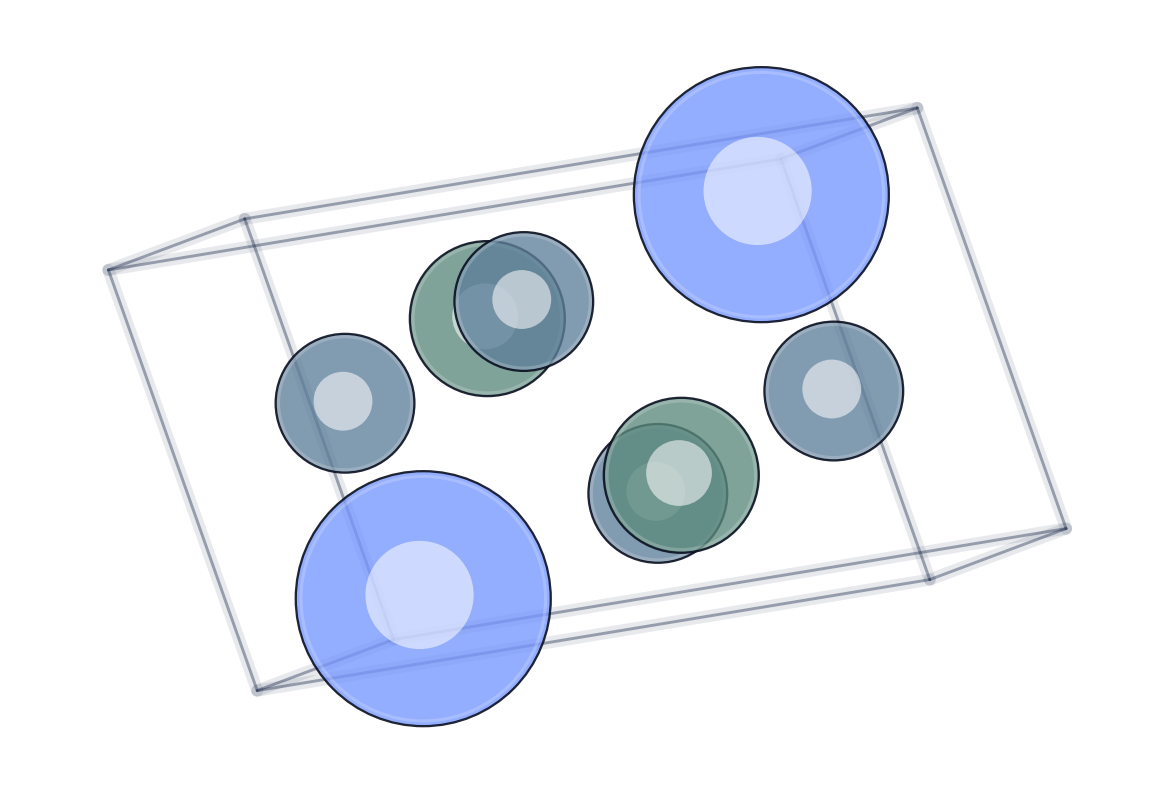}}{CeSi$_2$Ni} &
\cell{}{\includegraphics[width=\linewidth,height=\cellcardheight,keepaspectratio,trim=10 20 18 20,clip]{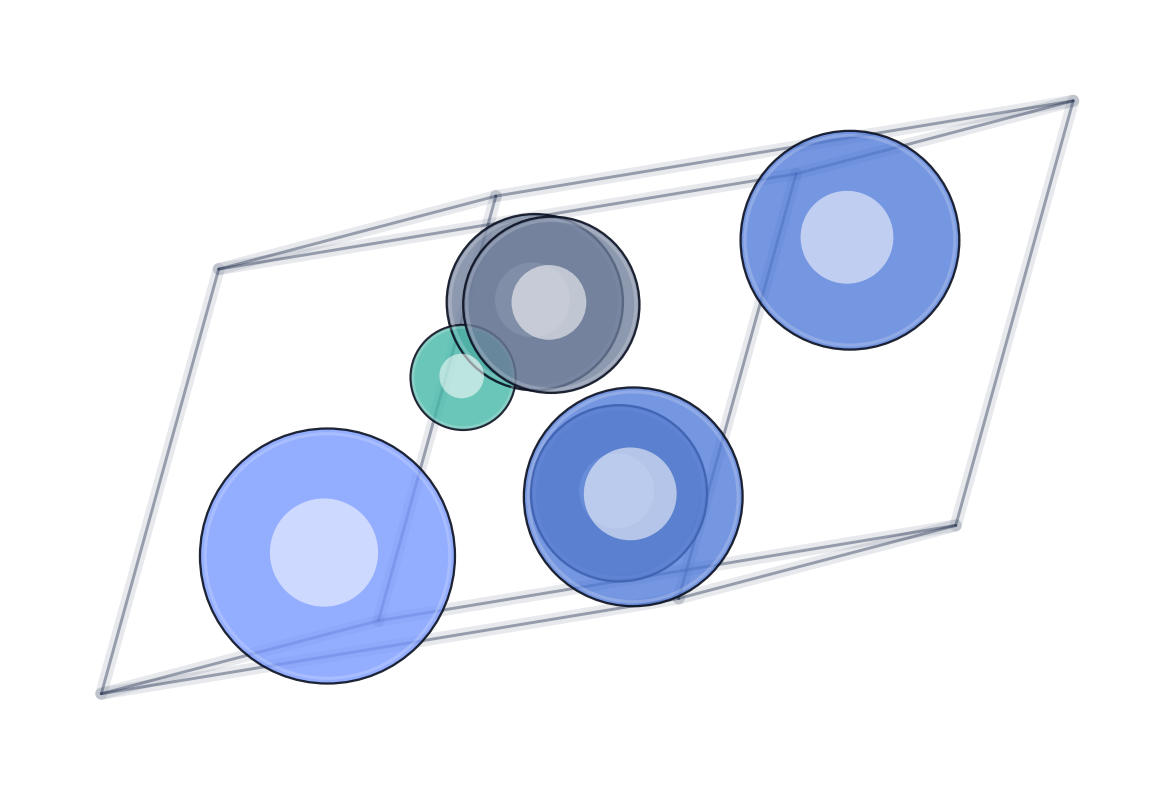}}{CeZr$_2$BIr$_3$} &
\cell{}{\includegraphics[width=\linewidth,height=\cellcardheight,keepaspectratio,trim=10 20 18 20,clip]{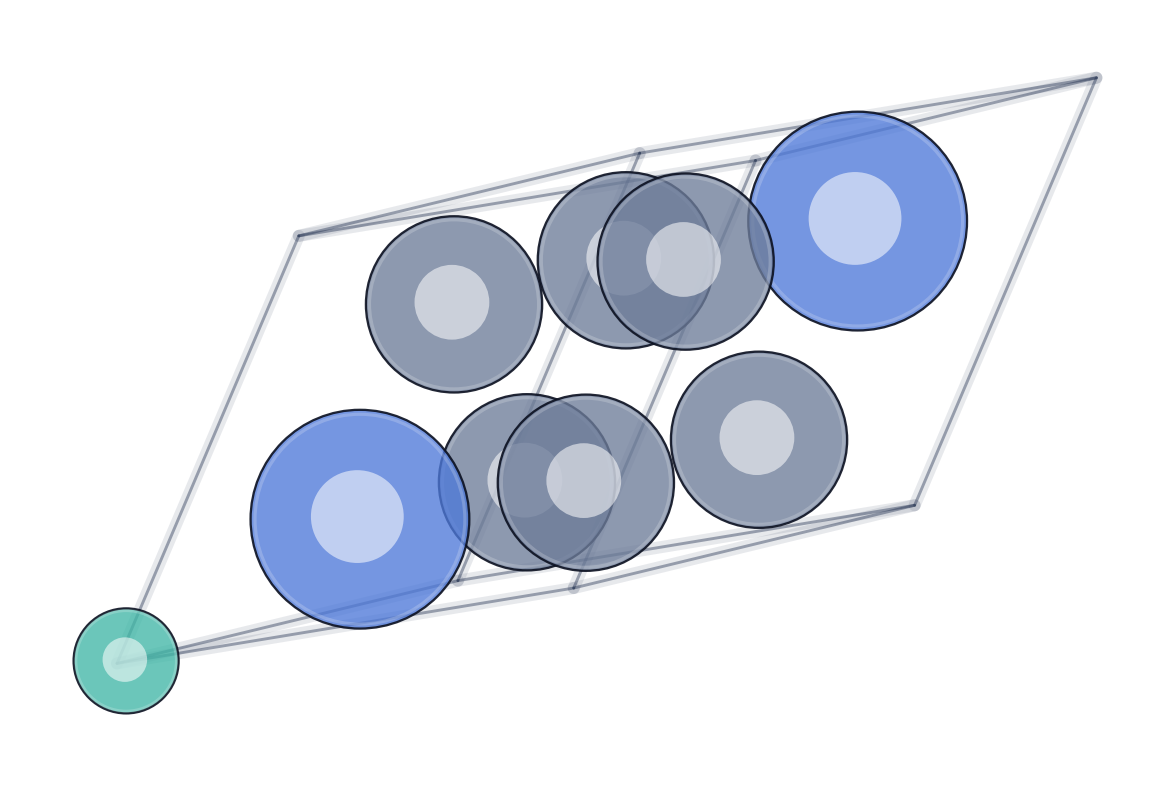}}{Zr$_2$BIr$_6$} &
\cell{}{\includegraphics[width=\linewidth,height=\cellcardheight,keepaspectratio,trim=10 20 18 20,clip]{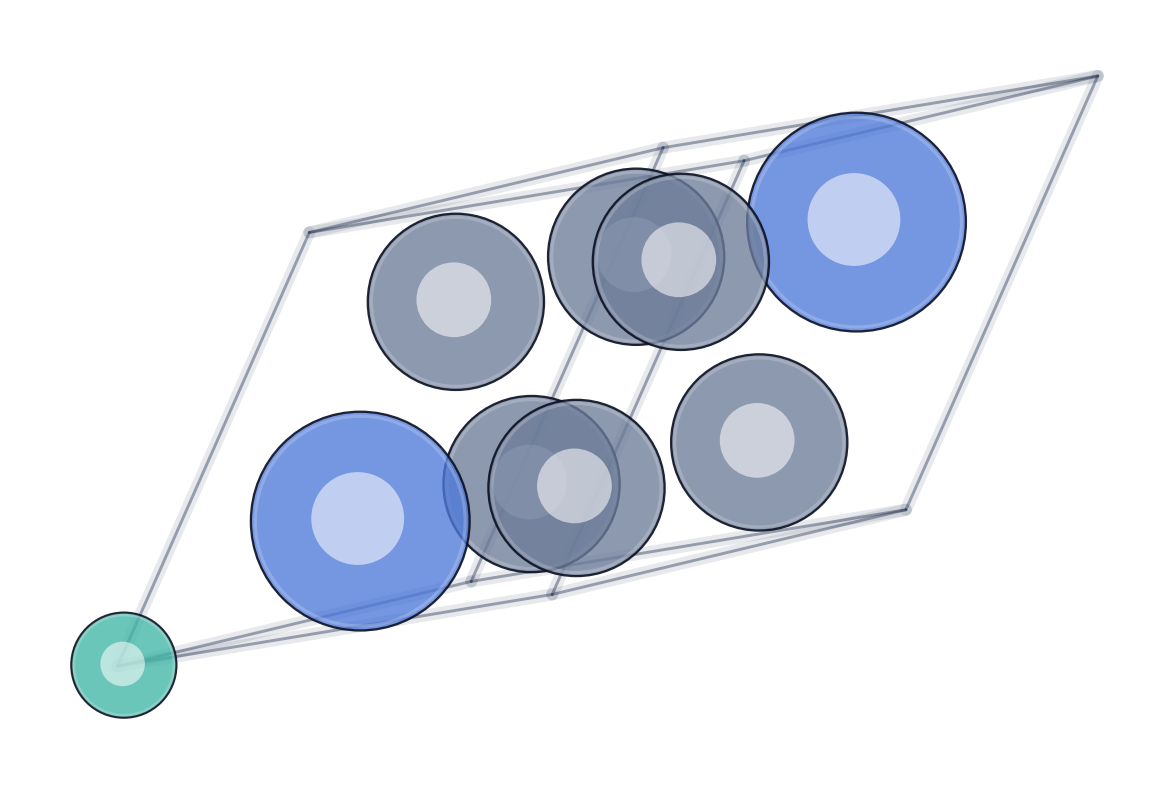}}{Zr$_2$BIr$_6$} 
\\[\rowgap]
\cell{}{\includegraphics[width=\linewidth,height=\cellcardheight,keepaspectratio,trim=10 20 18 20,clip]{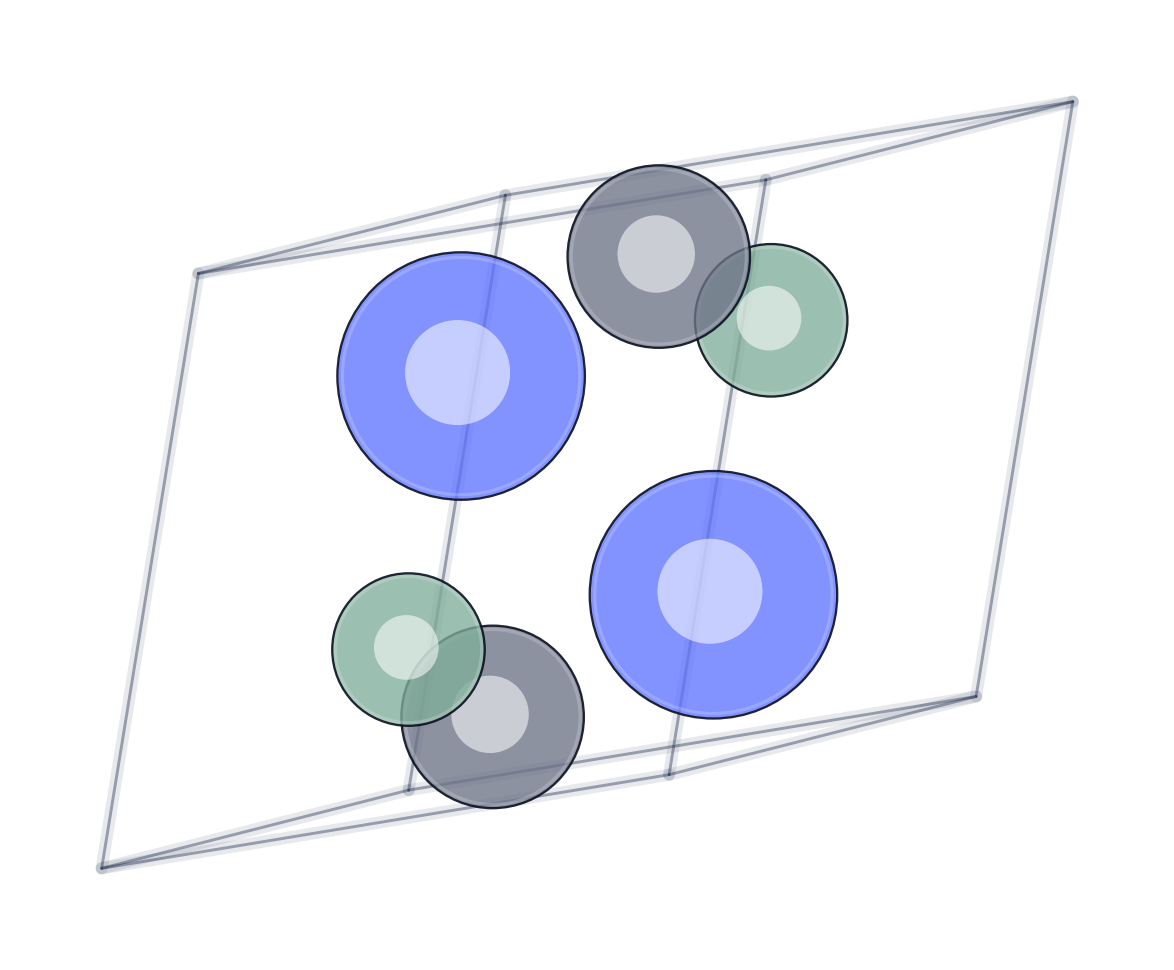}}{EuZnPb} &
\cell{}{\includegraphics[width=\linewidth,height=\cellcardheight,keepaspectratio,trim=10 20 18 20,clip]{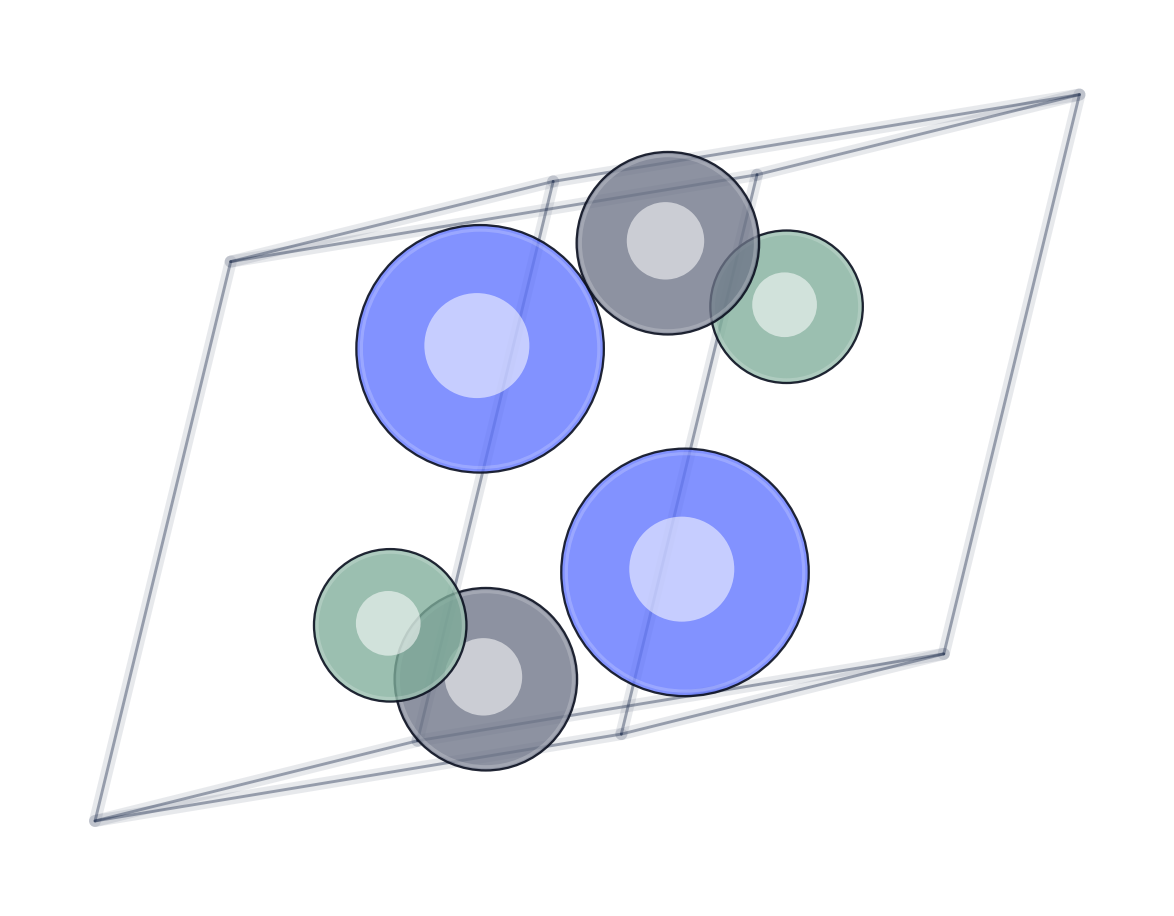}}{EuZnPb} &
\cell{}{\includegraphics[width=\linewidth,height=\cellcardheight,keepaspectratio,trim=10 20 18 20,clip]{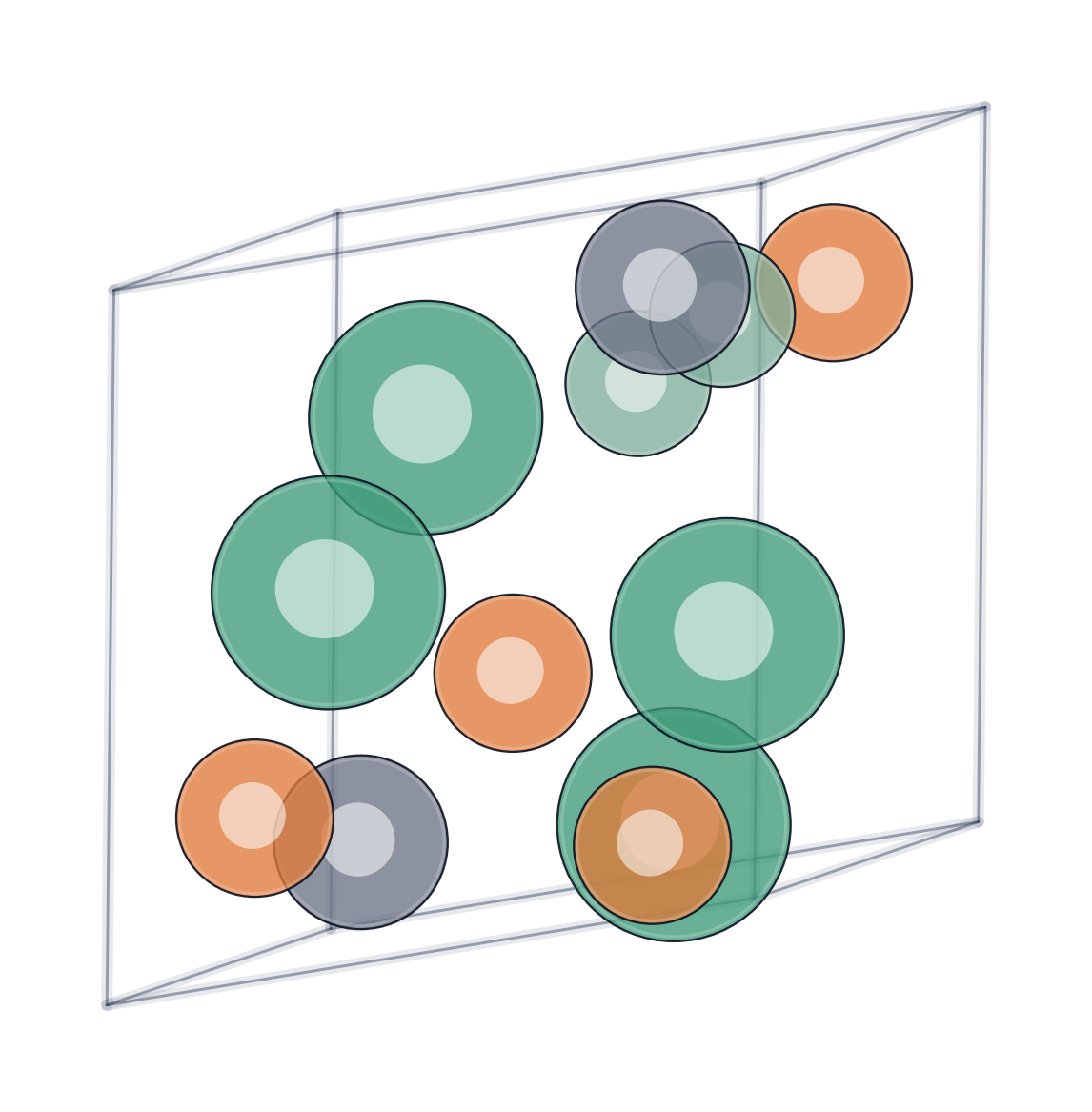}}{U$_2$ZnCu$_2$Pb} &
\cell{}{\includegraphics[width=\linewidth,height=\cellcardheight,keepaspectratio,trim=10 20 18 20,clip]{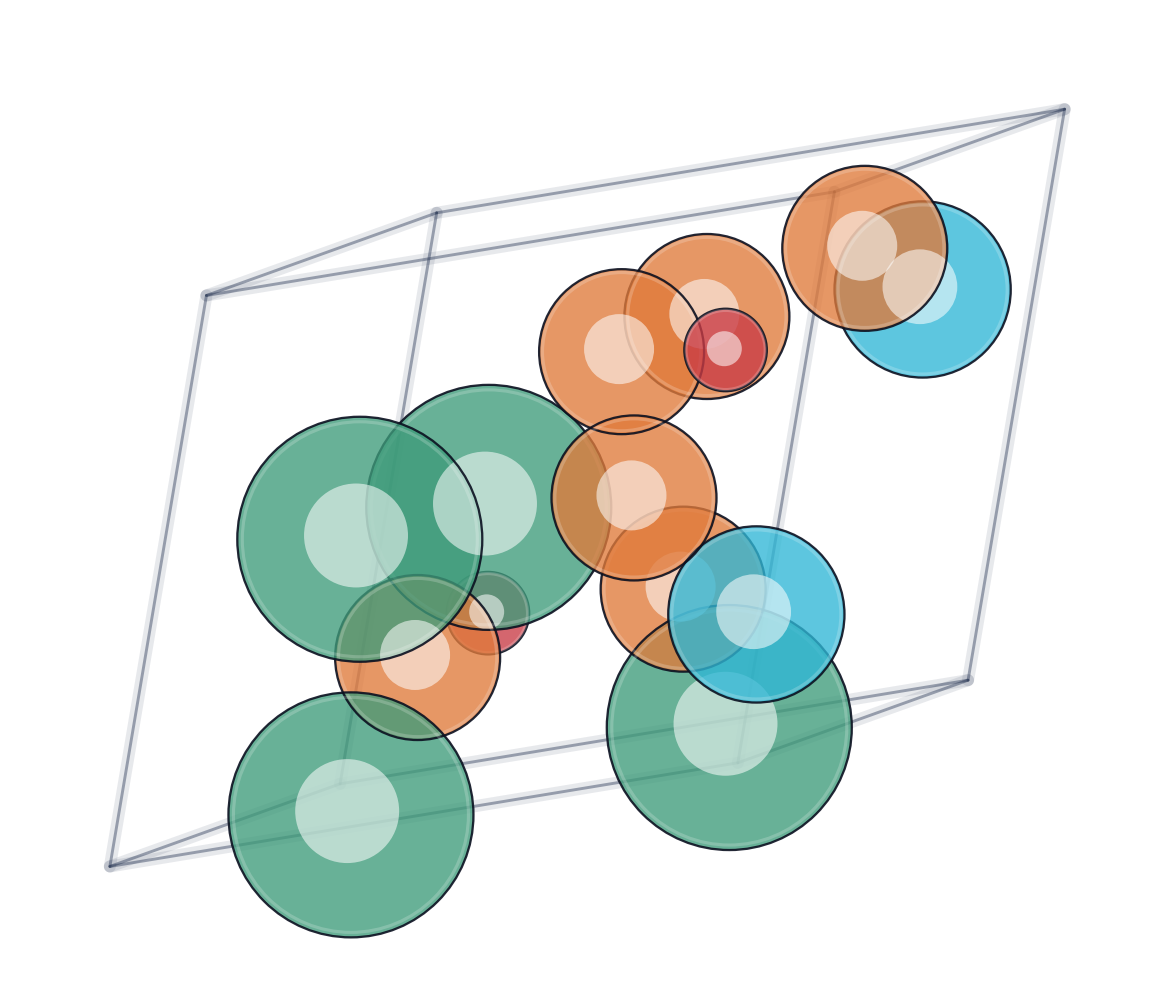}}{MgU$_2$Cu$_3$O} &
\cell{}{\includegraphics[width=\linewidth,height=\cellcardheight,keepaspectratio,trim=10 20 18 20,clip]{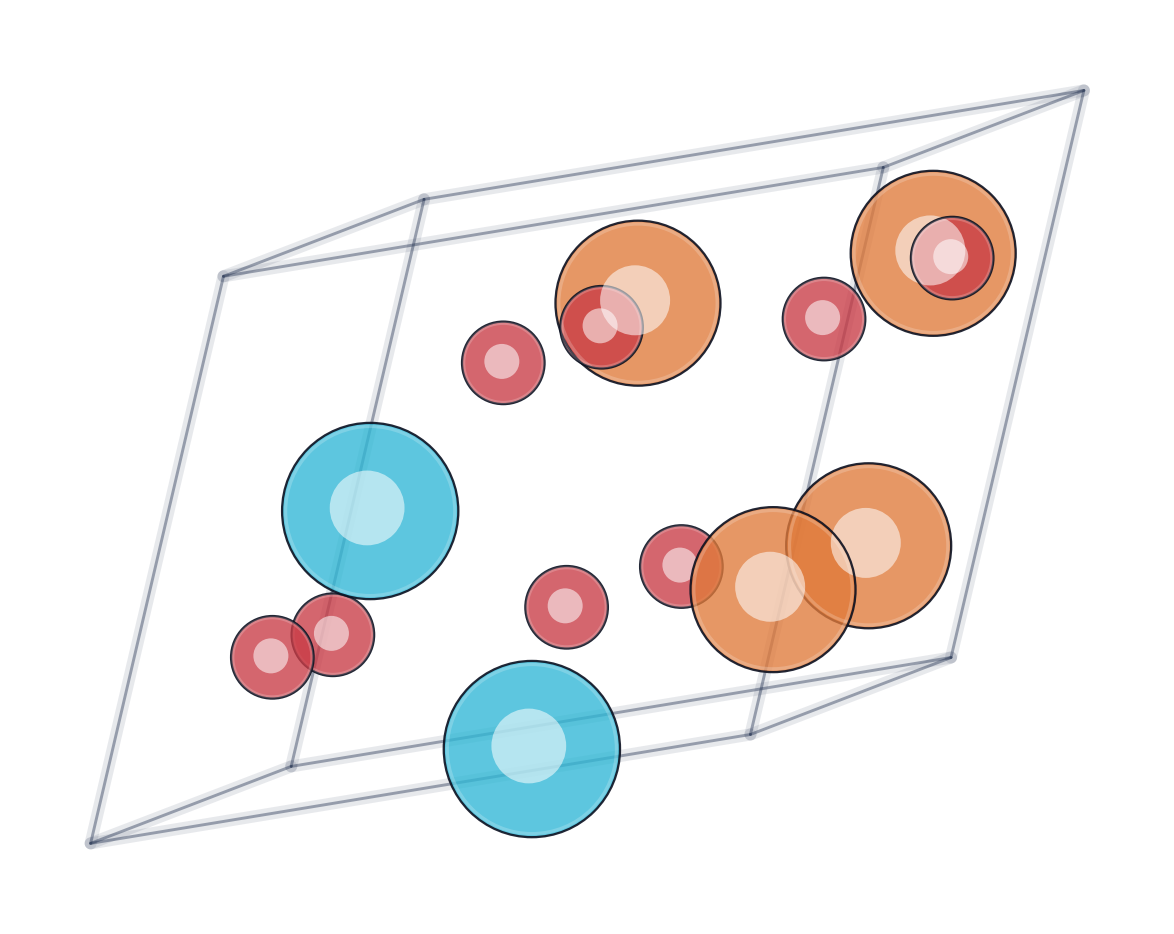}}{Mg(CuO$_2$)$_2$} 
\\[\rowgap]
\cell{}{\includegraphics[width=\linewidth,height=\cellcardheight,keepaspectratio,trim=10 20 18 20,clip]{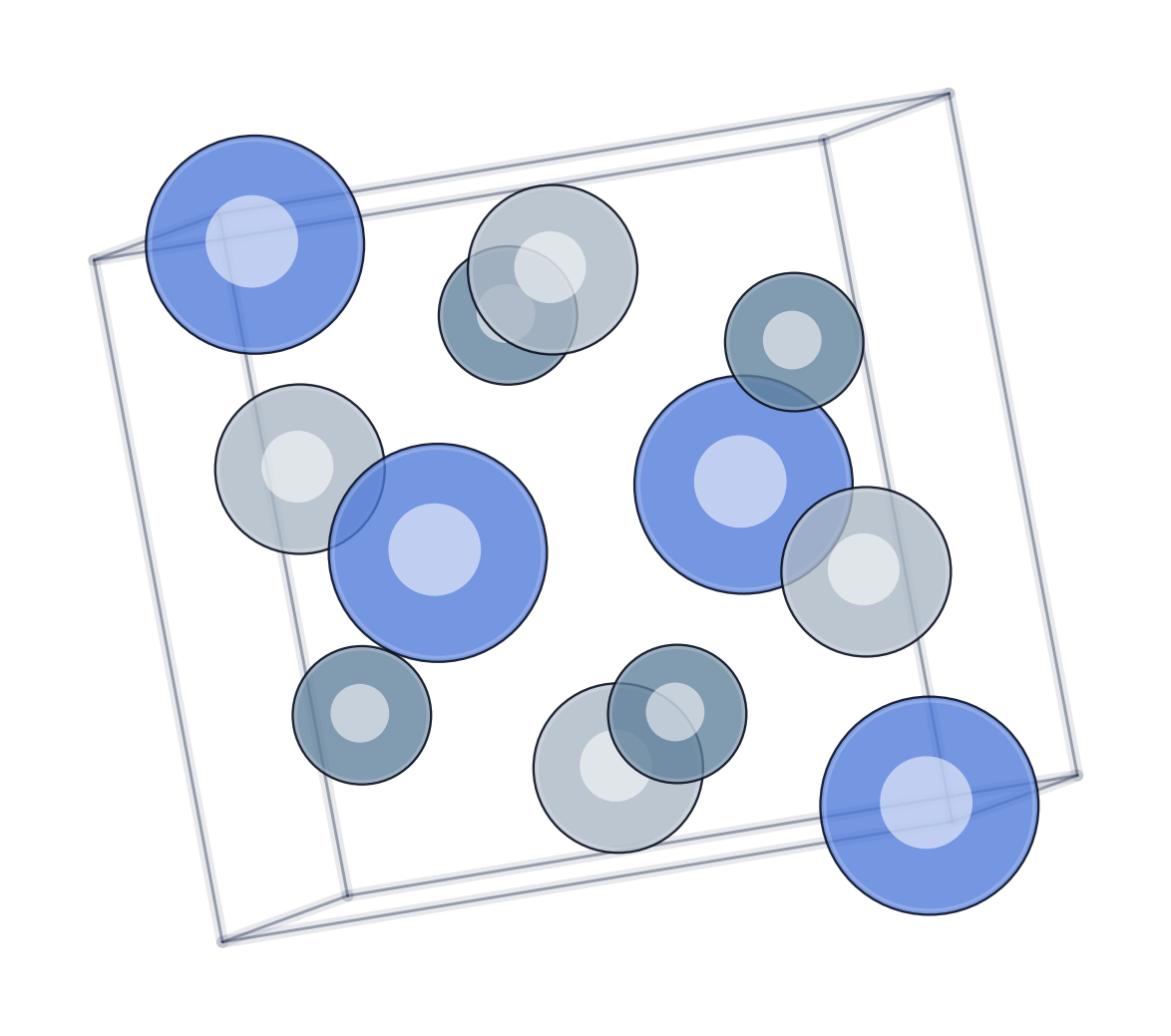}}{ZrSiPt} &
\cell{}{\includegraphics[width=\linewidth,height=\cellcardheight,keepaspectratio,trim=10 20 18 20,clip]{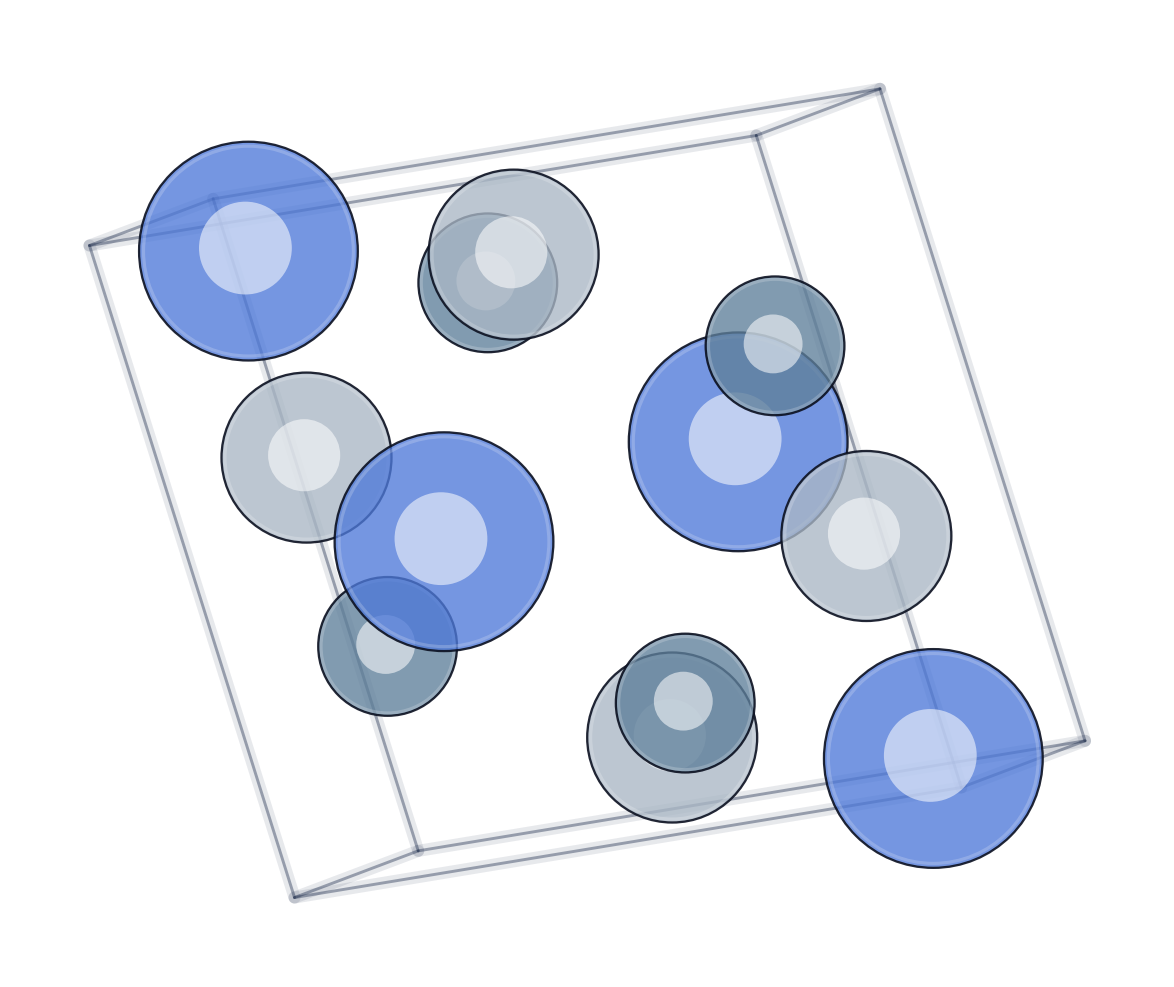}}{ZrSiPt} &
\cell{}{\includegraphics[width=\linewidth,height=\cellcardheight,keepaspectratio,trim=10 20 18 20,clip]{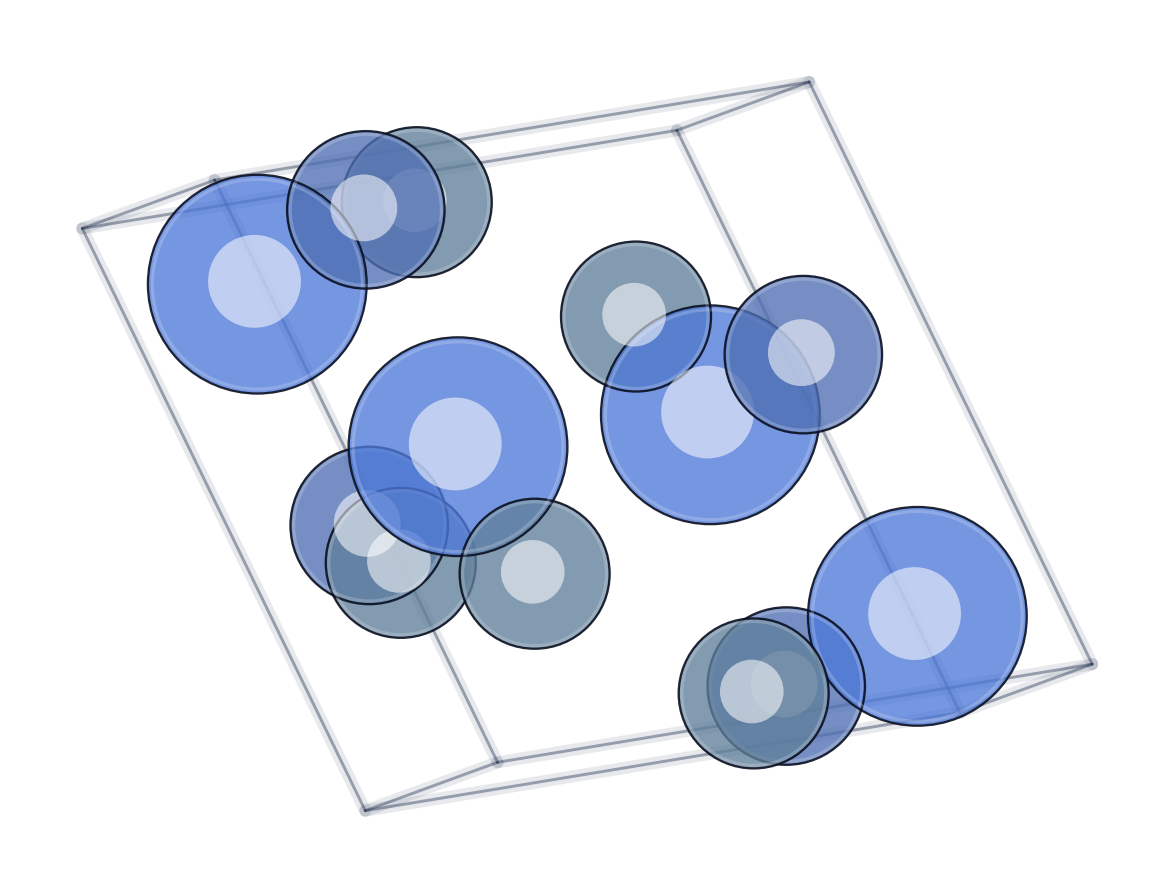}}{Zr$_4$Co$_4$Ge$_5$} &
\cell{}{\includegraphics[width=\linewidth,height=\cellcardheight,keepaspectratio,trim=10 20 18 20,clip]{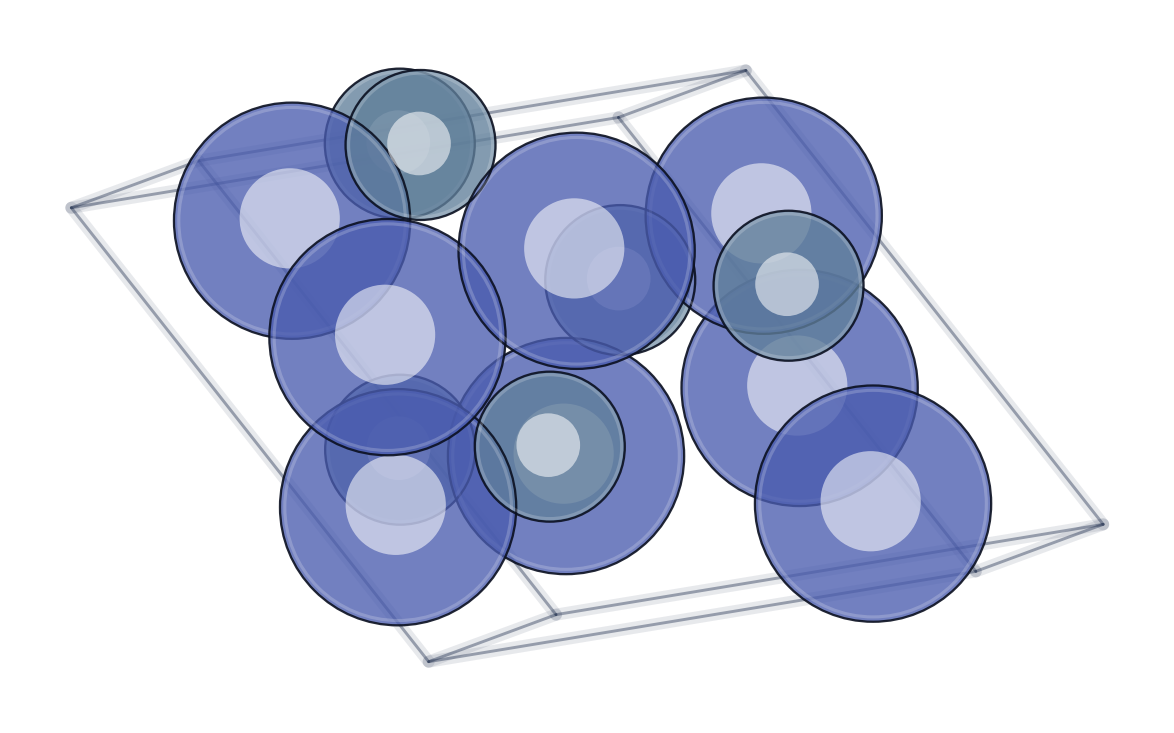}}{Er$_4$Ge$_3$} &
\cell{}{\includegraphics[width=\linewidth,height=\cellcardheight,keepaspectratio,trim=10 20 18 20,clip]{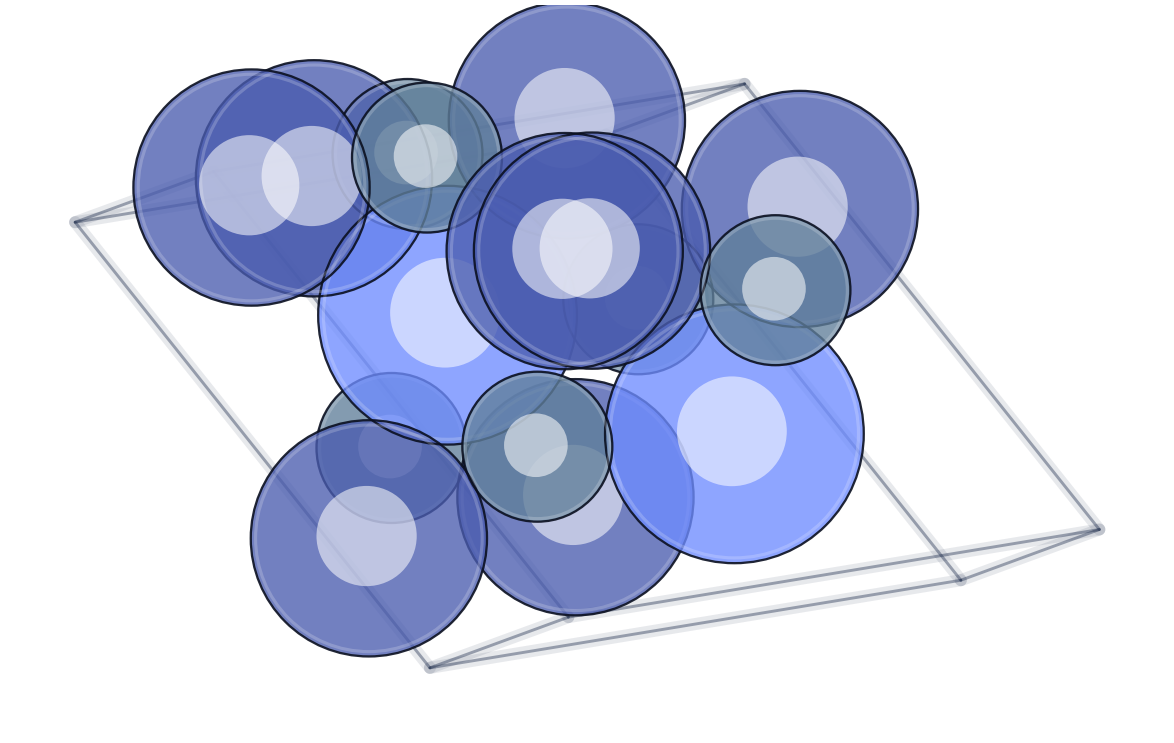}}{LaEr$_4$Ge$_3$} 
\\[\rowgap]
\end{tabular}

\vspace{\platepad}\par
{\color{plate}\rule{\textwidth}{\platerule}}\par

\vspace{-2pt}
\caption{\textbf{Latent interpolation of materials.} Each row decodes $\rvz^{(t)}=(1-t)\rvz^{(0)}+t\rvz^{(1)}$ between a pair of MP-20 materials. Columns show $t=0,0.25,0.50,0.75,1$. Cell shapes, atom positions, and compositions vary along the paths; most composition changes occur at intermediate values of $t$.}
\label{fig:latent-interpolation}
\vspace{-10pt}
\end{figure*}

\subsection{Material Representations}
\vspace{-5pt}
Our model learns representations that are meant to navigate the transdimensional space of materials smoothly. 
To demonstrate their ability to do so, we interpolate two materials, $\mathsf{M}^{(0)}$ and $\mathsf{M}^{(1)}$, by linearly mixing their representations $\rvz^{(0)}$ and $\rvz^{(1)}$.
Figure~\ref{fig:latent-interpolation} shows decoded interpolations for five pairs of MP-20 materials. 
We use Pymatgen's 2D plots for better clarity \citep{ong2013python}.
The resulting materials gradually evolve from $\mathsf{M}^{(0)}$ to $\mathsf{M}^{(1)}$ by altering their unit cell shape, atom positions, and the atomic composition (which happens mainly between the inner timesteps), indicating that the latents smoothly navigate the materials space.
This result motivates examining if one can obtain well-performing (in terms of $f(\mathsf{M})$) materials by mixing existing structures, similarly to a study in \citet{gomez2018automatic}.
To quantify how promising that is, we sample $N_{\text{pair}}=10$ pairs of existing materials and conduct an 8-step linear interpolation, at the same time evaluating $f(\mathsf{M})$ of the decoded, mixed structures.
The results in Figure (\ref{fig:interpolation-f}) show that the target property, unfortunately, tends to be higher along the interpolation trajectory.
In particular, it reaches its apex in the middle of interpolation.
Nevertheless, the failure of this na\"ive approach strengthens the motivation for more sophisticated tools, like MBO.

\begin{figure*}[t]
\centering
\setlength{\tabcolsep}{3pt}
\newcommand{\cellcardheight}{0.68\linewidth} 
\newcommand{\cellxshift}{-0.6mm}

\newcommand{\platerule}{0.55pt}
\newcommand{\platepad}{3pt}
\newcommand{\headgap}{1.5pt}   
\newcommand{\capgap}{-4pt}    
\newcommand{\rowgap}{6pt}      

\definecolor{plate}{RGB}{80,80,80}
\newcommand{\cell}[3]{
  \begin{minipage}[t]{0.18\textwidth}\centering
    \if\relax\detokenize{#1}\relax\else
      {\footnotesize\textbf{#1}}\par\vspace{0.4pt}%
      \vspace{\headgap}%
    \fi
    #2\par\vspace{\capgap}%
    {\footnotesize #3}%
  \end{minipage}%
}

\vspace{-5pt}
{\color{plate}\rule{\textwidth}{\platerule}}\par
\vspace{\platepad}
\noindent\begin{tabular}{@{}ccccc@{}}
\cell{$t=0.00$}{\includegraphics[width=\linewidth,height=\cellcardheight,keepaspectratio,trim=10 20 18 20,clip]{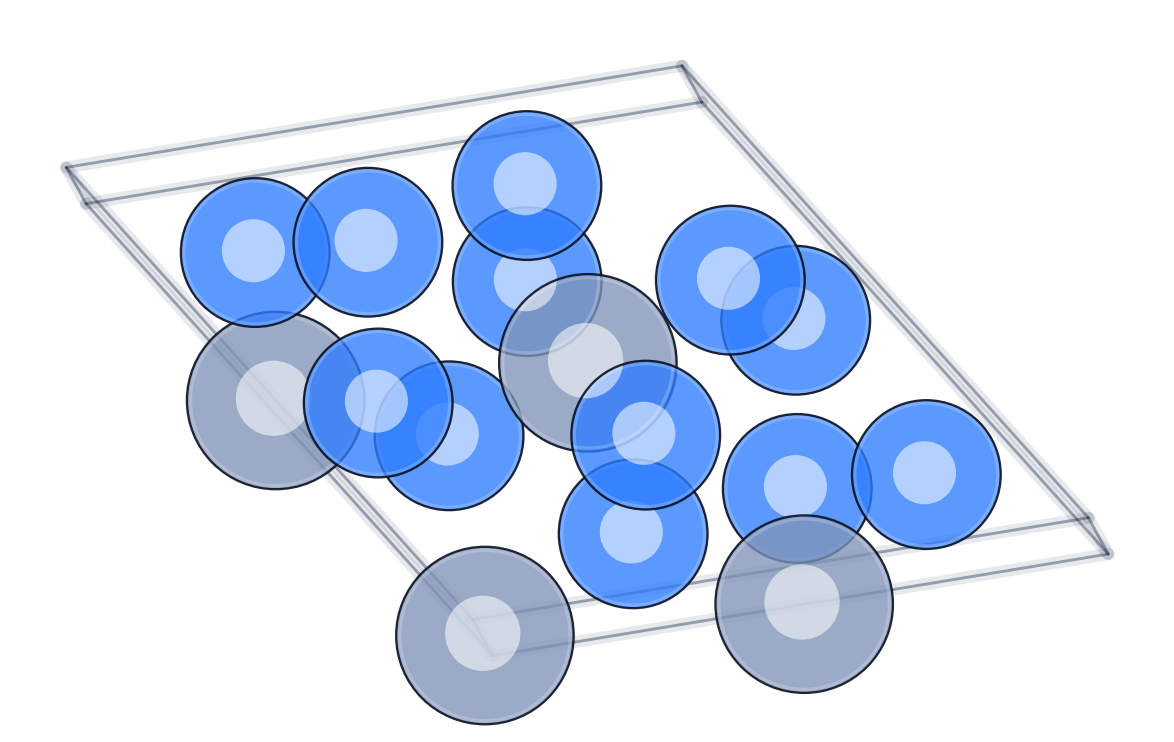}}{As$_3$Rh} &
\cell{$t=0.25$}{\includegraphics[width=\linewidth,height=\cellcardheight,keepaspectratio,trim=10 20 18 20,clip]{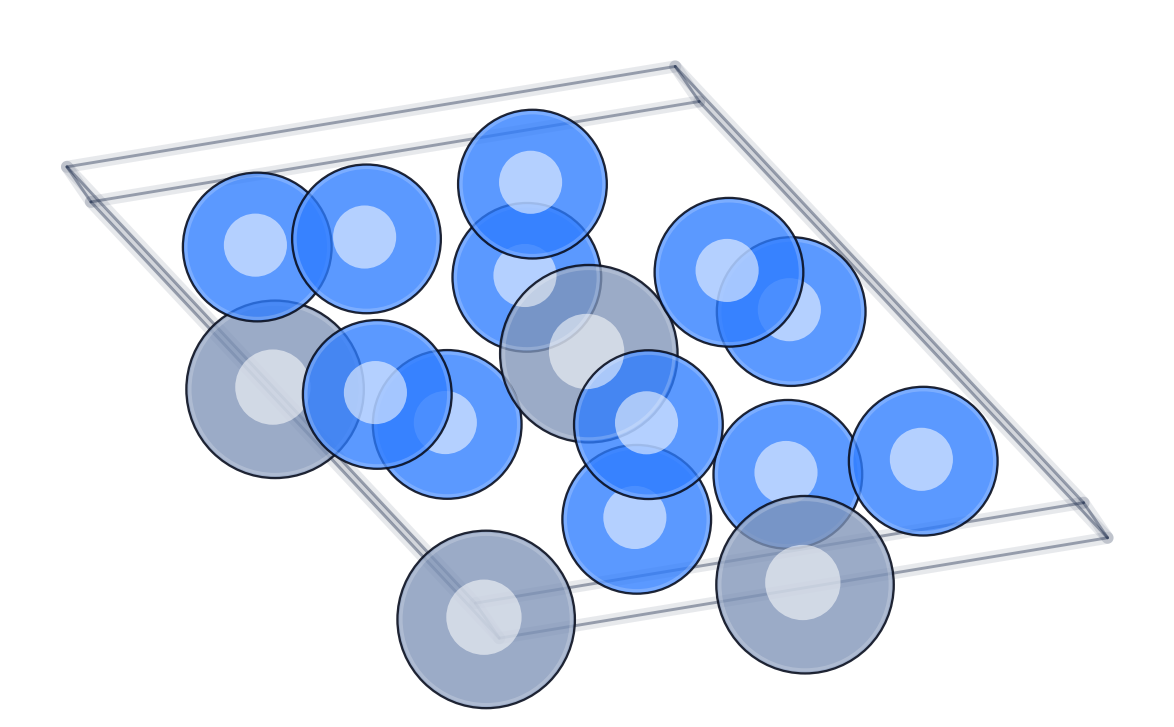}}{As$_3$Rh} &
\cell{$t=0.50$}{\includegraphics[width=\linewidth,height=\cellcardheight,keepaspectratio,trim=10 20 18 20,clip]{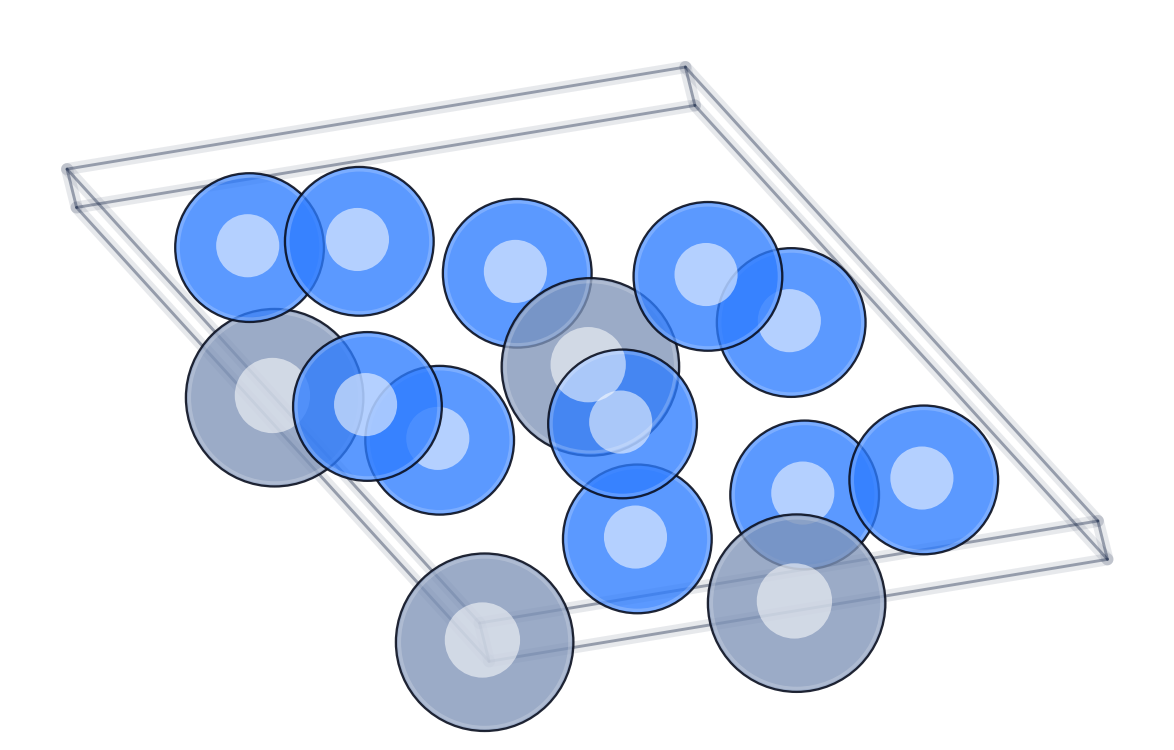}}{As$_{11}$Rh$_4$} &
\cell{$t=0.75$}{\includegraphics[width=\linewidth,height=\cellcardheight,keepaspectratio,trim=10 20 18 20,clip]{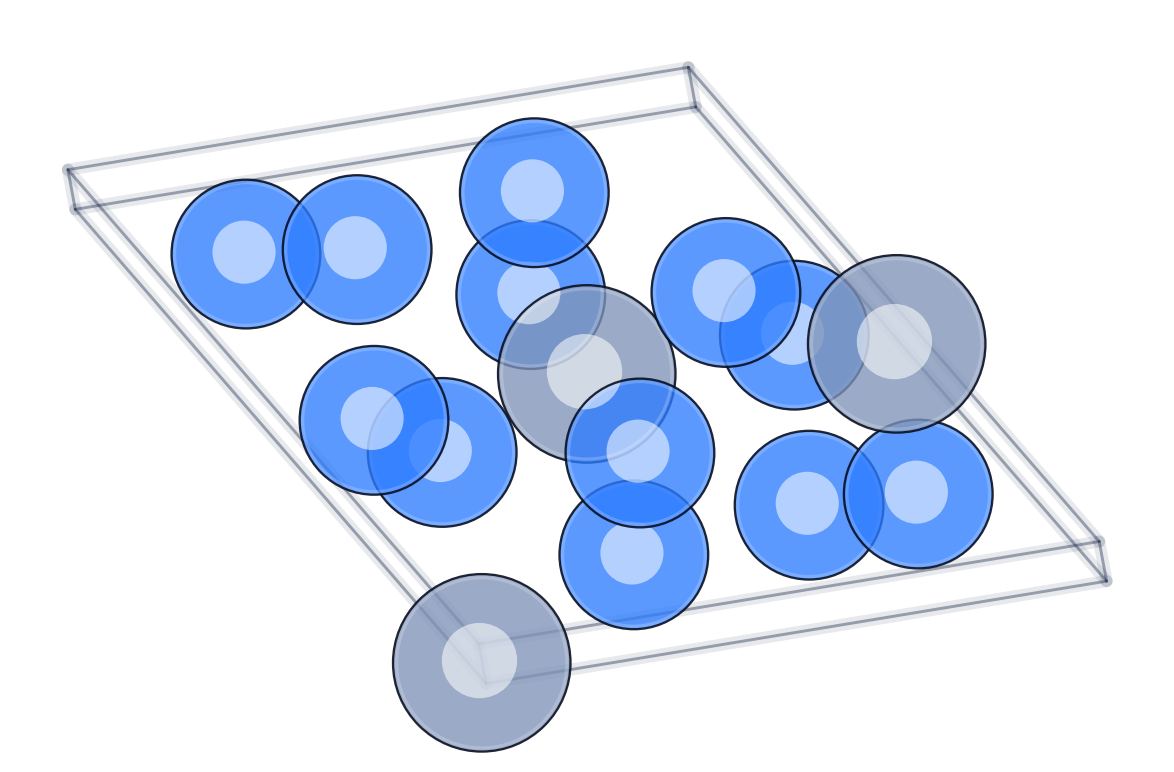}}{As$_4$Rh} &
\cell{$t=1.00$}{\includegraphics[width=\linewidth,height=\cellcardheight,keepaspectratio,trim=10 20 18 20,clip]{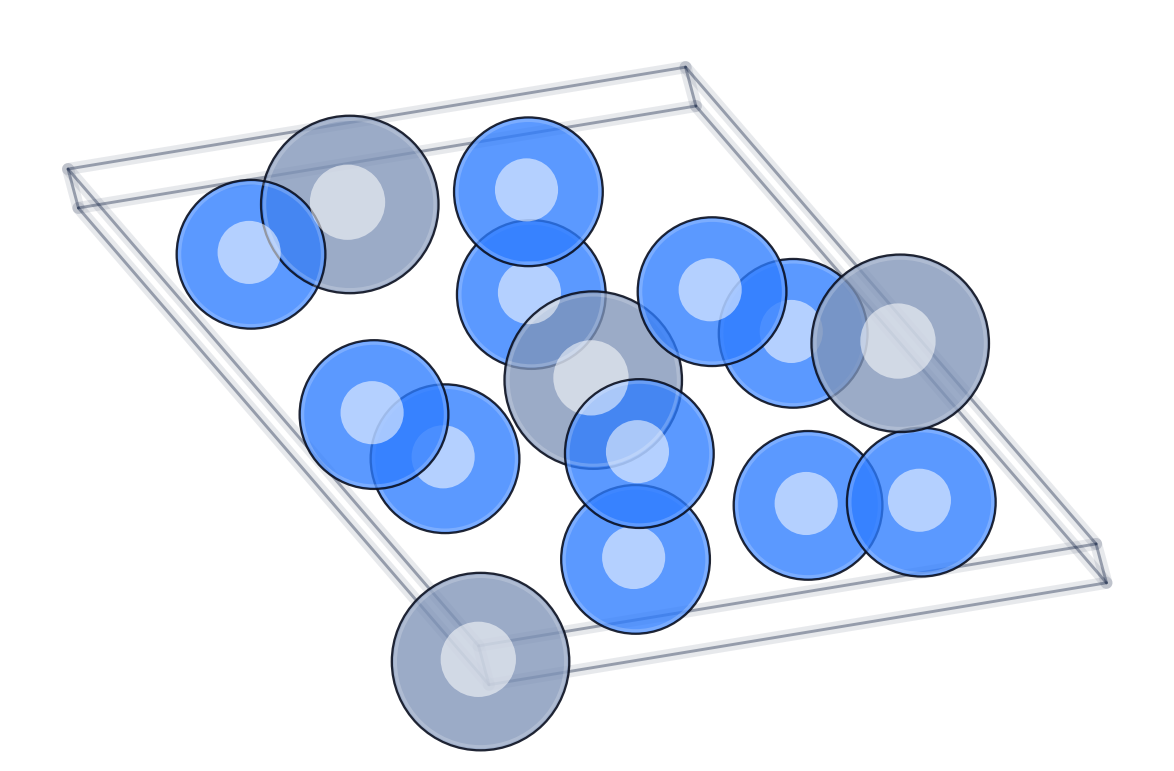}}{As$_{11}$Rh$_4$} \\[\rowgap]
\cell{}{\includegraphics[width=\linewidth,height=\cellcardheight,keepaspectratio,trim=10 20 18 20,clip]{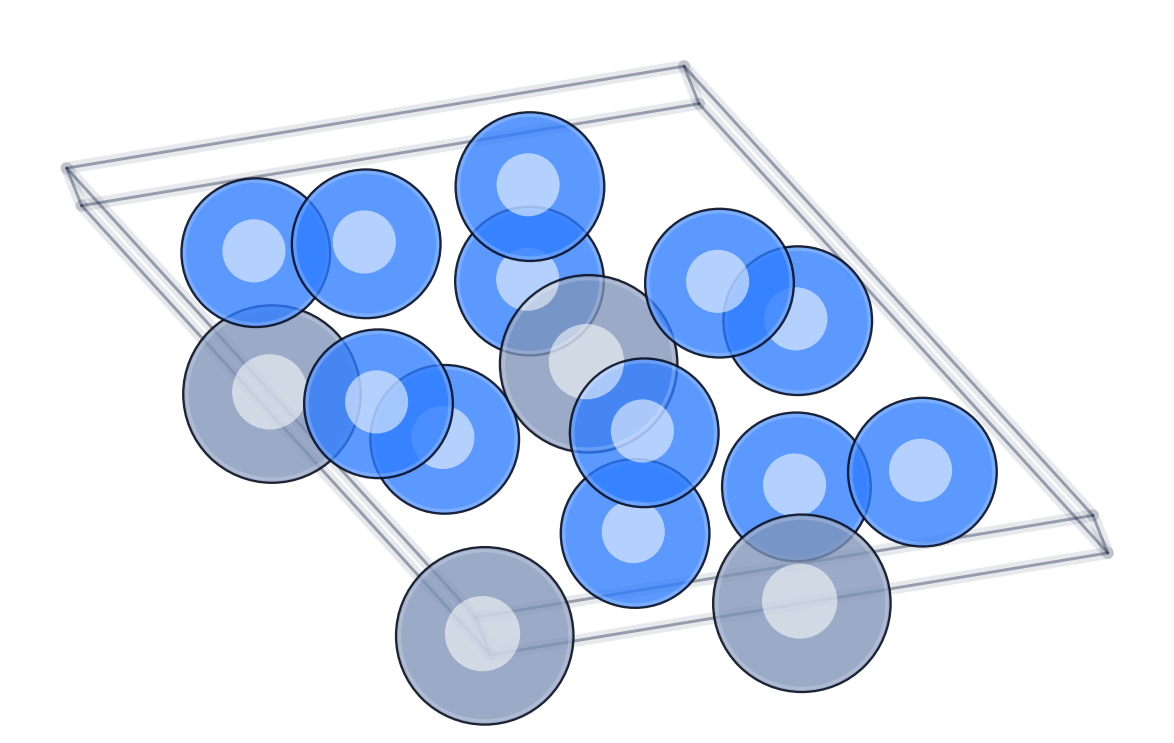}}{As$_3$Rh} &
\cell{}{\includegraphics[width=\linewidth,height=\cellcardheight,keepaspectratio,trim=10 20 18 20,clip]{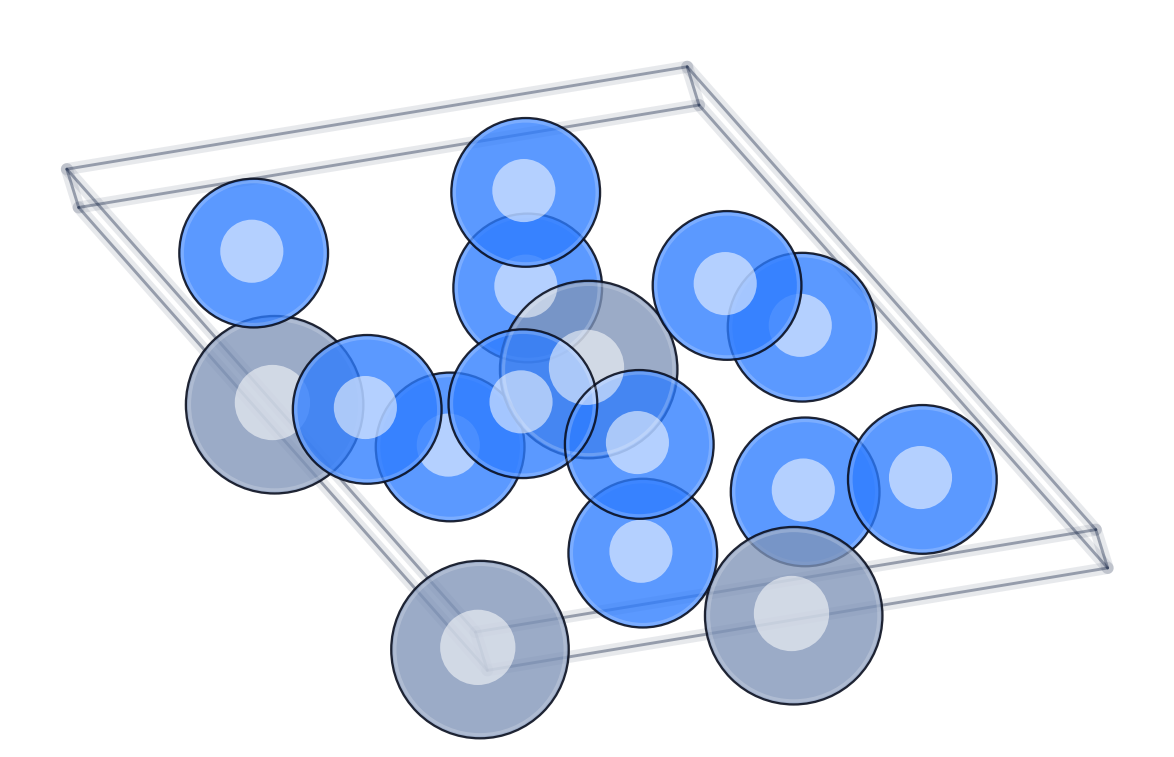}}{As$_3$Rh} &
\cell{}{\includegraphics[width=\linewidth,height=\cellcardheight,keepaspectratio,trim=10 20 18 20,clip]{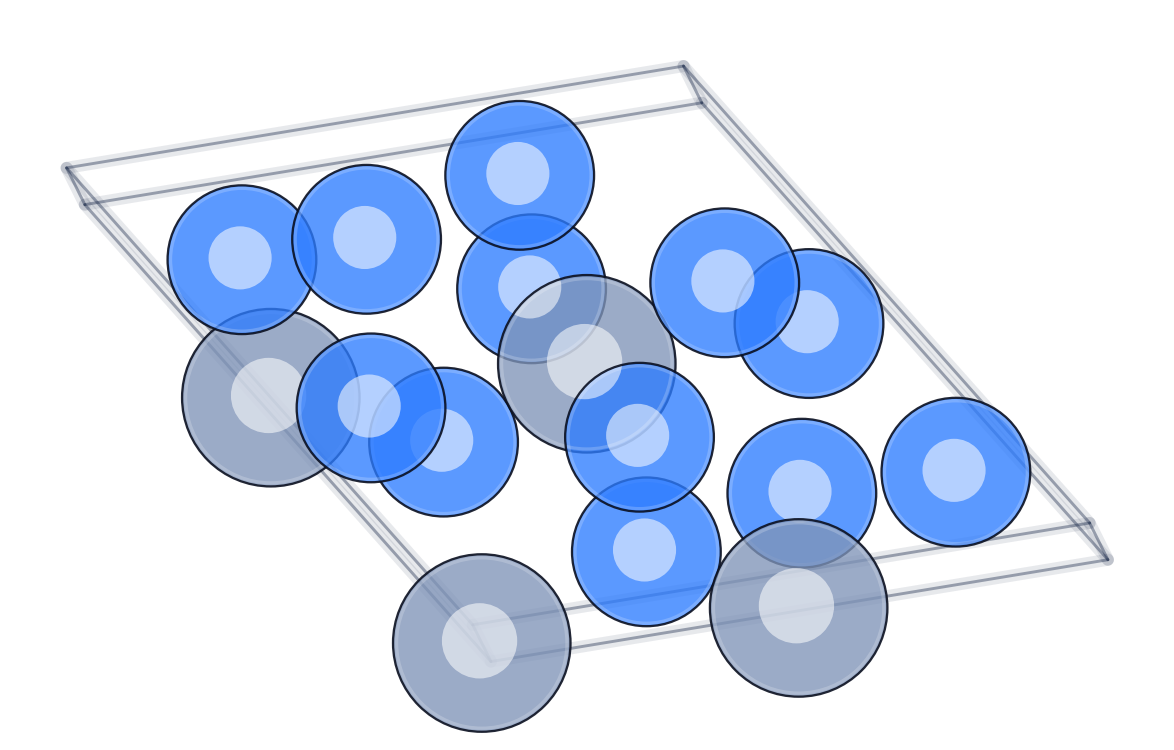}}{As$_3$Rh} &
\cell{}{\includegraphics[width=\linewidth,height=\cellcardheight,keepaspectratio,trim=10 20 18 20,clip]{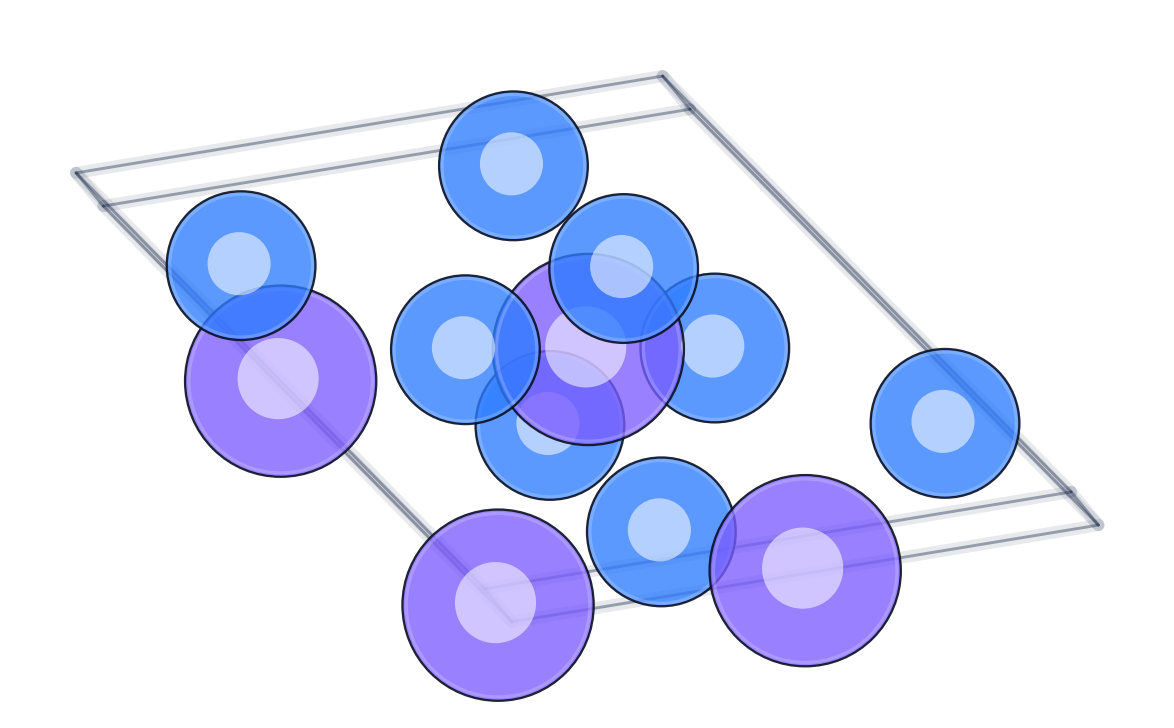}}{VAs$_2$} &
\cell{}{\includegraphics[width=\linewidth,height=\cellcardheight,keepaspectratio,trim=10 20 18 20,clip]{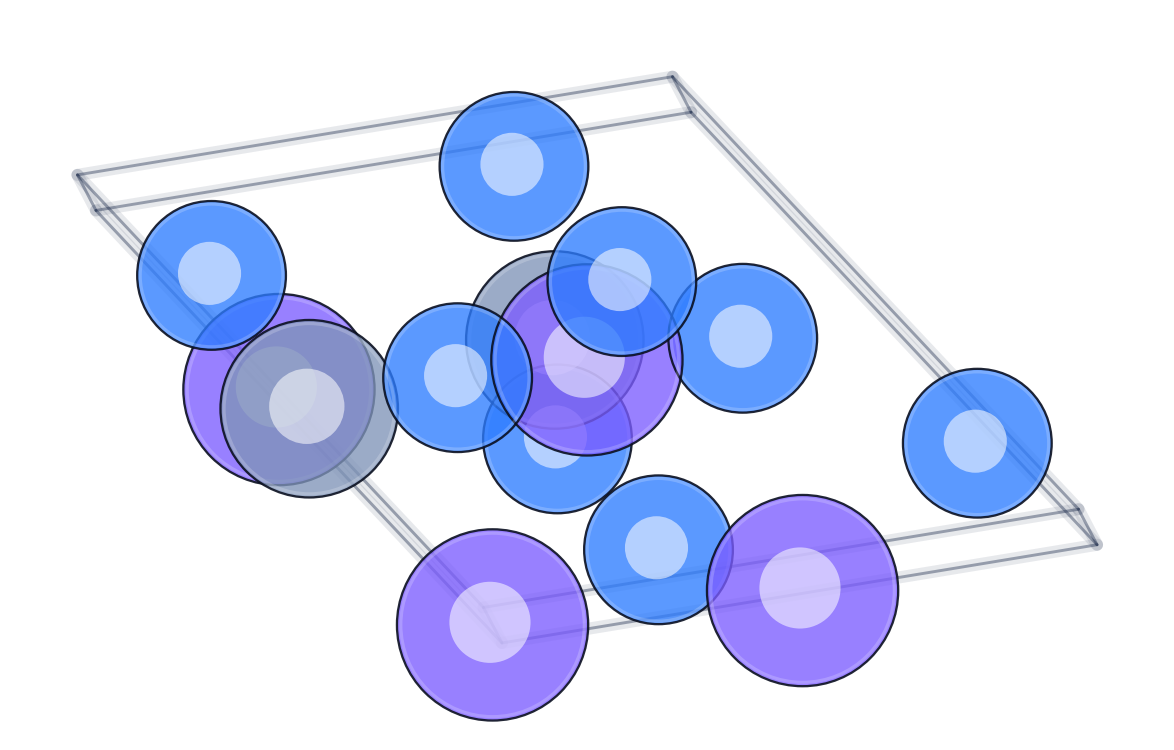}}{V$_2$As$_4$Rh} \\
\end{tabular}

\vspace{\platepad}\par
{\color{plate}\rule{\textwidth}{\platerule}}\par

\vspace{-2pt}
\caption{\textbf{Single-clique interpolation.} Coordinates of clique 1 (top) or clique 6 (bottom) are interpolated from As$_3$Rh toward MgInBr$_3$, while coordinates outside the selected clique remain fixed. Clique 1 retains As and Rh while changing geometry and atom count; clique 6 introduces V at later timesteps. Only a subset of coordinates is interpolated, so $t=1$ need not decode to MgInBr$_3$.}
\label{fig:two-row-clique}
\vspace{-10pt}
\end{figure*}

\paragraph{Intervening on individual cliques.}
We also interpolate a single clique between the latent encodings $\rvz^{(0)}$ and $\rvz^{(1)}$ of As$_3$Rh and MgInBr$_3$. Let $I_c$ denote the coordinate indices belonging to clique $c$ (Section~\ref{sub:pred}). For $t\in[0,1]$, we construct
\begin{equation}
    \label{eq:single-clique-interpolation}
    z_i^{(t)} =
    \begin{cases}
        (1-t)z_i^{(0)} + t z_i^{(1)}, & i\in I_c,\\
        z_i^{(0)}, & i\notin I_c.
    \end{cases}
\end{equation}
Coordinates outside the selected clique remain fixed; shared coordinates also affect neighboring cliques. We decode the resulting latents for $c\in\{1,6\}$ and $t\in\{0,0.25,0.50,0.75,1\}$. Figure~\ref{fig:two-row-clique} shows that clique 1 changes atomic positions and atom counts while retaining As and Rh, whereas clique 6 introduces V. Note that, since only part of the latent changes, interpolations at $t=1$ need not decode to MgInBr$_3$. These examples suggest distinct effects of clique interventions, without establishing a unique physical interpretation for each clique.

\vspace{-8pt}
\section{Related Work}
\vspace{-8pt}
Automating computational materials discovery has motivated growing interest in deep generative modeling for materials \citep{jain2013commentary}. 
Representative approaches include diffusion-based models such as CDVAE \citep{xie2021crystal} and MatterGen \citep{zeni2023mattergen}, flow-based generators such as FlowMM \citep{miller2024flowmm}, and autoregressive models such as CrystalFormer \citep{cao2025space}. 
These methods primarily focus on likelihood-based generation or property-conditioned sampling, whereas our focus is direct property optimization. Unlike prior materials generators, CliqueFlowmer compresses transdimensional crystal structures into fixed-dimensional latent representations explicitly designed for optimization. It combines transformers and latent-variable modeling similarly in spirit to \citet{joshi2025all}, but, crucially, supports compression and gradient-based search in materials space while avoiding expensive equivariant graph architectures \citep{li2025e2former}.

A parallel line of work steers diffusion samplers toward property-favorable Boltzmann distributions \citep{li2408derivative,uehara2024bridging,tan2025scalable,liu2025adjoint}. While these approaches tilt generative sampling toward desirable regions, they do not directly solve optimization over materials.
More closely related are offline model-based optimization methods such as MINs \citep{kumar2020model}, COMs \citep{trabucco2021conservative}, MatchOpt \citep{hoang2025learning}, and especially Cliqueformer \citep{kuba2024cliqueformer}, which introduced clique-based MBO in transformers. However, prior MBO methods target simplified benchmark domains and are not compatible with crystalline materials. CliqueFlowmer bridges this gap by integrating structured MBO with a domain-specific generative architecture for direct optimization in computational materials discovery.

\vspace{-8pt}
\section{Limitations \& Conclusion}
\vspace{-8pt}
In this work, we proposed addressing the problem of computational materials discovery (CMD) with the tools from offline model-based optimization (MBO).
We introduced CliqueFlowmer---a model 
that represents materials as continuous vectors and optimizes them with clique-based MBO. 
Our experimental results with CMD apparatus, such as MP-20 data and target property oracles, show that CliqueFlowmer enables efficient optimization of materials offline.
The major limitation of our work is the impact of optimized properties---while the most easily measurable with oracles, they are not of the greatest interest of industrial labs that design materials for specialized applications. 
Together with increased access to oracles of more properties, we expect CliqueFlowmer to solve more impactful problems.
Appropriate regulatory oversight is essential to ensure that discovered materials are deployed safely and for societal benefit.
Another limitation is the accuracy of our evaluation.
For example, some works estimate properties of generated materials with software conducting (high-cost) density functional theory calculations which are known to be very accurate.
Meanwhile we employed machine learning-based oracles in the majority of our evaluations.
These can induce errors, as confirmed by our limited DFT checks.
Overall, the most strongly validated conclusion of our study is that CliqueFlowmer is highly effective at offline property optimization; by contrast, absolute stability-based metrics such as S.U.N. are more sensitive to the surrogate-oracle stack and should be interpreted with greater caution.
Furthermore, while we focused on property optimization in our research, the continuous representations derived from CliqueFlowmer can be employed by researchers in any manner desired. 
Lastly, we hope that this work will inspire the CMD community to add MBO to its toolkit.

\clearpage
\begin{ack}
Kuba Grudzien's research is supported by a grant from NSF AI4OPT AI Institute for Advances in Optimization (Award 2112533).
Pieter Abbeel holds concurrent appointments as a Professor at UC Berkeley and as an Amazon Scholar. This paper describes work performed at UC Berkeley and is not associated with Amazon.
\end{ack}

\bibliography{final}
\bibliographystyle{plainnat}
\newpage
\appendix
\etocdepthtag.toc{appendix} 
\etocsettagdepth{main}{none}
\etocsettagdepth{appendix}{subsection} 

\section*{Appendix}

\tableofcontents

\clearpage

\section{Supplementary Experimental Results}
\label{app:more-exp}

\subsection{Additional Metrics} 
\label{app:additional}
In addition to the metrics provided in Table \ref{tab:eform_sun_comparison}, we report the metastable rate, uniqueness rate, and novelty rate, of materials provided by the studied models, in Table \ref{tab:all-metrics}. 
Materials that satisfy these three properties (stability, uniqueness, and novelty), are counted in while computing the S.U.N. rate that we report in Section \ref{sec:exp}.

\begin{table*}[t]
    \centering
    {\small
    \CFwidetable{
        \toprule[0.8pt]
        \rowcolor{CFheaderfill}
        \CFheading{Metric} & \CFheading{CrystalFormer} & \CFheading{DiffCSP} & \CFheading{DiffCSP++} & \CFheading{MatterGen} & \CFheading{\textbf{CliqueFlowmer}} & \CFheading{\textbf{-Top}} \\
        \midrule[0.4pt]
        \rowcolor{CFmetricfill}
        \CFmetric{E$_{\text{form}}$ ($\downarrow$)} & \CFnumber{\textcolor{CFbodyink}{0.71}} & \CFnumber{\textcolor{CFbodyink}{0.59}} & \CFnumber{\textcolor{CFbodyink}{0.65}} & \CFnumber{\textcolor{CFbodyink}{0.60}} & \CFnumber{\CFsecond{\ensuremath{-}0.81}} & \CFnumber{\CFbest{\ensuremath{-}0.99}} \\
        Metastable ($\uparrow$) & \CFnumber{\textcolor{CFbodyink}{47.1}} & \CFnumber{\CFsecond{55.5}} & \CFnumber{\textcolor{CFbodyink}{50.2}} & \CFnumber{\CFbest{57.0}} & \CFnumber{\textcolor{CFbodyink}{26.0}} & \CFnumber{\textcolor{CFbodyink}{24.1}} \\
        Stable ($\uparrow$) & \CFnumber{\textcolor{CFbodyink}{14.1}} & \CFnumber{\CFsecond{19.2}} & \CFnumber{\CFbest{19.8}} & \CFnumber{\textcolor{CFbodyink}{18.5}} & \CFnumber{\textcolor{CFbodyink}{14.5}} & \CFnumber{\textcolor{CFbodyink}{13.6}} \\
        Unique ($\uparrow$) & \CFnumber{\textcolor{CFbodyink}{99.6}} & \CFnumber{\CFsecond{99.8}} & \CFnumber{\CFsecond{99.8}} & \CFnumber{\CFbest{100}} & \CFnumber{\textcolor{CFbodyink}{99.5}} & \CFnumber{\textcolor{CFbodyink}{99.6}} \\
        Novel ($\uparrow$) & \CFnumber{\textcolor{CFbodyink}{70.7}} & \CFnumber{\textcolor{CFbodyink}{96.9}} & \CFnumber{\textcolor{CFbodyink}{83.8}} & \CFnumber{\textcolor{CFbodyink}{86.0}} & \CFnumber{\CFsecond{99.5}} & \CFnumber{\CFbest{99.6}} \\
        S.U.N. ($\uparrow$) & \CFnumber{\textcolor{CFbodyink}{12.8}} & \CFnumber{\CFbest{18.6}} & \CFnumber{\CFsecond{18.5}} & \CFnumber{\textcolor{CFbodyink}{17.6}} & \CFnumber{\textcolor{CFbodyink}{14.4}} & \CFnumber{\textcolor{CFbodyink}{13.4}} \\
        E$_{\text{form}}$ / S.U.N. ($\downarrow$) & \CFnumber{\textcolor{CFbodyink}{1.06}} & \CFnumber{\textcolor{CFbodyink}{1.02}} & \CFnumber{\textcolor{CFbodyink}{1.10}} & \CFnumber{\textcolor{CFbodyink}{0.98}} & \CFnumber{\CFsecond{\ensuremath{-}0.65}} & \CFnumber{\CFbest{\ensuremath{-}1.06}} \\
        \midrule[0.35pt]
        \rowcolor{CFmetricfill}
        \CFmetric{Band Gap ($\downarrow$)} & \CFnumber{\textcolor{CFbodyink}{0.52}} & \CFnumber{\textcolor{CFbodyink}{0.63}} & \CFnumber{\textcolor{CFbodyink}{0.48}} & \CFnumber{\textcolor{CFbodyink}{0.57}} & \CFnumber{\CFbest{0.03}} & \CFnumber{\CFsecond{0.07}} \\
        Metastable ($\uparrow$) & \CFnumber{\textcolor{CFbodyink}{47.1}} & \CFnumber{\textcolor{CFbodyink}{55.5}} & \CFnumber{\textcolor{CFbodyink}{50.2}} & \CFnumber{\textcolor{CFbodyink}{57.0}} & \CFnumber{\CFbest{99.9}} & \CFnumber{\CFsecond{99.7}} \\
        Stable ($\uparrow$) & \CFnumber{\textcolor{CFbodyink}{14.1}} & \CFnumber{\textcolor{CFbodyink}{19.2}} & \CFnumber{\textcolor{CFbodyink}{19.8}} & \CFnumber{\textcolor{CFbodyink}{18.5}} & \CFnumber{\CFsecond{61.3}} & \CFnumber{\CFbest{69.4}} \\
        Unique ($\uparrow$) & \CFnumber{\textcolor{CFbodyink}{99.6}} & \CFnumber{\textcolor{CFbodyink}{99.8}} & \CFnumber{\textcolor{CFbodyink}{99.8}} & \CFnumber{\CFbest{100}} & \CFnumber{\CFbest{100}} & \CFnumber{\CFbest{100}} \\
        Novel ($\uparrow$) & \CFnumber{\textcolor{CFbodyink}{70.7}} & \CFnumber{\textcolor{CFbodyink}{96.9}} & \CFnumber{\textcolor{CFbodyink}{83.8}} & \CFnumber{\textcolor{CFbodyink}{86.0}} & \CFnumber{\CFbest{100}} & \CFnumber{\CFbest{100}} \\
        S.U.N. ($\uparrow$) & \CFnumber{\textcolor{CFbodyink}{12.8}} & \CFnumber{\textcolor{CFbodyink}{18.6}} & \CFnumber{\textcolor{CFbodyink}{18.5}} & \CFnumber{\textcolor{CFbodyink}{17.6}} & \CFnumber{\CFsecond{61.3}} & \CFnumber{\CFbest{69.4}} \\
        Band Gap / S.U.N. ($\downarrow$) & \CFnumber{\textcolor{CFbodyink}{0.37}} & \CFnumber{\textcolor{CFbodyink}{0.55}} & \CFnumber{\textcolor{CFbodyink}{0.23}} & \CFnumber{\textcolor{CFbodyink}{0.40}} & \CFnumber{\CFbest{0.02}} & \CFnumber{\CFsecond{0.05}} \\
        \bottomrule[0.65pt]
    }
    \caption{\small
    Target properties and quality metrics across generative baselines and CliqueFlowmer. Formation energies are in eV/atom, band gaps in eV, and all rates in \%. Property / S.U.N. denotes the mean property within the S.U.N. subset, in the same units.
    Row-wise best results are \CFbest{bold teal}; second-best are \CFsecond{underlined blue}.
    We observe that, apart from superior property values, in each task, 
    \textbf{CliqueFlowmer} and CliqueFlowmer\textbf{-Top} attain a very high (near or equal to 100\%) novelty rate. 
    This result satisfies our motivation for development of CMD methods that incorporate MBO and explore the materials space more aggressively than generative baselines.}
        \label{tab:all-metrics}
    }
\end{table*}

\subsection{DFT Evaluations} 
\label{appendix:dft}
\paragraph{Selective DFT.} Due to resource constraints, our full-scale evaluations in Tables \ref{tab:eform_sun_comparison} \& \ref{tab:all-metrics} were done with machine learning oracles, such as M3GNet and MEGNet. 
To ground these results in rigorous physics, however, we have also conducted density functional theory \citep[DFT]{kohn1965self} evaluations of proposed materials. 
Namely, for each method evaluated in Section \ref{sec:exp}, we selected the 100 most stable structures (100 structures with the lowest $E_{\text{hull}}$ values), as predicted by M3GNet.
We relaxed these candidates with DFT and computed formation energies and band gaps.

\paragraph{DFT setup and energy corrections.}
Structural relaxations were performed using VASP \citep{kresse1996efficiency,kresse1996efficient} with inputs generated by pymatgen's \texttt{MPRelaxSet} \citep{ong2013python}, following the standard Materials Project relaxation protocol \citep{jain2013commentary}. The calculated total-energy entries were subsequently processed using \texttt{MaterialsProject2020Compatibility}, which applies the Materials Project 2020 energy-correction scheme and enables consistent comparison of GGA and GGA+$U$ entries \citep{wang2021framework,jain2011formation}. Formation energies are reported per atom (eV/atom). For band-gap calculations, we used approximate HSE \citep{heyd2003hse}, with gaps reported in eV.

\paragraph{DFT results.}
The results in Table \ref{tab:dft} show that CliqueFlowmer retains a strong advantage on the optimized property under DFT evaluation, confirming the central optimization result of the paper. At the same time, the DFT re-evaluation reveals that stability-based metrics such as S.U.N. are substantially more sensitive to bias in the machine-learning oracle pipeline used for development and candidate selection.
As a result, the S.U.N. (and S.U.N.$^\star$) rates of materials proposed by our method are much lower than we previously estimated.
Nevertheless, among S.U.N.$^\star$ materials, CliqueFlowmer (or CliqueFlowmer-Top) delivers the most optimal structures.

\paragraph{The Effect of Oracle-Based Development.}
To study the effect of bias of our evaluation pipeline, we repeated the whole CMD procedure; this time, we ran only $T/2=1000$ latent optimization (ES) steps.
Such a modification can be viewed as \emph{early stopping} \citep{prechelt2002early}, and thus we denote the base length of the latent optimization as \emph{Base}, and the shorter one as \emph{Early}.
Similarly to before, we selected the weight decay of ES by measuring the formation energy and S.U.N. rate derived from M3GNet for $N=100$ structures. 
We found $\lambda=0.2$ to perform best. 
Thus, we repeat our experiments with $1000$ latent optimization steps and weight decay $0.2$, and report the comparative results in Table \ref{tab:oracle_base_vs_early} (oracle evaluation) and Table \ref{tab:dft_base_vs_early} (DFT evaluation).
Under oracle evaluation (Table~\ref{tab:oracle_base_vs_early}), Base achieves lower unfiltered average property values and higher S.U.N. rates than Early. However, Early-Top achieves lower formation energy than Base-Top. DFT evaluation (Table~\ref{tab:dft_base_vs_early}) reveals a tradeoff: early stopping improves stability-based rates, while Base achieves lower band gaps. The best formation-energy result depends on whether all candidates or only the S.U.N.$^\star$ subset are considered.

\begin{figure}
    \centering
    \includegraphics[width=0.9\linewidth]{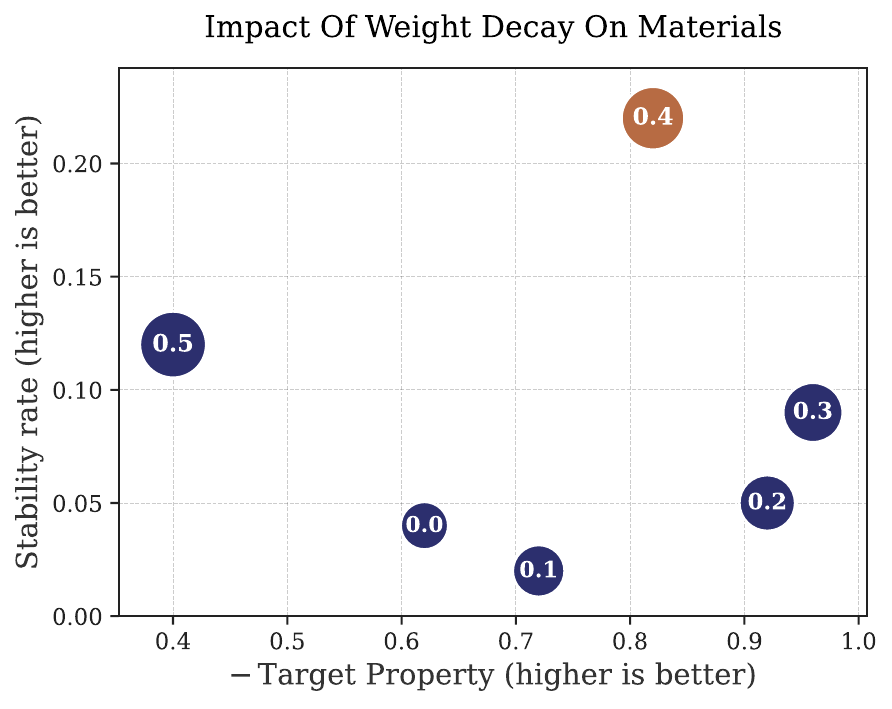}
    \caption{Ablation of the weight decay strength $\lambda$ in the latent-space optimization algorithm. The horizontal axis shows negative formation energy per atom (eV/atom); the vertical axis shows the stable fraction. The selected weight decay $\lambda=0.4$ ({\color{cliquecopper}copper}) delivers the strongest compromise between target property values and stability.}
    \label{fig:tuning-decay}
\end{figure}

These results highlight both a weakness of our paper, as well as a major strength of our methods: hyper-parameter tuning of our methods with ML oracles (like we do, e.g., in Figure \ref{fig:tuning-decay}) introduces unphysical bias into their performance in terms of stability, while they are very robust in terms of property optimization---we find this result satisfying since solving the optimization problem is the main purpose of this paper.
With enhanced DFT infrastructure for development and higher-quality data, we expect CliqueFlowmer to improve its ability to deliver optimized and stable materials. 

\begin{table*}[t]
    \centering
    {\small
    \CFwidetable{
        \toprule[0.8pt]
        \rowcolor{CFheaderfill}
        \CFheading{Metric} & \CFheading{CrystalFormer} & \CFheading{DiffCSP} & \CFheading{DiffCSP++} & \CFheading{MatterGen} & \CFheading{\textbf{CliqueFlowmer}} & \CFheading{\textbf{-Top}} \\
        \midrule[0.4pt]
        \rowcolor{CFmetricfill}
        \CFmetric{E$_{\text{form}}$ ($\downarrow$)} & \CFnumber{\textcolor{CFbodyink}{\ensuremath{-}0.59}} & \CFnumber{\textcolor{CFbodyink}{\ensuremath{-}0.56}} & \CFnumber{\textcolor{CFbodyink}{\ensuremath{-}0.47}} & \CFnumber{\textcolor{CFbodyink}{\ensuremath{-}0.68}} & \CFnumber{\CFsecond{\ensuremath{-}2.43}} & \CFnumber{\CFbest{\ensuremath{-}2.87}} \\
        S.U.N.$^\star$ ($\uparrow$) & \CFnumber{\CFbest{47.00}} & \CFnumber{\textcolor{CFbodyink}{27.00}} & \CFnumber{\textcolor{CFbodyink}{33.00}} & \CFnumber{\CFsecond{38.00}} & \CFnumber{\textcolor{CFbodyink}{12.00}} & \CFnumber{\textcolor{CFbodyink}{4.00}} \\
        S.U.N. ($\uparrow$) & \CFnumber{\CFbest{15.00}} & \CFnumber{\textcolor{CFbodyink}{2.00}} & \CFnumber{\CFsecond{7.00}} & \CFnumber{\textcolor{CFbodyink}{6.00}} & \CFnumber{\textcolor{CFbodyink}{2.00}} & \CFnumber{\textcolor{CFbodyink}{0.00}} \\
        E$_{\text{form}}$ / S.U.N.$^\star$ ($\downarrow$) & \CFnumber{\textcolor{CFbodyink}{\ensuremath{-}0.62}} & \CFnumber{\textcolor{CFbodyink}{\ensuremath{-}0.64}} & \CFnumber{\textcolor{CFbodyink}{\ensuremath{-}0.76}} & \CFnumber{\textcolor{CFbodyink}{\ensuremath{-}0.84}} & \CFnumber{\CFsecond{\ensuremath{-}0.99}} & \CFnumber{\CFbest{\ensuremath{-}2.79}} \\
        \midrule[0.35pt]
        \rowcolor{CFmetricfill}
        \CFmetric{Band Gap ($\downarrow$)} & \CFnumber{\textcolor{CFbodyink}{0.81}} & \CFnumber{\textcolor{CFbodyink}{0.86}} & \CFnumber{\textcolor{CFbodyink}{0.77}} & \CFnumber{\textcolor{CFbodyink}{1.00}} & \CFnumber{\CFsecond{0.20}} & \CFnumber{\CFbest{0.17}} \\
        S.U.N.$^\star$ ($\uparrow$) & \CFnumber{\CFbest{47.00}} & \CFnumber{\textcolor{CFbodyink}{27.00}} & \CFnumber{\textcolor{CFbodyink}{33.00}} & \CFnumber{\CFsecond{38.00}} & \CFnumber{\textcolor{CFbodyink}{2.00}} & \CFnumber{\textcolor{CFbodyink}{3.00}} \\
        S.U.N. ($\uparrow$) & \CFnumber{\CFbest{15.00}} & \CFnumber{\textcolor{CFbodyink}{2.00}} & \CFnumber{\CFsecond{7.00}} & \CFnumber{\textcolor{CFbodyink}{6.00}} & \CFnumber{\textcolor{CFbodyink}{0.00}} & \CFnumber{\textcolor{CFbodyink}{0.00}} \\
        Band Gap / S.U.N.$^\star$ ($\downarrow$) & \CFnumber{\textcolor{CFbodyink}{0.81}} & \CFnumber{\textcolor{CFbodyink}{0.78}} & \CFnumber{\textcolor{CFbodyink}{1.01}} & \CFnumber{\textcolor{CFbodyink}{1.37}} & \CFnumber{\CFbest{0.11}} & \CFnumber{\CFsecond{0.28}} \\
        \bottomrule[0.65pt]
    }
    \caption{\small
    DFT evaluation of 100 structures per method selected by lowest M3GNet-predicted energy above hull. Formation energies are in eV/atom, band gaps in eV, and S.U.N. rates in \%. Property / S.U.N.$^\star$ denotes the mean property within that subset, in the same units.
    Row-wise best results are \CFbest{bold teal}; second-best are \CFsecond{underlined blue}.
    We write S.U.N.$^\star$ when we count metastable materials as stable, and S.U.N. when we only include strictly stable materials.
    }
    \label{tab:dft}
    }
    \vspace{-10pt}
\end{table*}

\begin{table}[t]
\centering
\caption{Comparison, in terms of \textbf{oracle-based} target properties (formation energy and band gap), S.U.N. rates, and properties per S.U.N. for two settings of the latent-space optimization: the \textbf{Base} one ($2000$ ES steps) and the \textbf{Early}-stopped one ($1000$ ES steps). The results are calculated with target property oracles and local hulls. 
Row-wise best results are \CFbest{bold teal}; second-best are \CFsecond{underlined blue}.
Formation energies are in eV/atom, band gaps in eV, and S.U.N. rates in \%; subset-property rows retain the property's units. Base improves the unfiltered average properties and S.U.N. rates, while Early-Top attains lower formation energy.}
\label{tab:oracle_base_vs_early}
{
\begingroup
\arrayrulecolor{CFaccent}
\renewcommand{\arraystretch}{1.16}
\setlength{\extrarowheight}{0.25pt}
\setlength{\aboverulesep}{0pt}
\setlength{\belowrulesep}{0pt}
\begin{tabular}{lcccc}
\toprule[0.8pt]
\rowcolor{CFheaderfill}
\textcolor{CFheading}{\textbf{Metric}} & \textcolor{CFheading}{Base} & \textcolor{CFheading}{Base-Top} & \textcolor{CFheading}{Early} & \textcolor{CFheading}{Early-Top} \\
\midrule[0.35pt]
\rowcolor{CFmetricfill}
\textcolor{CFheading}{\textbf{\boldmath E$_{\text{form}}$ ($\downarrow$)}} 
    & \ensuremath{-}0.81 & \CFsecond{\ensuremath{-}0.99} & \ensuremath{-}0.34 & \CFbest{\ensuremath{-}1.22} \\
S.U.N. ($\uparrow$) 
    & \CFbest{14.4} & \CFsecond{13.4} & 8.3 & 7.6 \\
E$_{\text{form}}$ / S.U.N. ($\downarrow$) 
    & \ensuremath{-}0.65 & \CFsecond{\ensuremath{-}1.06} & \ensuremath{-}0.46 & \CFbest{\ensuremath{-}1.42} \\
\midrule[0.35pt]
\rowcolor{CFmetricfill}
\textcolor{CFheading}{\textbf{\boldmath Band Gap ($\downarrow$)}} 
    & \CFbest{0.03} & \CFsecond{0.07} & 0.38 & 0.25 \\
S.U.N. ($\uparrow$) 
    & \CFsecond{61.3} & \CFbest{69.4} & 28.4 & 30.7 \\
Band Gap / S.U.N. ($\downarrow$) 
    & \CFbest{0.02} & \CFsecond{0.05} & 0.09 & 0.11 \\
\bottomrule[0.65pt]
\end{tabular}
\arrayrulecolor{black}
\endgroup
}
\end{table}

\begin{table}[t]
\centering
\caption{Comparison, in terms of \textbf{DFT-calculated} target properties (formation energy and band gap), S.U.N. rates, and properties per S.U.N. for two settings of the latent-space optimization: the \textbf{Base} one ($2000$ ES steps) and the \textbf{Early}-stopped one ($1000$ ES steps). The results are calculated with DFT and HSE. 
Row-wise best results are \CFbest{bold teal}; second-best are \CFsecond{underlined blue}.
Formation energies are in eV/atom, band gaps in eV, and S.U.N. rates in \%; subset-property rows retain the property's units. Early stopping improves stability-based rates, whereas Base achieves lower band gaps. Early-Top has the lowest overall formation energy; Base-Top has the lowest formation energy within the S.U.N.$^\star$ subset.
}
\label{tab:dft_base_vs_early}
{
\begingroup
\arrayrulecolor{CFaccent}
\renewcommand{\arraystretch}{1.16}
\setlength{\extrarowheight}{0.25pt}
\setlength{\aboverulesep}{0pt}
\setlength{\belowrulesep}{0pt}
\begin{tabular}{lcccc}
\toprule[0.8pt]
\rowcolor{CFheaderfill}
\textcolor{CFheading}{\textbf{Metric}} & \textcolor{CFheading}{Base} & \textcolor{CFheading}{Base-Top} & \textcolor{CFheading}{Early} & \textcolor{CFheading}{Early-Top} \\
\midrule[0.35pt]
\rowcolor{CFmetricfill}
\textcolor{CFheading}{\textbf{\boldmath E$_{\text{form}}$ ($\downarrow$)}} 
    & \ensuremath{-}2.43 & \CFsecond{\ensuremath{-}2.87} & \ensuremath{-}2.15 & \CFbest{\ensuremath{-}3.25} \\
S.U.N.$^\star$ ($\uparrow$) 
    & 12.00 & 4.00 & \CFsecond{21.00} & \CFbest{24.00} \\
S.U.N. ($\uparrow$) 
    & 2.00 & 0.00 & \CFsecond{5.00} & \CFbest{8.00} \\
E$_{\text{form}}$ / S.U.N.$^\star$ ($\downarrow$) 
    & \ensuremath{-}0.99 & \CFbest{\ensuremath{-}2.79} & \CFsecond{\ensuremath{-}1.93} & \ensuremath{-}1.42 \\
\midrule[0.35pt]
\rowcolor{CFmetricfill}
\textcolor{CFheading}{\textbf{\boldmath Band Gap ($\downarrow$)}} 
    & \CFsecond{0.20} & \CFbest{0.17} & 0.42 & 0.30 \\
S.U.N.$^\star$ ($\uparrow$) 
    & 2.00 & \CFsecond{3.00} & \CFbest{7.00} & \CFbest{7.00} \\
S.U.N. ($\uparrow$) 
    & 0.00 & 0.00 & 0.00 & \CFbest{1.00} \\
Band Gap / S.U.N.$^\star$ ($\downarrow$) 
    & \CFbest{0.11} & 0.28 & 0.57 & \CFsecond{0.27} \\
\bottomrule[0.65pt]
\end{tabular}
\arrayrulecolor{black}
\endgroup
}
\end{table}

\subsection{Histogram of Optimized Band Gaps} 
\label{appendix:histogram}

\begin{figure}
    \centering
    \hspace{-10pt}
    \includegraphics[width=\linewidth]{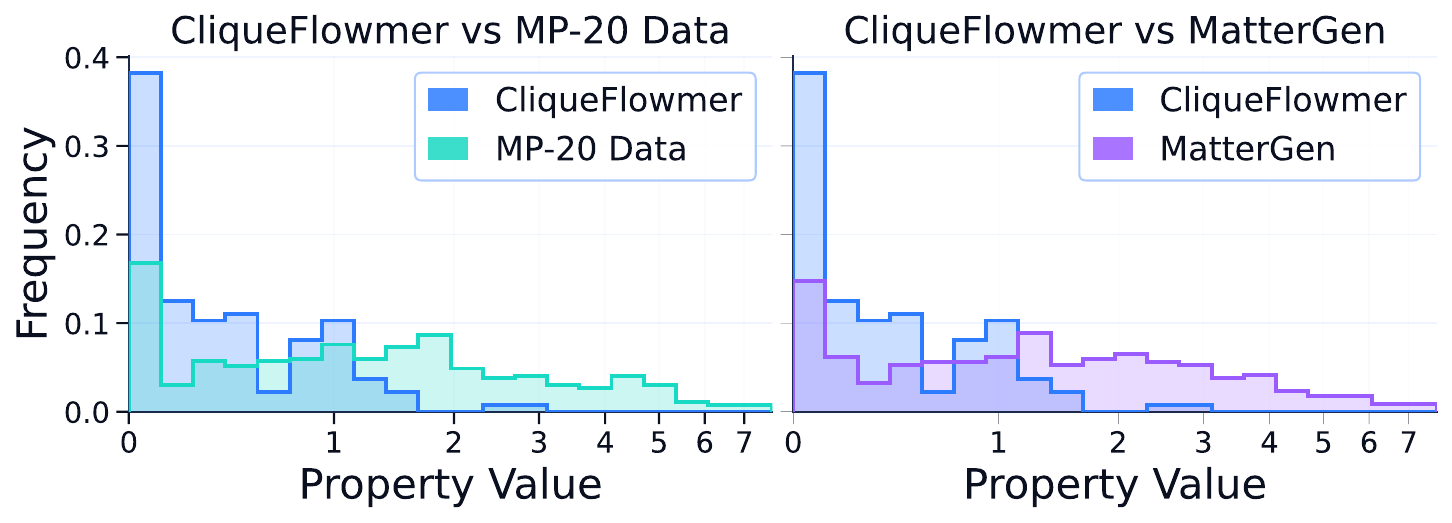}
    \caption{Band-gap distributions among discovered materials. We compare CliqueFlowmer (blue) to materials from the MP-20 dataset (green, \emph{left}) and to those generated by MatterGen (purple, \emph{right}). For this visualization, we omit structures with $\Delta_{\text{band}}=0$; the horizontal axis reports band gap in eV. Table~\ref{tab:eform_sun_comparison} reports the unfiltered property and S.U.N. metrics.
    Materials optimized by CliqueFlowmer have values accumulated near zero while others are more evenly spread out.}
    \label{fig:bg-histograms}
    \vspace{-10pt}
\end{figure}

\begin{figure}
    \centering
    \includegraphics[width=0.95\linewidth]{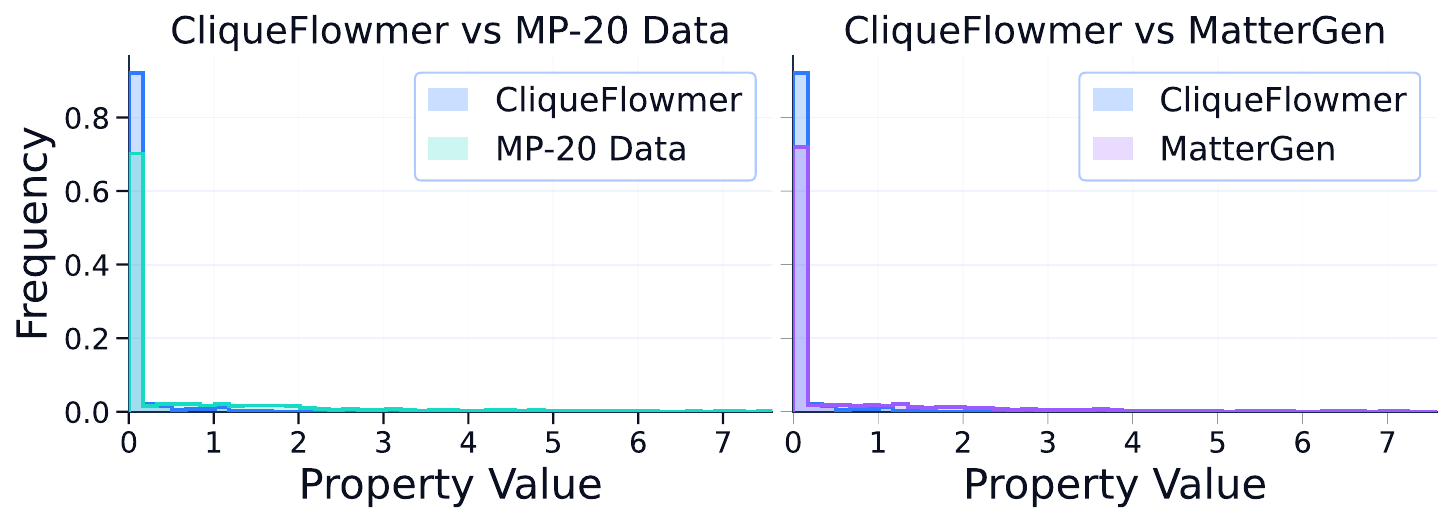}
    \caption{Unfiltered version of Figure~\ref{fig:bg-histograms}, including zero-band-gap structures. The horizontal axis reports band gap in eV.}
    \label{fig:full-histogram}
\end{figure}

Figure \ref{fig:bg-histograms} shows a filtered histogram of band gaps of optimized materials from CliqueFlowmer, materials generated by MatterGen, and materials from the MP-20 dataset.
Figure \ref{fig:full-histogram} presents, for completeness, the unprocessed histogram of ``raw" band gap values.
Both CliqueFlowmer and the reference materials put a substantial probability mass on near-zero band gap materials, obstructing interpretation of the histograms. Hence, Figure \ref{fig:bg-histograms} filtered out materials with band gap exactly equal to zero.

\label{appendix:add-fig}

\subsection{Inference Time Analysis}
\label{app:inference}
Unlike the generative baselines, our model does not simply sample new materials---it optimizes them in the latent space, with evolution strategies (ES, Appendix \ref{app:es}) beforehand. 
This optimization phase, of course, incurs extra computational cost.
In this section, we are going to discuss the resulting computational overhead.
We note that, while doing so, we are not going to account for GPU-accelerated matrix multiplication algorithms. However, we are going to account for in-batch parallelism. 

Recall that, at each latent-space optimization step, we sample $N_{\text{pert}}$ perturbations $\epsilon$ of the optimized latent $\rvz$. 
Then, we predict the target property of antithetic samples with the predictive head, $f_{\theta}(\rvz\pm \sigma \epsilon)$, which is an MLP.
Since the sampling and the evaluation step happen in parallel, the time induced in this step is $\mathcal{O}(N_{\text{layers}}d_{\text{model}}^2)$, where $N_{\text{layers}}$ and $d_{\text{model}}$ are the number of MLP layers and its hidden dimension, respectively. The obtained predictions are then ranked, which induces time cost of $\mathcal{O}(N_{\text{pert}}\log N_{\text{pert}})$ due to a sorting algorithm, such as Quicksort \citep{hoare1961quicksort}.
Then, the ranks are used to derive the ES gradient (Equation (\ref{eq:es-anti})) which, not requiring back-propagation, is a simple algebraic expression that can be omitted. Altogether, a single optimization step induces time cost of $\mathcal{O}(N_{\text{layers}}d_{\text{model}}^2 + N_{\text{pert}}\log N_{\text{pert}})$, and the total time of $T$-step latent-space optimization is
\begin{align}
    \operatorname{Time}(\operatorname{MBO}) = \mathcal{O}\big(T(N_{\text{layers}}d_{\text{model}}^2 + N_{\text{pert}}\log N_{\text{pert}})\big).\nonumber
\end{align}

Note that this time does not depend on the number of materials optimized thanks to the batch parallelism.
Additionally, since the model used in the latent space is a relatively small MLP, the resulting time overhead induced by the MBO phase is negligible in comparison to the time cost of the remaining phase consisting of generative modeling techniques. 
In particular, it stands in contrast with the beam search step (Appendix \ref{app:beam}) which, unlike the flow step, we do not parallelize.
As a result, as the number of optimized materials increases, the beam search step becomes the most costly, although it remains at most linear in the number of structures. 
We note that transformer decoding and beam search can be optimized for efficiency as well \citep{yan2021fastseq, frohner2022parallel} but it is out of scope of this paper.

\begin{figure}
    \centering
    \includegraphics[width=\linewidth]{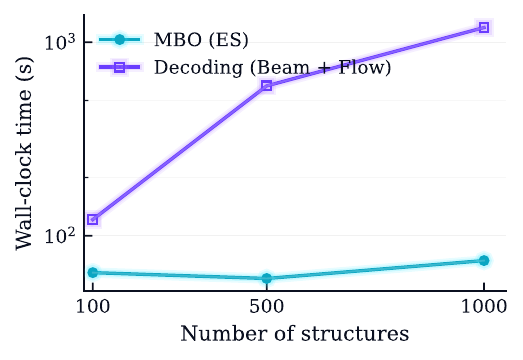}
    \caption{Wall clock time of material optimization with CliqueFlowmer. The optimization (MBO) step remains negligible, especially as the number of materials increases. The wall clock time is dominated by beam search but it remains linear in the number of structures.}
    \label{fig:timing}
    \vspace{-10pt}
\end{figure}

To demonstrate this, we conduct the following experiments. We encode and optimize, in the latent space, $N=\{100, 500, 1000\}$ materials and decode them back into the material form, without relaxation. 
We measure how much time, in seconds, the MBO and the decoding phases took in each experiment. 
For sample sizes 100, 500, and 1000, the MBO step took 64, 60, and 74 seconds, respectively, thus staying relatively constant.
Meanwhile, the decoding step took 121, 595, and 1,192 seconds, respectively.
See Figure \ref{fig:timing} for visualization of the results.
For the batch of 1000 structures, the mean decoding time is 1.19 seconds per structure. This excludes the 74-second latent-optimization stage and structure relaxation. 

\section{Latent-Space Optimization}
In this section, we describe, in more detail, our latent-space optimization algorithm. 
It is this step of our method that delivers new, optimized materials (see Figure \ref{fig:vesta-materials} for their visualization).
\label{appendix:latent}

\subsection{Evolution Strategies}
\label{app:es}
Evolution Strategies (ES) are a class of black-box optimization methods that estimate gradients of an expected objective using random perturbations of the parameters rather than back-propagation through the objective itself. In the scalable formulation introduced by \citet{salimans2017evolution}, ES optimizes parameters $\phi \in \mathbb{R}^d$ by minimizing the smoothed objective
\begin{align}
    \label{eq:es-og}
    F(\phi) = \mathbb{E}_{\epsilon \sim \mathcal{N}(0, I)} \big[ f(\phi + \sigma \epsilon) \big], 
\end{align}
where $\sigma$ controls the perturbation scale. The gradient of $F(\phi)$ can be estimated via Monte Carlo sampling thanks to the following identity,
\begin{align}
    \nabla_{\phi}F(\phi) = \frac{1}{\sigma}\E_{\epsilon \sim \mathcal{N}(0, I)}\big[f(\phi + \sigma \epsilon)\epsilon\big]\nonumber
\end{align}
and used in a standard stochastic gradient descent loop.
In our work, we substitute a representation $\rvz$ of a material $\mathsf{M}$ into the place of $\phi$, and we minimize the expected perturbed approximation $f_{\theta}(\rvz)$ of the target property. 

\paragraph{Rank-Based ES.} A key practical component of modern ES is \emph{rank-based fitness shaping}. Instead of using raw function values $f(\phi + \sigma \epsilon_i)$, ES replaces them with normalized ranks computed across the population of perturbations. 
For example, if the values of, say 3, perturbed parameters $(\phi + \sigma \epsilon_1, \phi + \sigma \epsilon_2, \phi + \sigma \epsilon_3)$  are $(10, -1, 5)$,
then the values are turned into ranks $(3, 1, 2)$.
Then, they are standardized to have mean zero and unit variance, ultimately taking on values $(\sqrt{1.5}, -\sqrt{1.5}, 0)$.
This transformation makes the update invariant to monotone rescalings of the objective, improves robustness to outliers and heavy-tailed noise, and stabilizes optimization when the scale of $f$ varies over time \citep{salimans2017evolution, wierstra2014natural}. As a result, ES behaves less like a noisy gradient estimator and more like a robust search-direction method.

\paragraph{Antithetic Sampling.} To further reduce estimator variance, ES commonly employs \emph{antithetic sampling}. For each sampled perturbation $\epsilon_i$, its negation $-\epsilon_i$ is also evaluated, producing paired function values $f(\phi + \sigma \epsilon_i)$ and $f(\phi - \sigma \epsilon_i)$. The gradient estimate aggregates these symmetric evaluations.
In the function value case, this causes odd-order noise terms to cancel in expectation and significantly reducing variance. 
When used with ranks, it verifies for each perturbation whether to go along or against the direction it points to.

For completeness, we write down the rank-based ES estimator with antithetic sampling for $N_{\text{pert}}$ perturbations.
First, note that each perturbation $\epsilon$ produces values $f(\phi + \sigma \epsilon)$ and $f(\phi - \sigma \epsilon)$, resulting in $2N_{\text{pert}}$ values from all perturbations.
We rank all these values in the ascending order (the lowest value gets the lowest rank). 
Then, for clarity, we rank the perturbations $\epsilon$ based off the ranks of $f(\phi + \sigma \epsilon)$ (where the perturbation) was added, giving $\epsilon^1, \dots, \epsilon^{N_{\text{pert}}}$.
Let $r(i, +)$ and $r(i, -)$ be the ranks of $f(\phi + \sigma \epsilon^i)$ and $f(\phi - \sigma \epsilon^i)$ among all values.
Then, let $R^{i}_{+}$ and $R^{i}_{-}$ be their standardized versions.
The final estimator takes the form
\begin{align}
    \label{eq:es-anti}
    \hat{\nabla}^{\text{ES}}(\phi) = \frac{1}{2\sigma \cdot N_{\text{pert}}}\sum_{i=1}^{N_{\text{pert}}}\big(R^{i}_{+} - R^{i}_{-}\big)\epsilon^i.
\end{align}
Antithetic sampling effectively doubles sample efficiency and is especially beneficial in high-dimensional settings \citep{salimans2017evolution, glasserman2004monte}.

Together, rank-based fitness shaping and antithetic sampling enable ES to scale to millions of parameters while remaining simple, highly parallelizable, and entirely gradient-free, making it a compelling alternative when back-propagation is unavailable, unreliable, or misleading (\emph{i.e.} our case).

\subsection{Examination of Optimization Algorithms}
\label{app:opt}

\begin{figure}
    \centering
    \includegraphics[width=0.95\linewidth]{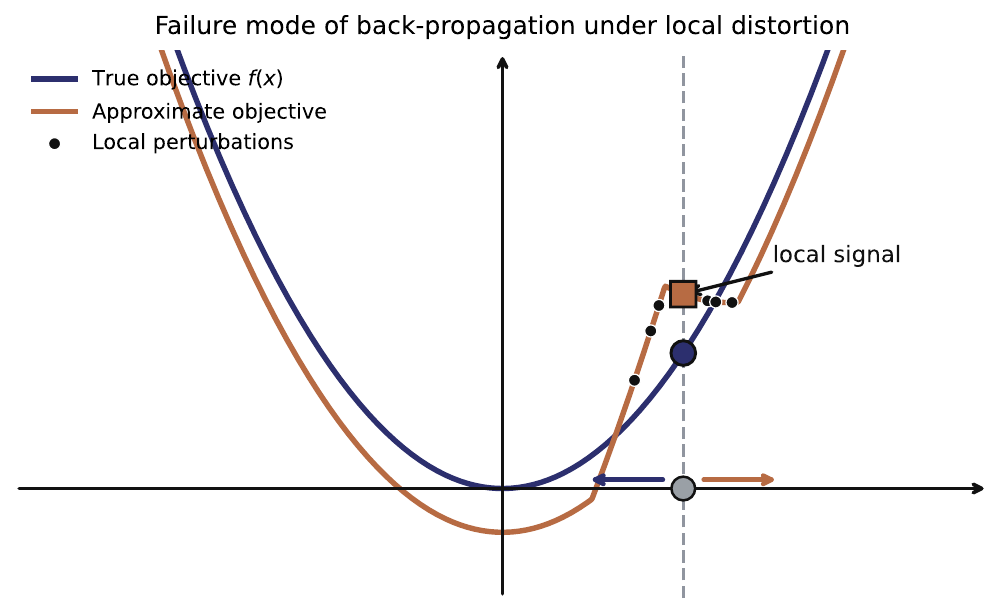}
    \caption{Intuition behind performance discrepancy between BP and ES. The function approximator ({\color{cliquecopper}copper}) of the ground truth function ({\color{cliqueindigo}indigo}) has an artifact that is inconsistent with the ground truth. At the {\color{cliquecopper}copper square} near that artifact ({\color{Gray}gray}), gradient descent on the approximator updates to the right. Meanwhile, exploring the region with ES allows identification of the correct (left) update direction.
    The predicted values of the perturbed points are, from left to right: 0.99, 1.44, 1.67, 1.71, 1.70, and 1.70.}
    \label{fig:errors-bp}
\end{figure}

We verified, in Section \ref{subsec:opt} (Figure (\ref{fig:optimization}), that latent-space optimization of materials with backpropagation-based (BP) gradients diverges and gradients based on rank-based evolution strategies \citep[ES]{salimans2017evolution} are effective.
In this section, we investigate this phenomenon deeper and ablate the weight decay strength of Adam optimization that we use \citep{kingma2014adam, loshchilov2017fixing}.
First, we discuss the failure of back-propagation (BP) from Section \ref{subsec:opt}.
It is known that MBO algorithms are sensitive to distribution shift, wherein optimized designs $\rvx$ become inputs to a predictive model $f_{\theta}(\rvx)$ that makes erroneous predictions about them \citep{kumar2020model, kumar2020conservative, trabucco2021conservative}.
Meanwhile, back-propagation is the most exact way of calculating the steepest direction of change with respect to such an imprecise function approximator.
As a result, a gradient-based algorithm that fully ``trusts" this estimator, commits to the artifacts of its errors, leading to poor designs (see Figure (\ref{fig:errors-bp}) for intuition).

To choose the final value of weight decay $\lambda$, we tested the value of the target property and the stability rate  of materials (fraction of materials below the convex hull) it delivers.
We evaluated weight decay values $\lambda\in\{0, 0.1, 0.2, 0.3, 0.4, 0.5\}$ by applying them to ES and AdamW in a process of optimization of $N=100$ random materials (with M3GNet relaxation). 
We note that we could use more materials to get more accurate estimates, but that would be computationally costly and thus, potentially, not scalable.
This time, we stress-tested the algorithms and ran $T=2000$ optimization steps (twice as many as in Section \ref{subsec:opt}).
As a result, we found higher values of $\lambda$ more effective at optimizing the target property and material stability. 
Ultimately, we found $\lambda=0.4$ to offer the most promising compromise between the target property and stability and thus use it in our experiments.

\subsection{Clique Decomposition Ablation}
\label{appendix:clique}
\begin{figure}[t]
    \centering
    \includegraphics[width=0.95\linewidth]{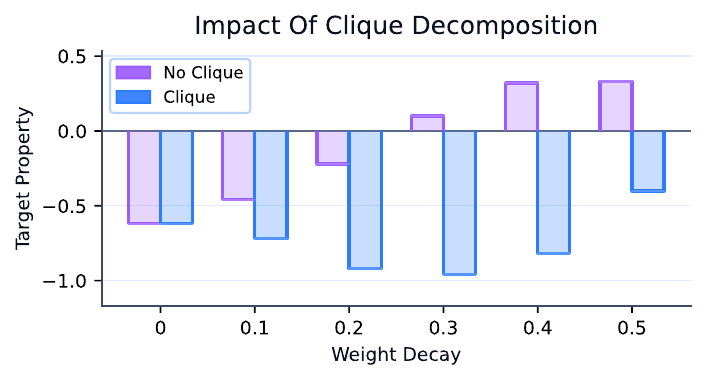}
    \caption{
    Impact of clique decomposition on optimization of formation energy per atom (vertical axis, eV/atom) across different values of ES weight decay.
    Clique decomposition enables substantially stronger optimization performance over a broad range of weight decay values, whereas the no-clique variant degrades rapidly as regularization increases.
    }
    \label{fig:clique_ablation}
\end{figure}

CliqueFlowmer structures its latent space into \emph{cliques} over which the target property prediction is decomposed (see Section \ref{sub:pred}). 
Such a decomposition was proven to reduce the regret of the corresponding MBO problem by \cite{grudzien2024functional}.
The intuition behind this phenomenon is simple---it is easier to optimize functions of fewer variables and, then, compose the solutions, than to optimize an ultra-high-dimensional function offline at once.
For example, if one knows that a target function $f(\rx_1, \rx_2)$ satisfies $f(\rx_1, \rx_2)=f_1 (\rx_1) + f_2 (\rx_2)$, then even if favorable values of the joint $f(\rx_1, \rx_2)$ are rare in the dataset, it suffices to find favorable values of $f_1 (\rx_1)$ and $f_2 (\rx_2)$.
The corresponding $\rx_1^\star$ and $\rx_2^\star$ can then be \emph{stitched} together to form a strong solution $\rvx^\star=(\rx_1^\star, \rx_2^\star)$.
This finding was corroborated empirically by Cliqueformer \citep{kuba2024cliqueformer}, which demonstrated strong empirical gains from incorporating the clique decomposition.
Nevertheless, before ultimately deploying the decomposition in our work, we ablated its impact against a plain latent space (no clique decomposition)---see Figure \ref{fig:clique_ablation}.
Indeed, while we did not sweep the structure's configuration (we just chose the latent space size to be a reasonable power of 2), our results indicate that the structured latent space delivers large MBO gains, in particular when paired with weight decay.
Thus, in this work, we employ the clique decomposition.

\section{Model Design Details}
\label{appendix:model}

\subsection{Flow Conditioning via Cross-Attention}
\label{appendix:cross-attention}
The simplest way to condition the flow denoiser is, just as in the case of the atom-type decoder, via adaptive layer-norm that employs the flat representation $\rvz\in\mathbb{R}^{d_{\rvz}}$ \citep[AdaLN]{peebles2023scalable}. 
Throughout the architecture's development, we tested our design choices with the reconstruction experiment from Appendix \ref{sec:appendix-cfg}.
That is, by studying the rate at which we can encode a material $\mathsf{M}$ and decode it into $\hat{\mathsf{M}}$, so that the two match.
In those experiments, the ``flat" conditioning with AdaLN performed poorly.
A breakthrough came from employing cross-attention as a form of conditioning \citep{vaswani2017attention}.
That is, we would use the clique form of the latent, $\rmZ\in\mathbb{R}^{N_{\text{clique}}\times d_{\text{clique}}}$, and treat it as if it was a sequence of hidden states.
We would then insert a cross-attention layer between the actual flow transformer's hidden states and this ``sequence" between the regular attention and feed-forward layers \citep{vaswani2017attention}.

We examined both conditioning schemes in the reconstruction task with different CFG strengths and present the results in Table \ref{tab:cross-attention}.
The results indicate that cross-attention conditioning delivers superior reconstruction quality. 
Notably, the lowest match ratio recorded with cross-attention (71\%) is higher than the highest match ratio of flat conditioning (63\%).
Furthermore, cross-attention conditioning displayed much lower volatility to CFG strength (the range of results being $80\%-71\%=9\%$) than flat conditioning ($63\%-32\%=31\%$). 
In our final experiments, we used a transformer flow denoiser with cross-attention. 

\begin{table}
\centering
\caption{Effect (match ratio in \%) of cross-attention on encode--decode consistency. The best match ratio is \CFbest{bold teal}.}
\label{tab:cross-attention}
\begingroup
\arrayrulecolor{CFaccent}
\renewcommand{\arraystretch}{1.16}
\setlength{\extrarowheight}{0.25pt}
\setlength{\aboverulesep}{0pt}
\setlength{\belowrulesep}{0pt}
\begin{tabular}{lccc}
\toprule[0.8pt]
\rowcolor{CFheaderfill}
\textcolor{CFheading}{Conditioning} & \textcolor{CFheading}{$\omega=0$} & \textcolor{CFheading}{$\omega=2$} & \textcolor{CFheading}{$\omega=4$} \\
\midrule[0.35pt]
Flat & 63 & 45 & 32 \\
Cross-attention & 71 & \CFbest{80} & 77 \\
\bottomrule[0.65pt]
\end{tabular}
\arrayrulecolor{black}
\endgroup
\end{table}

\subsection{Prior Distribution of Lattice Lengths}
\label{app:prior}
When we decode the geometry $\rmG$ of a material represented by $\rvz$ with composition $\rva$, we initialize it with a sample from a prior distribution 
\begin{align}
    \label{eq:prior-rewrite}
    p_{0}(\rmG|\rva)={\color{RoyalBlue}p^{\text{len}}_{0}(a,b,c|\rva)}\cdot p^{\text{ang}}_{0}(\alpha, \beta, \gamma|\rva)\cdot p^{\text{pos}}_{0}(\rmX|\rva).
\end{align}
Similarly to \citet{miller2024flowmm}, we chose to model ${\color{RoyalBlue}p^{\text{len}}_{0}(a,b,c|\rva)}={\color{RoyalBlue}p^{a}_{0}(a|\rva)} \cdot 
{\color{RoyalBlue}p^{b}_{0}(b|\rva)} \cdot 
{\color{RoyalBlue}p^{c}_{0}(c|\rva)}$ 
as an independent log-normal distribution.
That is, for every $l\in\{a, b, c\}$, we consider it transformation of a variable $l_{\text{nor}}$ via the logarithmic function,
\begin{align}
    \label{eq:logit-normal}
    l = \exp(l_{\text{nor}}) \quad \text{equiv.} \quad l_{\text{nor}} = \log(l) 
\end{align}
and we model $l_{\text{nor}}$ as normally distributed, $l_{\text{nor}}\sim \mathcal{N}(\mu_{\text{l}}, \sigma^{2}_{l})$.

That said, we introduce a small modification that accounts for the physical meaning of this distribution. 
First, we recall that the volume of material $\mathsf{M}$'s unit cell is
\begin{align}
\label{eq:volume}
\text{Vol}(\mathsf{M})
=
abc \cdot
\sqrt{
1
- \cos^2\alpha
- \cos^2\beta
- \cos^2\gamma
+ 2\cos\alpha\cos\beta\cos\gamma
} \ \propto \ abc.
\end{align}
The distribution we described in Equations (\ref{eq:prior-rewrite}) \& (\ref{eq:logit-normal}) does not account for the number of atoms that the initialized unit cell is meant to fit.
Meanwhile, materials with more atoms clearly need larger, in terms of volume, cells.
To impose it, we introduce a simple change---we modify our prior distribution so that the average atom density,
\begin{align}
    \label{eq:den}
    \text{Den}(\mathsf{M}) = \frac{N_{\text{atom}}}{\text{Vol}(\mathsf{M})}\propto \frac{N_{\text{atom}}}{abc}.
\end{align}
is invariant to the atom count.  
We accomplish that by constructing \emph{canonical} lattice lengths, $(\underline{a}, \underline{b}, \underline{c})$, such that 
$\text{Vol}(\mathsf{M})\propto N_{\text{atom}}\underline{a}\underline{b}\underline{c}$, and thus
\begin{align}
    \label{eq:invariant}
    \text{Den}(\mathsf{M}) = \frac{N_{\text{atom}}}{\text{Vol}(\mathsf{M})}\propto \frac{N_{\text{atom}}}{N_{\text{atom}}\underline{a}\underline{b}\underline{c}} = \frac{1}{\underline{a}\underline{b}\underline{c}}.
\end{align}
To construct them, simply, for each variable $l\in\{a, b, c\}$, we define
\begin{align}
    \underline{l} = \frac{1}{\sqrt[3]{N_{\text{atom}}}} \cdot l \quad \text{equiv.} \quad l = \sqrt[3]{N_{\text{atom}}} \cdot \underline{l}, \nonumber
\end{align}
which immediately gives us Equation (\ref{eq:invariant}).
From then on, we model the canonical lengths as log-normal and estimate their distribution parameters (normal mean and standard deviation from data). 

\begin{figure*}[t]
    \centering
    \includegraphics[width=0.48\textwidth]{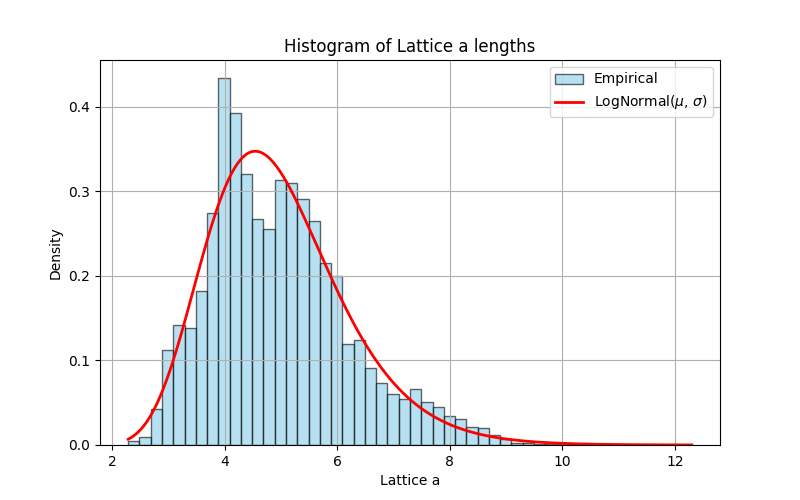}
    \hfill
    \includegraphics[width=0.48\textwidth]{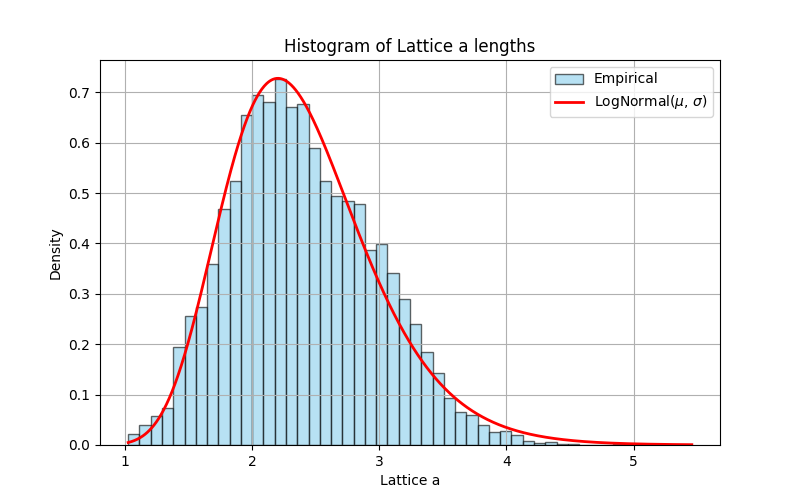}
    \caption{%
    Two ways to model the distribution of lattice lengths, demonstrated on variable $a$. In the first scheme (left), the length is naively modeled as log-normally distributed. In the second, final scheme (right), the length is scaled down by the cubic root of the atom count, and the canonical length is modeled as a log-normal variable.
    Both horizontal axes report lengths in \AA; the vertical axes report probability density in \AA$^{-1}$. The log-normal density ({\color{Red}red}) fits the histogram ({\color{Cyan}blue}) better in the canonical case (right).
    }
    \label{fig:lattice-log-normal}
\end{figure*}

We investigated if this prior fits the data better. Namely, we estimated the distribution of the lattice lengths in two ways---\emph{(1)} without the canonical (atom count-aware) transformation, and \emph{(2)} with the canonical transformation.
In \emph{(1)}, the lengths were modeled as log-normal, and in \emph{(2)}, they were modeled as atom count-scaled log-normal variables. 
We demonstrate the improved quality of method \emph{(2)} in Figure (\ref{fig:lattice-log-normal}).

\subsection{Classifier-Free Guidance for Diffusion and Flow Models}
\label{sec:appendix-cfg}

Classifier-free guidance (CFG) is a technique originally introduced in the context of diffusion models to control the trade-off between sample quality and diversity without requiring an explicit classifier \citep{ho2022classifier}. The key idea is to train a single conditional generative model that can operate both conditionally and unconditionally, and to combine the two modes at sampling time to bias generation toward the conditioning signal.

Formally, let $V_\theta(\rvx_t, t \mid c)$ denote a diffusion or flow model conditioned on some context $c$, and let $V_\theta(\rvx_t, t \mid \varnothing)$ denote the same model evaluated without conditioning. Classifier-free guidance constructs a guided vector field
\begin{align}
V^\omega_\theta(\rvx_t, t \mid c)
= (1 + \omega)\, V_\theta(\rvx_t, t \mid c)
\;-\;
\omega\, V_\theta(\rvx_t, t \mid \varnothing),\nonumber
\end{align}
where $\omega \ge 0$ is the guidance strength. Increasing $\omega$ amplifies features correlated with the conditioning signal, at the cost of reduced diversity and potential distributional shift.
Of course, $\omega=0$ is equivalent to not using CFG---it corresponds to the vanilla Euler method.

While CFG was originally developed for diffusion models, the same principle applies directly to continuous normalizing flows and flow-matching models, where the learned vector field defines an ordinary differential equation (ODE) rather than a stochastic reverse process \citep{lipman2022flow}. In this setting, guidance modifies the deterministic velocity field used during integration, biasing trajectories toward regions favored by the conditioning signal.

\paragraph{CFG in CliqueFlowmer.}
In CliqueFlowmer, classifier-free guidance is used during decoding of material geometry with the flow-matching model. The conditioning signal is the optimized latent representation $\rvz$, while the unconditional model is obtained by replacing $\rvz$ with Gaussian noise $\varepsilon_z \sim \mathcal{N}(0, I)$. During sampling, the guided vector field is integrated from $t=0$ to $t=1$ using an explicit ODE solver.

\paragraph{Reconstruction Experiment.}
To study the effect of guidance strength, we conducted an encode--decode consistency experiment. Given a material $\mathsf{M}$, we encoded it into a latent representation $\rvz$ using the CliqueFlowmer encoder and then decoded it back into a material $\hat{\mathsf{M}}$ using the geometry flow with different guidance strengths $\omega \in \{0, 2, 4\}$. We then evaluated whether $\mathsf{M}$ and $\hat{\mathsf{M}}$ represent the same crystal structure using \texttt{StructureMatcher} from \texttt{pymatgen}, which accounts for lattice symmetries, atomic species, and fractional coordinates.

We report the \emph{match ratio}, defined as the fraction of decoded structures that were deemed equivalent to their original inputs, over the sample of 100 structures. Results are shown in Table~\ref{tab:cfg-results}.

\begin{table}
\centering
\caption{Effect of classifier-free guidance strength on encode--decode consistency. The best match ratio is \CFbest{bold teal}.}
\label{tab:cfg-results}
\begingroup
\arrayrulecolor{CFaccent}
\renewcommand{\arraystretch}{1.16}
\setlength{\extrarowheight}{0.25pt}
\setlength{\aboverulesep}{0pt}
\setlength{\belowrulesep}{0pt}
\begin{tabular}{cc}
\toprule[0.8pt]
\rowcolor{CFheaderfill}
\textcolor{CFheading}{Guidance strength $\omega$} & \textcolor{CFheading}{Match ratio (\%)} \\
\midrule[0.35pt]
0 & 71 \\
2 & \CFbest{80} \\
4 & 77 \\
\bottomrule[0.65pt]
\end{tabular}
\arrayrulecolor{black}
\endgroup
\end{table}

\paragraph{Discussion.}
Moderate classifier-free guidance ($\omega = 2$) substantially improves reconstruction fidelity compared to no guidance, indicating that conditioning on the latent representation is underutilized without explicit amplification. However, excessive guidance ($\omega = 4$) degrades performance, likely due to over-sharpening of the vector field and reduced robustness to modeling errors. This behavior mirrors observations in diffusion-based image and molecule generation, where intermediate guidance strengths often yield the best balance between faithfulness and stability \citep{ho2022classifier}.

Based on these results, all main experiments in this work use a guidance strength of $\omega = 2$ during geometry decoding.

\subsection{Lifted Logit-Normal Distribution}
\label{app:lifted}
\begin{figure}
    \centering
    \includegraphics[width=\linewidth]{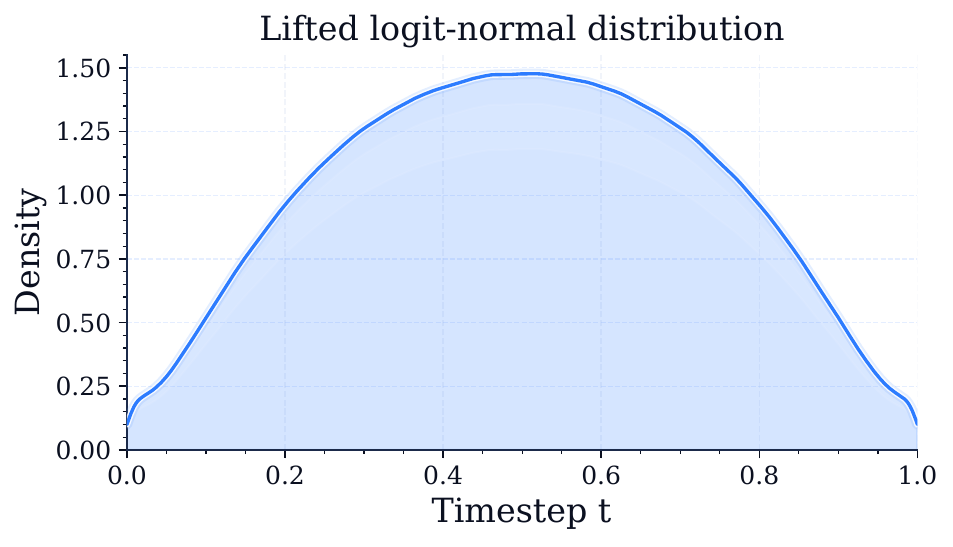}
    \caption{Illustrative density of the lifted logit-normal distribution with $\epsilon=0.2$; training uses $\epsilon=0.1$.
    Most of the mass is placed on values of $t$ in the middle of the range $(0,1)$
    and a substantial amount near the endpoints.}
    \label{fig:lifted-ln}
    \vspace{-10pt}
\end{figure}

While training our geometry decoder with flow matching, we sampled the denoising timestep from the lifted logit-normal distribution, defined as
\begin{align}
    t &= (1-\rb) \cdot t_{LN} + \rb \cdot t_{U} \nonumber \\ 
    \rb\sim \operatorname{Ber}(\epsilon), \ & t_{LN} \sim \operatorname{LogitNorm}(0, 1), \ t_{U} \sim U[0,1],\nonumber
    \vspace{-15pt}
\end{align}
to prioritize the challenging denoising tasks from the midpoint timesteps \citep{esser2024scaling} and ensure sufficient coverage of the edge timesteps (see Figure \ref{fig:lifted-ln}). We set $\epsilon = 0.1$.
\section{Additional Background}
\label{sec:appendix-background}
This section provides additional background that helps understand the contribution of our work. 

\subsection{Computational Materials Discovery Practices}
\label{sec:appendix-CMD}

Computational materials discovery (CMD) aims to identify materials $\mathsf{M}$ whose physical or chemical properties, often, optimize a desired objective $f(\mathsf{M})$, such as formation energy or electronic behavior. 
Simultaneously, the materials should be thermodynamically stable in order to be plausible to synthesize.
Stability can be determined by calculating the structure's energy above hull $E_{\text{hull}}$, which can be done with density functional theory \citep[DFT]{kohn1965self} calculations and a reference set in form of a convex hull.
A material is (strictly) \emph{stable} if its energy above hull is non-positive $E_{\text{hull}}(\mathsf{M}) \leq 0$.
Nevertheless, in practice, many materials with slightly positive $E_{\text{hull}}\in(0, 0.1]\text{eV/atom}$ are synthesizable as well.
To account for this phenomenon, a material is said to be \emph{metastable} if its energy above hull is at most 0.1 eV/atom \citep{jain2013commentary, kirklin2015open, sun2016thermodynamic}.
The MP-20 dataset used in this work consists of metastable materials \citep{xie2021crystal}.
Evaluating candidate materials using first-principles methods such as density DFT is accurate but computationally expensive, motivating the construction of large offline datasets such as the Materials Project \citep{jain2013commentary}.

Recent machine-learning approaches to CMD broadly fall into two categories. \emph{Surrogate modeling} methods learn predictors of material properties and use them for screening or ranking candidates. \emph{Generative modeling} methods—based on VAEs, diffusion models, or flows—learn to sample new materials resembling those in a reference dataset \citep{xie2021crystal,zeni2023mattergen,cao2025space}. While generative models can produce valid and diverse structures, their likelihood-based training objective concentrates probability mass near the empirical data distribution, limiting their ability to aggressively explore property-optimal regions.

Offline model-based optimization (MBO) offers an alternative paradigm by directly optimizing a learned surrogate objective using only offline data \citep{kumar2020conservative,trabucco2022design}. Applying MBO to CMD is challenging due to the hybrid discrete-continuous structure of materials and their transdimensionality. CliqueFlowmer addresses this challenge by learning a fixed-dimensional latent representation of materials that admits structured optimization and can be decoded back into valid crystal structures.

\subsection{Next-Token Prediction with Transformers}
\label{sec:appendix-ntp}

Next-token prediction is a standard training paradigm for autoregressive sequence models, including transformers \citep{vaswani2017attention}. Given a sequence of discrete tokens $(x_1,\dots,x_T)$, a causal transformer models the factorized distribution
\begin{align}
    p(\rx_1,\dots,\rx_T) = \prod_{t=1}^T p(\rx_t \mid \rx_{<t}), \nonumber
\end{align}
and is trained by minimizing the negative log-likelihood of each token conditioned on its prefix.

In CliqueFlowmer, next-token prediction is used to model the sequence of atom types in a material. Each structure is represented as a variable-length sequence augmented with explicit $\langle\operatorname{Start}\rangle$ and $\langle\operatorname{Stop}\rangle$ tokens. The autoregressive transformer is conditioned on the latent representation via adaptive layer normalization (AdaLN) \citep{peebles2023scalable}, allowing global structural information to influence all token predictions. This approach provides a flexible mechanism for handling discrete, variable-length outputs while integrating naturally with transformer architectures and beam search at generation time.

\subsection{Beam Search}
\label{app:beam}
Beam search \cite{wu2016googles} is a heuristic decoding algorithm for approximately maximizing the joint probability of an output sequence under an autoregressive model. Given a conditional next-token distribution $p_\theta(\rx_{k}\mid \rvx_{<k})$, beam search maintains a set (the \emph{beam}) of the top-$K$ partial hypotheses at each step, scored by accumulated log-probability,
\begin{align}
    s(\rvx_{1:k}) \;=\; \sum_{i=1}^{k} \log p_\theta(\rx_i \mid \rvx_{<i}).\nonumber
\end{align}
At the index $k$, each hypothesis in the beam is expanded by all (or a pruned subset of) next-token candidates, producing a pool of new hypotheses; the top-$N_{\text{beam}}$ by score are retained. Decoding terminates when all hypotheses have emitted the {$\langle$Stop$\rangle$} token, or when a maximum length is reached, and the best completed sequence is returned. 

In CliqueFlowmer, we apply beam search to decode the discrete atom-type sequence from the autoregressive transformer conditioned on the optimized latent representation, using beam width $N_{\text{beam}}=10$. This decoding step is used after latent-space optimization to produce a high-likelihood composition consistent with the latent code, before sampling the continuous geometry with the flow-based decoder.

\subsection{Functional Graphical Models and Clique-Based Representations}
\label{sec:appendix-fgm}

CliqueFlowmer builds on the framework of \emph{functional graphical models} introduced by \citet{grudzien2024functional}. In this framework, the target function of interest is modeled as an additive decomposition over overlapping cliques of latent variables. Concretely, a latent vector $\rvz \in \mathbb{R}^{d_z}$ is reshaped into a chain of overlapping cliques
\begin{align}
\rmZ = \texttt{chain}(\rvz, d_{\text{clique}}, d_{\text{knot}})\nonumber
\end{align}
where each clique shares a subset of variables (knots) with its neighbors. The surrogate objective is then parameterized as
\begin{align}
f_\theta(\rvz) = \sum_{c=1}^{N_{\text{cliques}}} f_\theta(\rmZ_c, c),\nonumber
\end{align}
where each term depends only on a local clique.

This structured decomposition induces a functional graphical model in which each clique corresponds to a factor, enabling \emph{compositional generalization}. In particular, clique-based models support \emph{stitching}: optimal in-distribution configurations of individual cliques can be recombined to form globally competitive solutions \citep{fu2020d4rl, kuba2024cliqueformer}.
To understand this intuitively, one can think of a 2D function $\ry=f(\rx_1, \rx_2)$, such that $f(\rx_1, \rx_2)=f_1(\rx_1)+f_2(\rx_2)$.
Knowing of this decomposition, given a dataset $\mathcal{D}=\{(\rx_{1}^i, \rx_{2}^i, \ry^i)\}_{i=1}^{N}$, one can \emph{stitch} the best examples $\rx_{1}^{\star}=\argmin_{\rx_1}f_1(\rx_1)$ and $\rx_{2}^{\star}=\argmin_{\rx_2}f_2(\rx_2)$, into a valid, joint solution $\rvx^{\star}=(\rx^{\star}_1, \rx^{\star}_2)$.
This property has been shown to substantially improve the effectiveness of offline MBO in high-dimensional design spaces.

In CliqueFlowmer, this clique structure is imposed on the latent representation of materials, enabling gradient-based or derivative-free optimization directly in latent space while maintaining a strong inductive bias toward in-distribution solutions. The decoder then maps optimized latent variables back into discrete atom types and continuous geometries, allowing clique-based MBO to operate over the transdimensional space of materials.

\clearpage
\section{Model and Optimization Hyperparameters}
\label{sec:appendix-hparams}

Table~\ref{tab:hyperparams} summarizes the hyperparameters used throughout all experiments. Unless otherwise stated, these values are fixed across runs. 
We trained our model on 8 \emph{NVIDIA GeForce GTX 1080} GPUs.

\begin{table}
\centering
\caption{CliqueFlowmer hyperparameters.}
\label{tab:hyperparams}
\small
\setlength{\tabcolsep}{4pt}
\begingroup
\arrayrulecolor{CFaccent}
\renewcommand{\arraystretch}{1.16}
\setlength{\extrarowheight}{0.25pt}
\setlength{\aboverulesep}{0pt}
\setlength{\belowrulesep}{0pt}
\begin{tabular}{llcl}
\toprule[0.8pt]
\rowcolor{CFheaderfill}
\textcolor{CFheading}{\textbf{Category}} & \textcolor{CFheading}{\textbf{Parameter}} & \textcolor{CFheading}{\textbf{Value}} & \textcolor{CFheading}{\textbf{Description}} \\
\midrule[0.35pt]
\rowcolor{CFmetricfill}
\textcolor{CFheading}{\textbf{Model}} & \texttt{n\_cliques} & 8 & Number of latent cliques \\
 & \texttt{clique\_dim} & 16 & Dimensionality of each clique \\
 & \texttt{knot\_dim} & 1 & Overlap between adjacent cliques \\
 & \texttt{transformer\_dim} & 256 & Transformer hidden dimension \\
 & \texttt{n\_blocks} & 4 & Transformer layers \\
 & \texttt{n\_heads} & 4 & Attention heads \\
 & \texttt{n\_registers} & 2 & Register tokens \\
 & \texttt{mlp\_dim} & 128 & MLP hidden dimension \\
 & \texttt{n\_mlp} & 2 & MLP depth \\
 & \texttt{dropout\_rate} & 0.1 & Dropout probability \\
\midrule[0.35pt]
\rowcolor{CFmetricfill}
\textcolor{CFheading}{\textbf{Training}} & \texttt{gradient\_steps} & $7 \times 10^{5}$ & Number of gradient steps\\
 & \texttt{alpha\_vae} & $10^{-4}$ & KL regularization limit \\
 & \texttt{alpha\_mse} & 1 & Prediction loss weight limit \\
 & \texttt{beta\_mse} & $10^{-4}$ & Prediction loss weight init \\
 & \texttt{temp\_atom} & 1 & Atom weight in the loss \\
 & \texttt{temp\_flow} & 16 & Position weight in flow loss \\
 & \texttt{warmup} & $10^{5}$ & Linear warmup steps \\
 & \texttt{learning\_rate} & $1.4\times10^{-4}$ & Model learning rate \\
\midrule[0.35pt]
\rowcolor{CFmetricfill}
\textcolor{CFheading}{\textbf{Latent-space Optimization}} & \texttt{algorithm} & $\operatorname{ES}$ & Algorithm used  \\
 & \texttt{antithetic sampling} & $\operatorname{True}$ & If antithetic sampling used \\ 
 & \texttt{n\_pert} & 20 & Number of perturbations\\
 & \texttt{pert\_scale} & 0.05 & Scale of perturbations \\
 & \texttt{learning\_rate} & $3\times10^{-4}$ & Latent optimization LR \\
 & \texttt{design\_steps} & 2000 & Latent optimization steps \\
 & \texttt{decay} & 0.4 & Weight decay \\
 \midrule[0.35pt]
\rowcolor{CFmetricfill}
\textcolor{CFheading}{\textbf{Decoding}} & \texttt{N\_beam} & 10 & Beam search width \\
& \texttt{w\_cfg} & 2 & CFG (guidance) strength \\ 
& \texttt{N\_step} & 1000 & Number of denoising steps\\
\bottomrule[0.65pt]
\end{tabular}
\arrayrulecolor{black}
\endgroup
\end{table}

Additionally, for the band gap task, we defined the target property to be the formation energy-regularized band gap
\begin{align}
    f(\mathsf{M}) = \Delta_{\text{band}}(\mathsf{M}) + \lambda_{\text{reg}} \cdot \max\big(0, E_{\text{form}}(\mathsf{M}) - \tau_{\text{reg}}\big),\nonumber 
\end{align}
where we set $\lambda_{\text{reg}}=20$ and $\tau_{\text{reg}}=-0.2$ eV/atom, using numerical band gaps in eV and formation energies in eV/atom. 
The goal of adding the term $\lambda_{\text{reg}} \cdot \max\big(0, E_{\text{form}}(\mathsf{M}) - \tau_{\text{reg}}\big)$ was to constrain the model's search space to low formation energy materials, as such tend to be more stable. 
Below this threshold, the objective would simply reduce to the band gap, and the penalty would only be induced for materials that exceed it.

\section{Released Resources}
\label{app:resources}
The implementation and usage instructions are available on \href{https://github.com/znowu/CliqueFlowmer}{GitHub}, pretrained checkpoints and a model card on \href{https://huggingface.co/iamkuba/CliqueFlowmer}{Hugging Face}, and an interactive demo on the \href{https://cliqueflowmer.znowu.workers.dev}{CliqueFlowmer website}.
The released weights include MP-20 formation-energy and band-gap checkpoints and an additional Alexandria-trained, MP-20-fine-tuned formation-energy checkpoint. The experiments reported in this paper use the MP-20 models.

\clearpage
\section*{NeurIPS Paper Checklist}


\begin{enumerate}

\item {\bf Claims}
    \item[] Question: Do the main claims made in the abstract and introduction accurately reflect the paper's contributions and scope?
    \item[] Answer: \answerYes{} 
    \item[] Justification: The paper uses CliqueFlowmer to solve offline materials optimization.

\item {\bf Limitations}
    \item[] Question: Does the paper discuss the limitations of the work performed by the authors?
    \item[] Answer: \answerYes{} 
    \item[] Justification: We wrote a Limitations \& Conclusion section.

\item {\bf Theory assumptions and proofs}
    \item[] Question: For each theoretical result, does the paper provide the full set of assumptions and a complete (and correct) proof?
    \item[] Answer: \answerNA{} 
    \item[] Justification: No theoretical results.

    \item {\bf Experimental result reproducibility}
    \item[] Question: Does the paper fully disclose all the information needed to reproduce the main experimental results of the paper to the extent that it affects the main claims and/or conclusions of the paper (regardless of whether the code and data are provided or not)?
    \item[] Answer: \answerYes{} 
    \item[] Justification: Sections 3--4 and Appendices A--E describe the method, evaluation protocol, and hyperparameters. Appendix~\ref{appendix:dft} specifies the DFT relaxation and energy-correction procedure; Appendix~\ref{app:resources} links the implementation and pretrained models.

\item {\bf Open access to data and code}
    \item[] Question: Does the paper provide open access to the data and code, with sufficient instructions to faithfully reproduce the main experimental results, as described in supplemental material?
    \item[] Answer: \answerYes{} 
    \item[] Justification: Appendix~\ref{app:resources} links the public implementation, pretrained checkpoints, and demo. The paper cites the MP-20 dataset and baseline structures used in the experiments. The release does not include the complete raw DFT calculation archive.

\item {\bf Experimental setting/details}
    \item[] Question: Does the paper specify all the training and test details (e.g., data splits, hyperparameters, how they were chosen, type of optimizer) necessary to understand the results?
    \item[] Answer: \answerYes{} 
    \item[] Justification: Section 4 and Appendices A--E describe the data split, training and optimization settings, evaluation criteria, and DFT procedure.

\item {\bf Experiment statistical significance}
    \item[] Question: Does the paper report error bars suitably and correctly defined or other appropriate information about the statistical significance of the experiments?
    \item[] Answer: \answerNo{} 
    \item[] Justification: We report sample sizes and standard errors for the interpolation experiment, but the main optimization comparisons do not include confidence intervals or variability across independent training runs.

\item {\bf Experiments compute resources}
    \item[] Question: For each experiment, does the paper provide sufficient information on the computer resources (type of compute workers, memory, time of execution) needed to reproduce the experiments?
    \item[] Answer: \answerNo{} 
    \item[] Justification: Appendix~\ref{sec:appendix-hparams} states the training hardware, and Appendix~\ref{app:inference} reports inference timings. Complete training and DFT runtime, memory, and total-compute accounting are not provided.
    
\item {\bf Code of ethics}
    \item[] Question: Does the research conducted in the paper conform, in every respect, with the NeurIPS Code of Ethics \url{https://neurips.cc/public/EthicsGuidelines}?
    \item[] Answer: \answerYes{} 
    \item[] Justification: We follow the code of ethics.

\item {\bf Broader impacts}
    \item[] Question: Does the paper discuss both potential positive societal impacts and negative societal impacts of the work performed?
    \item[] Answer: \answerYes{} 
    \item[] Justification: These are discussed.
    
\item {\bf Safeguards}
    \item[] Question: Does the paper describe safeguards that have been put in place for responsible release of data or models that have a high risk for misuse (e.g., pre-trained language models, image generators, or scraped datasets)?
    \item[] Answer: \answerNA{} 
    \item[] Justification: No such risk.

\item {\bf Licenses for existing assets}
    \item[] Question: Are the creators or original owners of assets (e.g., code, data, models), used in the paper, properly credited and are the license and terms of use explicitly mentioned and properly respected?
    \item[] Answer: \answerNo{} 
    \item[] Justification: We cite the sources of the code, datasets, and pretrained models used in this work, but the manuscript does not enumerate the license terms for every existing asset.

\item {\bf New assets}
    \item[] Question: Are new assets introduced in the paper well documented and is the documentation provided alongside the assets?
    \item[] Answer: \answerYes{} 
    \item[] Justification: The public code includes usage instructions, the released checkpoints have a model card, and the interactive demo illustrates inference (Appendix~\ref{app:resources}).

\item {\bf Crowdsourcing and research with human subjects}
    \item[] Question: For crowdsourcing experiments and research with human subjects, does the paper include the full text of instructions given to participants and screenshots, if applicable, as well as details about compensation (if any)? 
    \item[] Answer: \answerNA{} 
    \item[] Justification: No humans involved.

\item {\bf Institutional review board (IRB) approvals or equivalent for research with human subjects}
    \item[] Question: Does the paper describe potential risks incurred by study participants, whether such risks were disclosed to the subjects, and whether Institutional Review Board (IRB) approvals (or an equivalent approval/review based on the requirements of your country or institution) were obtained?
    \item[] Answer: \answerNA{} 
    \item[] Justification: No humans involved.

\item {\bf Declaration of LLM usage}
    \item[] Question: Does the paper describe the usage of LLMs if it is an important, original, or non-standard component of the core methods in this research? Note that if the LLM is used only for writing, editing, or formatting purposes and does \emph{not} impact the core methodology, scientific rigor, or originality of the research, declaration is not required.
    \item[] Answer: \answerNA{} 
    \item[] Justification: LLMs are not a component of the research methodology. Their use for manuscript editing and formatting does not affect the core methodology.

\end{enumerate}

\end{document}